\documentclass[fleqn,10pt]{wlscirep}
\usepackage[utf8]{inputenc}
\usepackage[T1]{fontenc}
\usepackage{bm}
\usepackage{tabularx}
\usepackage{booktabs}
\usepackage{array}
\usepackage{caption}
\usepackage{siunitx}  % 用于数字对齐美化，可选
\usepackage{multirow}
\usepackage{colortbl}
\usepackage[most]{tcolorbox} %用于彩色背景方框
\usepackage{xcolor}
\usepackage{hyperref} %跳转需要

\usepackage[utf8]{inputenc}
\usepackage[T1]{fontenc}
\usepackage{CJKutf8}

\usepackage{subcaption} % 用于实现 subfigure 环境

\DeclareCaptionType{case}[Case][List of Examples]

\newcommand{\cn}[1]{\begin{CJK}{UTF8}{gbsn}#1\end{CJK}}

\definecolor{natureblue}{RGB}{220, 235, 245}
\definecolor{natureblueframe}{RGB}{100, 150, 200}
\definecolor{naturedarkblue}{RGB}{0, 80, 150}
\definecolor{answergreen}{RGB}{220, 235, 245}
\definecolor{answergreenframe}{RGB}{0, 80, 150}
\definecolor{modelred}{RGB}{220, 235, 245}
\definecolor{modelredframe}{RGB}{0, 80, 150}

\newtcolorbox{casebox}[1]{
    colback=natureblue!30!white,
    colframe=natureblueframe,
    fonttitle=\bfseries,
    coltitle=white,
    colbacktitle=naturedarkblue,
    title=#1,
    boxrule=1pt,
    rounded corners,
    drop shadow={opacity=0.4},
}

\newtcolorbox{answerbox}{
  colback=modelred!40!white,        % 深蓝背景
  colframe=modelredframe,
  arc=0mm,
  boxrule=1pt,
  fonttitle=\bfseries\small\color{white}, % 白色标题文字
  title={\textcolor{white}{Answer}},
  top=1mm,
  bottom=1mm,
  left=1mm,                 % 左内边距（默认更大，调小）
  right=1mm, 
  boxsep=0.4mm,
  coltext=black,               % 内容文字黑色
  fontupper=\small % <<< 新增：将内容字体设置为 small 尺寸
}

\newtcolorbox{modelbox}{
  colback=modelred!40!white,        % 深蓝背景
  colframe=modelredframe,
  arc=0mm,
  boxrule=1pt,
  fonttitle=\bfseries\small\color{white}, % 白色标题文字
  title={\textcolor{white}{Output}},
  top=1mm,
  bottom=1mm,
  left=1mm,                 % 左内边距（默认更大，调小）
  right=1mm, 
  boxsep=0.4mm,
  coltext=black,               % 内容文字黑色
  fontupper=\small % <<< 新增：将内容字体设置为 small 尺寸
}

\newtcolorbox{analysisbox}{
  colback=modelred!40!white,        % 深蓝背景
  colframe=modelredframe,
  arc=0mm,
  boxrule=1pt,
  fonttitle=\bfseries\small\color{white}, % 白色标题文字
  title={\textcolor{white}{Analysis}},
  top=1mm,
  bottom=1mm,
  left=1mm,                 % 左内边距（默认更大，调小）
  right=1mm, 
  coltext=black,               % 内容文字黑色
  fontupper=\small % <<< 新增：将内容字体设置为 small 尺寸
}

\newtcolorbox{analysistwobox}{
    colback=white,
    colframe=naturedarkblue,
    arc=0mm,
    borderline={1pt}{0pt}{naturedarkblue},
    borderline={1pt}{2pt}{natureblueframe, dashed},
    fontupper=\small,
    top=1mm,
    bottom=1mm,
    left=1mm,                 % 左内边距（默认更大，调小）
    right=1mm
}

\definecolor{level1Color}{RGB}{255,214,10}  % 亮黄色
\definecolor{level2Color}{RGB}{41,167,69}   % 亮绿色
\definecolor{level3Color}{RGB}{237,30,121}  % 亮粉色
\definecolor{totalColor}{RGB}{22,160,133}   % 亮青色
\definecolor{taskColor}{RGB}{155,89,182}    % 亮紫色
\definecolor{headerColor}{RGB}{52,152,219}  % 亮蓝色（表头）

\definecolor{thoughtbg}{HTML}{F4F4F9}
\definecolor{toolbg}{HTML}{E7F3FF}
\definecolor{obsbg}{HTML}{F0FFF4}
\definecolor{stepbg}{HTML}{4A4A4A}
\definecolor{analysisbg}{HTML}{FDF2F2}

\newtcolorbox{thoughtbox}{
    colback=thoughtbg, colframe=thoughtbg, arc=0pt, outer arc=0pt,
    boxrule=0pt, breakable, left=5pt, right=5pt, top=5pt, bottom=5pt, fontupper=\small
}

\newtcolorbox{toolbox}{
    colback=toolbg, colframe=toolbg, arc=0pt, outer arc=0pt,
    boxrule=0pt, breakable, left=5pt, right=5pt, top=5pt, bottom=5pt, fontupper=\small
}

\newtcolorbox{obsbox}{
    colback=obsbg, colframe=obsbg, arc=0pt, outer arc=0pt,
    boxrule=0pt, breakable, left=5pt, right=5pt, top=5pt, bottom=5pt, fontupper=\small
}

\usepackage[most]{tcolorbox}
\usepackage{xcolor}
\usepackage{graphicx}
\usepackage{caption}
\usepackage{CJKutf8}

\definecolor{HdrGray}{RGB}{55,55,55}        % header bar
\definecolor{PastelBlue}{RGB}{232,244,255}  % light blue body
\definecolor{PastelGreen}{RGB}{232,248,240} % light green body
\definecolor{PastelGray}{RGB}{246,246,246}  % light gray body
\definecolor{TransGray}{RGB}{105,105,105}

\newcommand{\TransLabel}{\textcolor{TransGray}{\textbf{Trans.:}}}
\newcommand{\TransText}[1]{\textcolor{TransGray}{#1}}

\newtcolorbox{CaseWrap}[1][]{%
  enhanced,
  breakable,
  colback=white,
  colframe=black,
  boxrule=0.8pt,
  sharp corners,
  left=2.2mm,right=2.2mm,top=2.0mm,bottom=2.0mm,
  boxsep=0pt,
  before skip=2pt,
  after skip=2pt,
  #1
}

\newtcolorbox{CaseHdrBox}[1]{%
  enhanced,
  breakable,
  colback=HdrGray,
  colframe=HdrGray,
  boxrule=0pt,
  sharp corners,
  width=\linewidth,
  left=2.0mm,right=2.0mm,top=1.0mm,bottom=1.0mm,
  boxsep=0pt,
  before skip=0pt,
  after skip=2.2pt,
  fontupper=\bfseries\small\color{white}\centering,
}
\newcommand{\CaseHdr}[1]{\begin{CaseHdrBox}{}#1\end{CaseHdrBox}}

\newtcolorbox{CaseBody}[1][]{%
  enhanced,
  breakable,
  colback=PastelBlue,
  colframe=PastelBlue,
  boxrule=0pt,
  sharp corners,
  width=\linewidth,
  left=2.0mm,right=2.0mm,top=1.2mm,bottom=1.2mm,
  boxsep=0pt,
  before skip=0pt,
  after skip=4.0pt,
  #1
}

\title{Cognitive Mismatch in Multimodal Large Language Models for Discrete Symbol Understanding}
\author[1]{Yinghui Li}
\author[2]{Jiayi Kuang}
\author[1]{Peng Xing}
\author[1]{Daixian Liu}
\author[4]{Yongheng Zhang}
\author[3]{\\Junnan Dong}
\author[1]{Shu-Yu Guo}
\author[1]{Yangning Li}
\author[4]{Qingyu Zhou}
\author[5]{Wenhao Jiang}
\author[1]{\\Hai-Tao Zheng}
\author[2]{Ying Shen}
\author[2]{Liang Lin}
\author[6]{Philip S. Yu}
\affil[1]{Tsinghua University, China}
\affil[2]{Sun Yat-sen University, China}
\affil[3]{The Hong Kong Polytechnic University, China}
\affil[4]{Independent Researcher}
\affil[5]{Guangdong Laboratory of AI and Digital Economy (SZ), China}
\affil[6]{University of Illinois Chicago, USA}

\affil[*]{E-mail: liyinghuihhh@gmail.com}

\begin{abstract}
Multimodal large language models (MLLMs) perform strongly on natural images, yet their ability to understand discrete visual symbols remains unclear. We present a multi-domain benchmark spanning language, culture, mathematics, physics and chemistry, organized into three cognitive levels: perception and recognition, combination and reasoning, and association and critical thinking. Across leading MLLMs, we observe a consistent cognitive mismatch. Models frequently underperform on elementary symbol recognition while appearing relatively competent on more complex reasoning tasks. This recognition-reasoning inversion indicates that current systems often compensate with linguistic priors, template retrieval or procedural reasoning instead of robust visual grounding. The pattern is especially clear for sparse, low-redundancy symbols such as handwritten characters, formula graphs, circuit diagrams and chemical structures. These results show that symbolic understanding remains a major bottleneck for multimodal intelligence and motivate training and evaluation schemes that prioritize grounded perception in discrete semantic spaces.
\end{abstract}
\begin{document}

% Keywords: Multimodal Large Language Models, symbolic reasoning, visual perception, artificial intelligence, scientific understanding, discrete semantics

\flushbottom
\maketitle

\thispagestyle{empty}

\vspace{-10mm}

\section*{Introduction}

Since the advent of the Large Language Models era in Artificial Intelligence, Multimodal Large Language Models (MLLMs) have consistently remained one of the most cutting-edge and prominent research topics~\cite{caffagni2024revolution, song2025bridge, zhang2024mm}. Beyond textual media, MLLMs aim to endow artificial systems with the ability to see, perceive and reason about the physical world, thereby moving toward a more comprehensive form of intelligence. In recent years, the rapid rise of embodied intelligence has further elevated the significance of MLLMs~\cite{xu2025embodied, turgunbaev2025perception, szot2025multimodal}. The fundamental reason behind this trend is that the cognitive paradigm represented by MLLM is closer to human intelligence in understanding the world and represents an essential step towards achieving Artificial General Intelligence (AGI)~\cite{du2025human, fei2024multimodal, shen2025zoomeye}. Consequently, understanding and emulating fundamental mechanisms of human cognition in the real world has become crucial for advancing MLLMs.

Symbols have been the indispensable cornerstone of human cognition and the evolution of intelligence since the dawn of the human species~\cite{perlovsky2007symbols}. From prehistoric cave paintings encoding survival knowledge to the structured syntax of natural language, symbolic systems allow humans to transcend the limits of individual sensory experience and accumulate abstract knowledge across generations~\cite{taniguchi2018symbol, bickerton2003symbol, miyagawa2018cross}. As cognitive science represented by semiotic theories has elucidated, human intelligence is inherently symbolic, relying on the creation, manipulation and communication of symbols to support reasoning and collective understanding~\cite{rapoport1955role, ahn2017signals}. For multimodal intelligence, the open question is therefore not simply whether a model can produce a plausible answer, but whether it can truly perceive discrete symbols and ground its reasoning in the correct symbolic evidence.

\begin{figure}
    \centering
    \includegraphics[width=0.76\linewidth]{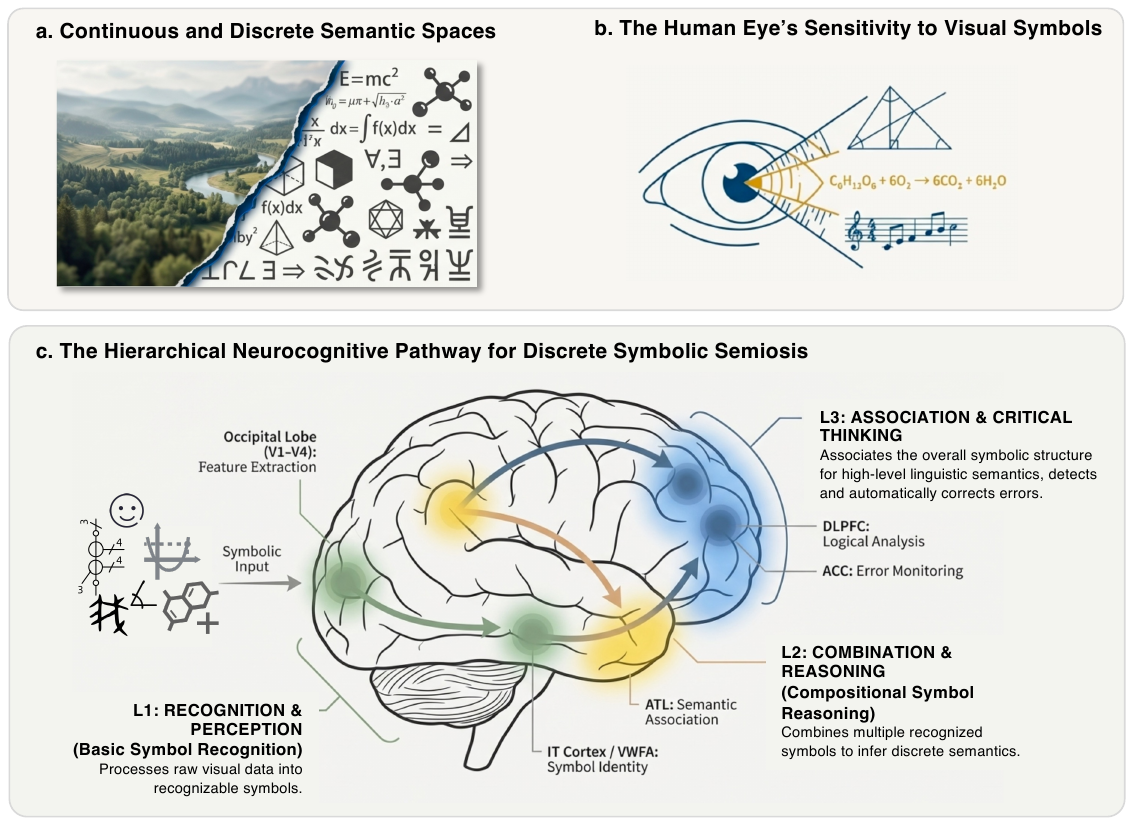}
    \caption{Continuous and discrete semantic spaces differ fundamentally in how visual information is organized. Discrete symbol understanding requires precise recognition of individual symbolic units and their structured relations.}
    \label{fig:continuous_vs_discrete_comparison}
    \vspace{-4mm}
\end{figure}

The distinction is especially sharp between continuous and discrete semantic spaces. Mainstream MLLMs are primarily trained on natural scenes and image-text pairs in which visual inputs map to relatively holistic semantic narratives, as in captioning, visual question answering and grounding~\cite{abdulgalil2025next, kuang2025natural, xiao2024towards}. By contrast, the images we focus on are composed of semantically independent symbolic units whose meaning depends on exact recognition and structured composition. In a handwritten Chinese sentence, a function graph, a circuit diagram or a chemical structure, a single misread stroke, bond or relation can overturn the whole interpretation. This representational gap is illustrated in Fig.~\ref{fig:continuous_vs_discrete_comparison}, which contrasts continuous semantic spaces with the discrete symbolic processing route that is far more natural to human cognition.

Across prior research on MLLMs, investigations into discrete symbols remain notably scarce. Existing evaluation suites cover broad multimodal understanding, OCR-heavy perception, expert reasoning and scientific problem solving~\cite{liu2024mmbench, li2024seed, fu2025mme, yue2024mmmu, li2025survey, xu2024lvlm, fu2025ocrbench, lu2024mathvista, zhang2024mathverse}. Specialized methods have also begun to address symbolic domains such as documents, music, art, geometry, chemistry and other structured scientific inputs~\cite{liu2024textmonkey, yu2024texthawk, nacson2025docvlm, tang-etal-2025-nota, jiang2025multimodal, fanelli2025artseek, gao2025gllava, shi-etal-2024-math, pan2025enhancing, shi2025multimodal, tan2025chemmllm, li2025chemvlm, zhao2024chemdfm}. However, these approaches remain fragmented across specific domains and rarely disentangle whether models truly recognize the symbols they manipulate.

To address this problem, we construct a unified benchmark spanning five symbolic domains, each organized into three cognitive levels: \emph{perception and recognition}, \emph{combination and reasoning}, and \emph{association and critical thinking}. This hierarchy follows the progressive route by which humans process symbols, from recognizing local visual units to integrating them compositionally and finally judging their semantic consistency in context. Across these domains, we find a recurrent and counterintuitive pattern: models often fail at foundational symbol recognition yet produce apparently competent higher-level reasoning. This recognition-reasoning inversion is the core empirical signature of the cognitive mismatch we study.

Our benchmark covers language symbols, cultural symbols, mathematical symbols, physical symbols and chemical symbols, thereby spanning both the social-scientific and natural-scientific symbolic systems that structure human knowledge. The task design is intentionally hierarchical. At the first level, models must identify the symbolic semantics of individual units, such as malformed handwritten characters, graph elements, circuit primitives or atoms and bonds in a molecular structure. At the second level, they must integrate multiple symbols and reason over their joint semantics, for example by combining local evidence in a graph or diagram into a higher-order conclusion. At the third level, they must move beyond straightforward recognition and perform correction, consistency judgment or culturally mediated interpretation. This progression allows us to ask not only whether a model can answer a question, but also what cognitive route is likely supporting that answer.

\begin{figure}
    \centering
    \includegraphics[width=0.90\linewidth]{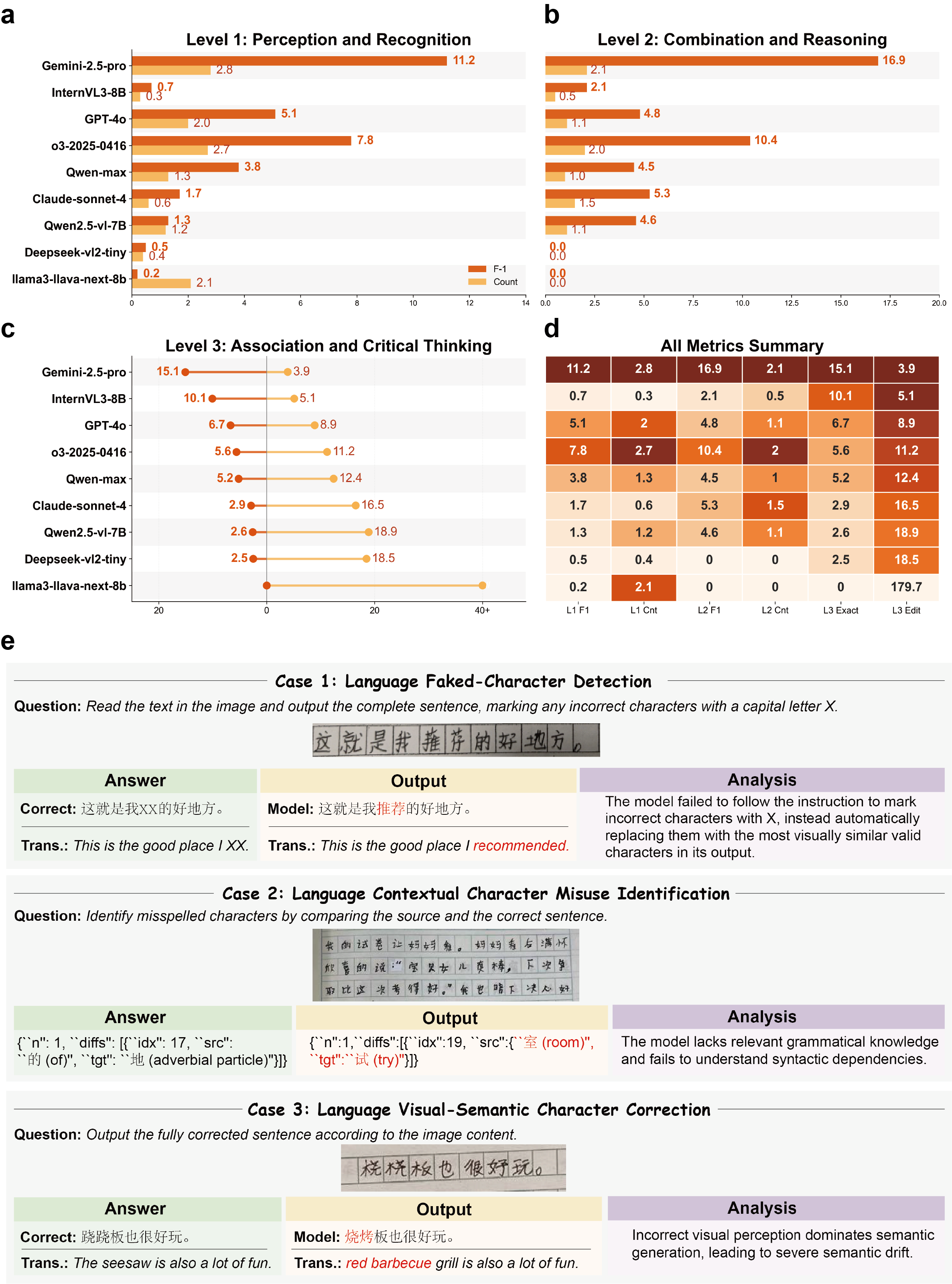}
    \caption{Performance and representative cases for language-symbol understanding across three cognitive levels. Models are evaluated on faked-character detection, misspelled-character detection and visual-semantic correction.}
    \label{fig:language_results}
\end{figure}

Using this design, we curate a large-scale benchmark containing more than 13,000 image-question-answer instances and 38 sub-tasks. The central motivation is diagnostic rather than competitive. In natural images, an approximately correct semantic summary is often sufficient; in discrete symbolic images, a minor local deviation can cause a total semantic shift. A missing stroke can turn a correct character into an invalid one, an omitted bond can alter a molecule, and a misread axis relation can invalidate a mathematical conclusion. This property makes discrete semantic spaces especially suitable for testing whether MLLMs truly ground their outputs in visual evidence, or instead rely on linguistic priors, memorized templates and shortcuts.

This perspective also explains why we place equal emphasis on social-symbolic and scientific-symbolic domains. Language and cultural symbols probe whether models can handle irregularity, convention and context-sensitive meaning in settings close to everyday cognition. Mathematics, physics and chemistry, by contrast, stress formal structure, low tolerance for local error and the need to preserve symbolic consistency over longer reasoning chains. Bringing these domains together lets us examine whether current MLLMs possess a general capacity for symbolic understanding, or whether their successes remain local, domain-dependent and heavily contingent on the kinds of patterns most represented in pre-training data.

\section*{Results}

\subsection*{Language symbols}

Language symbols expose the most immediate gap between visual perception and language-driven generation. As shown in Figure~\ref{fig:language_results}, performance remains weak across all three levels. In the faked-character detection task, the overall performance of most models is extremely poor, often failing to distinguish an invalid handwritten character from its nearest legal counterpart. The dominant failure mode is forced normalization: the model silently repairs an anomalous glyph into the closest familiar character and thereby erases the very evidence it is asked to detect. A second failure mode is unstable localization, in which models follow the instruction to mark errors but confuse legal and illegal characters because they lack fine-grained structural discrimination. Both behaviors indicate that the symbolic image is being collapsed into a coarse linguistic prior before a reliable character-space representation is formed.

This weakness further propagates into the more difficult tasks of misspelled-character detection and visual-semantic correction. Only a few models achieve usable exact-match accuracy, and even then, many corrections are only directionally relevant rather than fully correct. Once the first visual decision is wrong, later correction becomes semantically unstable: a misperceived character can trigger an entirely unrelated lexical continuation. The language domain therefore, provides a particularly clean and compelling illustration of our central finding. Higher-level outputs may appear fluent, but this outward fluency often masks a fundamental failure to anchor the answer rigorously in the correct symbolic evidence. A fuller breakdown of the language results is provided in Supplementary Sec.~\ref{Sec:SupplementaryLanguage}.

Qualitative cases reinforce this interpretation. Some models do not treat malformed characters as anomalies at all, but instead automatically replace them with the nearest valid glyph during generation, effectively overwriting the perceptual evidence at the earliest stage. Others attempt to follow the task instruction but still fail to separate structurally incorrect forms from merely difficult handwriting, producing a high number of predicted error positions with poor precision. In the correction setting, the same instability can lead to severe semantic drift: once a character is misread, the model may continue with a semantically coherent but irrelevant phrase. This shows that the major obstacle is not only linguistic correction itself, but the absence of a stable mechanism for maintaining semantic consistency from visual symbol recognition to textual output.

The language tasks, therefore, illuminate a broader issue in multimodal learning. Because large language models are optimized to produce well-formed text, they appear inclined to normalize uncertainty into familiar lexical forms. That tendency is useful for open-ended generation, but detrimental when the task requires preserving anomaly, ambiguity, or fine-grained deviation. In this sense, the weakness of handwritten symbols is not a marginal OCR problem. It is a structural sign that current MLLMs still struggle to keep visual uncertainty alive long enough for higher-level reasoning to operate on it faithfully.

\subsection*{Cultural symbols}

\begin{figure}
    \centering
    \includegraphics[width=0.98\linewidth]{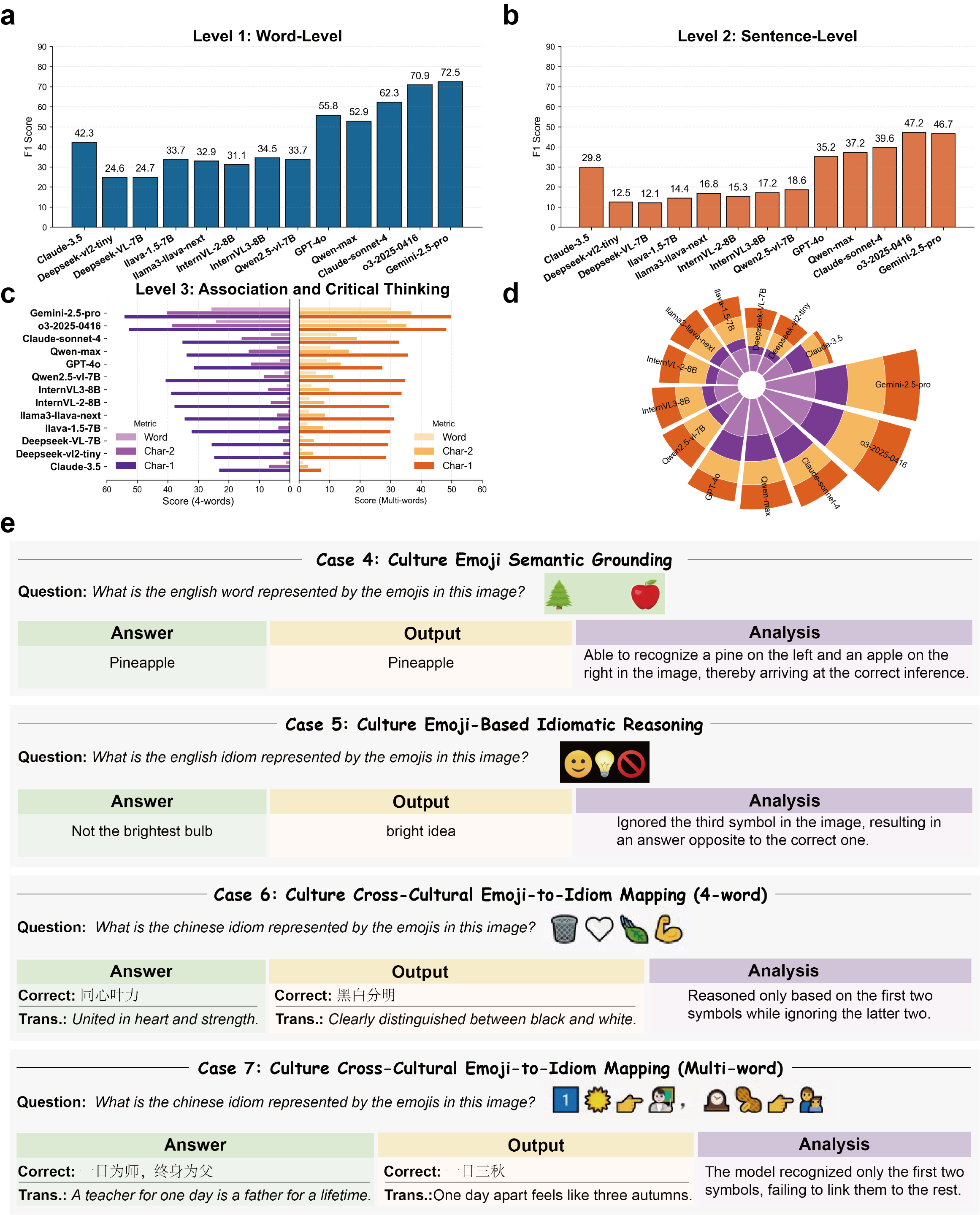}
    \caption{Performance and representative cases for cultural-symbol understanding. The tasks progress from lexical grounding of emojis to idiomatic composition and culturally mediated interpretation.}
    \label{fig:culture_exp}
\end{figure}

Compared with handwritten characters, emoji-based tasks produce stronger low-level performance, especially for common English expressions. As summarized in Figure~\ref{fig:culture_exp}, models often recover the literal semantics of individual emojis and, in simple cases, compose them into familiar words, suggesting that high-frequency and socially standardized visual symbols are easier to align with language than sparse handwritten forms. Even so, the apparent strength is fragile. Once a sequence requires the model to preserve a negative operator, a discourse constraint or a culturally conventional reading, many systems fall back on the most statistically accessible phrase rather than the meaning actually licensed by the whole symbolic configuration in context.

The limitation becomes more evident in Chinese idiom tasks, where correct interpretation depends not only on object semantics but also on homophony, cultural convention and cross-symbol associative reasoning. Here, character-level overlap remains substantially higher than exact idiom recovery, showing that models can often map isolated emojis to local meanings while failing to complete the intended symbolic leap. In other words, the bottleneck is no longer simple recognition of the visual token itself, but the controlled integration of multiple symbolic cues without hallucinating a more familiar expression.

This domain is especially revealing because it sits between literal recognition and culturally grounded semiosis. In English emoji tasks, some models already show competent low-level composition, indicating that they can align familiar pictographic symbols with lexical units learned from web-scale corpora. However, once the task requires suppressing an overly salient local meaning, performance drops rapidly. In Chinese idiom tasks the challenge is even sharper: correct answers may depend on object semantics, homophonic correspondence and shared cultural convention at the same time. The resulting gap between partial character recovery and full idiom recovery indicates that many models can access fragments of the intended semantics, yet remain unable to consolidate them into the culturally appropriate symbolic interpretation.

This pattern matters because cultural symbols are often treated as comparatively easy for multimodal models. Our results suggest a more nuanced picture. Emoji understanding is indeed easier when the task can be reduced to lexical substitution, but it remains difficult when symbolic interpretation must be constrained by context or by a social convention that cannot be read directly from the icon itself. The cultural domain thus shows that even highly familiar visual symbols can expose the limits of multimodal grounding once meaning becomes relational, conventional and compositionally constrained. Detailed analyses for both English emoji composition and Chinese idiom interpretation are given in Supplementary Sec.~\ref{Sec:SupplementaryCultural}.

\subsection*{Mathematical symbols}

\begin{figure}
    \centering
    \includegraphics[width=1\linewidth]{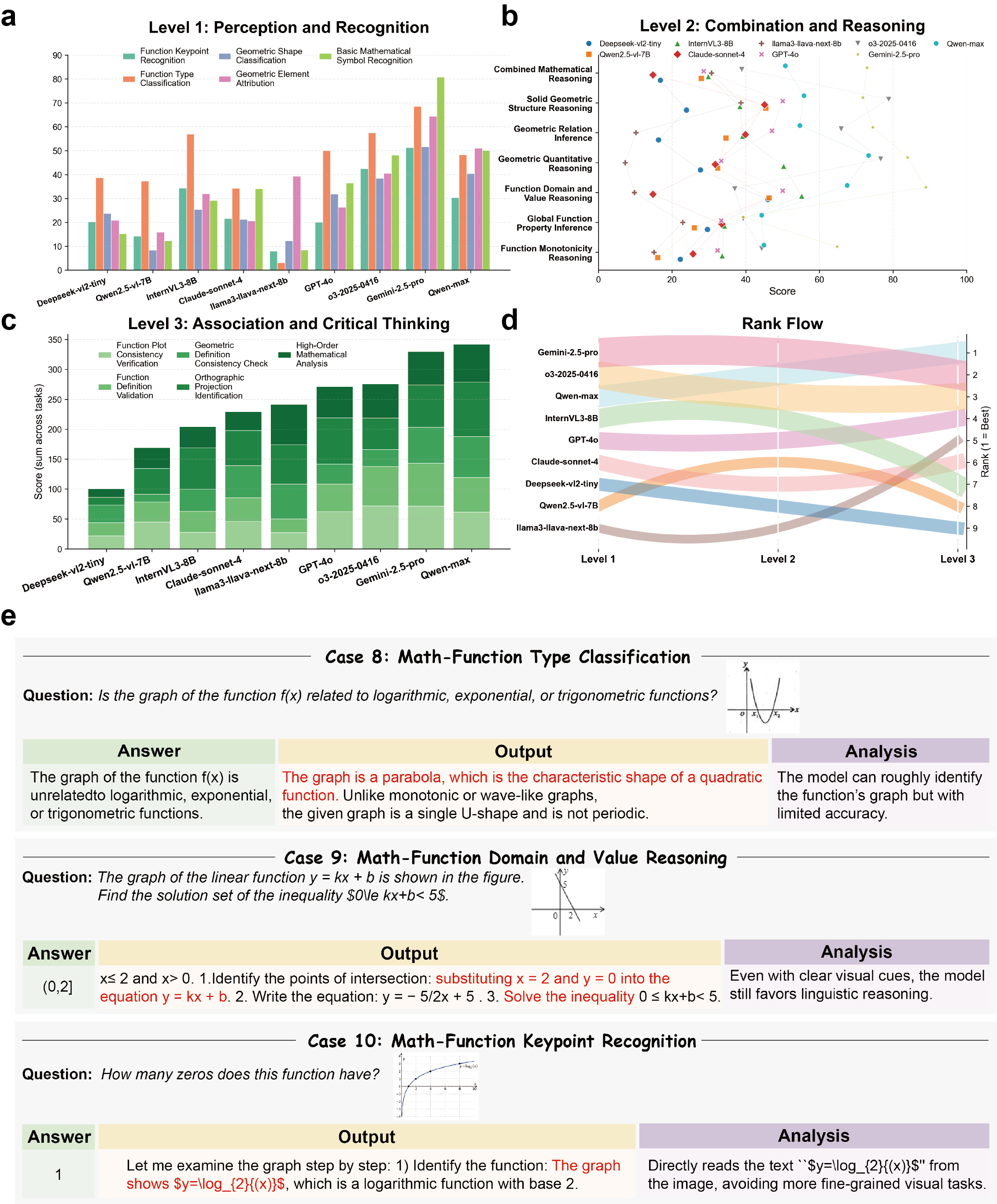}
    \caption{Performance and representative cases for mathematical-symbol understanding, covering function graphs and geometric figures across recognition, reasoning, and verification tasks.}
    \label{fig:math_exp}
\end{figure}

Mathematical symbols reveal a sharp inversion between low-level perception and apparently stronger higher-level reasoning. Figure~\ref{fig:math_exp} illustrates this contrast. In Level 1 tasks, models frequently miss basic visual distinctions such as function type, graph shape or local geometric attributes, because these tasks demand precise localization and have extremely low error tolerance. Yet they sometimes perform better on Level 2 and 3 tasks, where linguistic rules and familiar mathematical templates can partially compensate for visual uncertainty. Models may fail to read a curve faithfully while still producing a plausible explanation by invoking properties of logarithmic, exponential or quadratic functions from prior knowledge.

Qualitative cases show that even when the answer is correct, the path is often language-dominant rather than image-dominant. Models tend to bypass direct visual judgment and instead reconstruct the problem through symbolic calculation, textual cues or memorized proof patterns. This behavior explains why stronger reasoning scores should not be read as evidence of stronger perception. Rather, the mathematical domain shows very clearly how procedural inference can conceal deficits in the visual grounding of graphs, shapes and spatial relations.

The inversion is informative as it suggests poor Level 1 performance reflects not total incapacity, but a language-dominant preference. When questions contain enough textual cues or resemble familiar mathematical patterns, the model often mobilizes internal rule knowledge to compensate for incomplete perception. However, this compensation is selective and unstable. It works best when a problem is reconstructed from stereotyped functional or geometric schemas, yet fails when success depends on faithfully reading local visual structure. Thus, the mathematical domain reveals a core MLLM contradiction: stronger symbolic reasoning coexists with weaker symbolic seeing, because the former frequently bypasses the latter.

Meanwhile, the mathematical results caution against interpreting reasoning traces at face value. A long and apparently rigorous derivation can coexist with a weak visual parse of the graph or geometric figure. In several representative cases, the model's explanation sounds mathematically competent because it is anchored in general rule knowledge, not because it has securely extracted the local visual details required by the task. This is why mathematical symbolic evaluation must distinguish between procedural fluency and grounded diagram understanding rather than treating them as the same capability in practice. Further mathematical analyses are reported in Supplementary Sec.~\ref{Sec:SupplementaryMathematical}.

\subsection*{Physical symbols}

\begin{figure}
    \centering
    \includegraphics[width=0.9\linewidth]{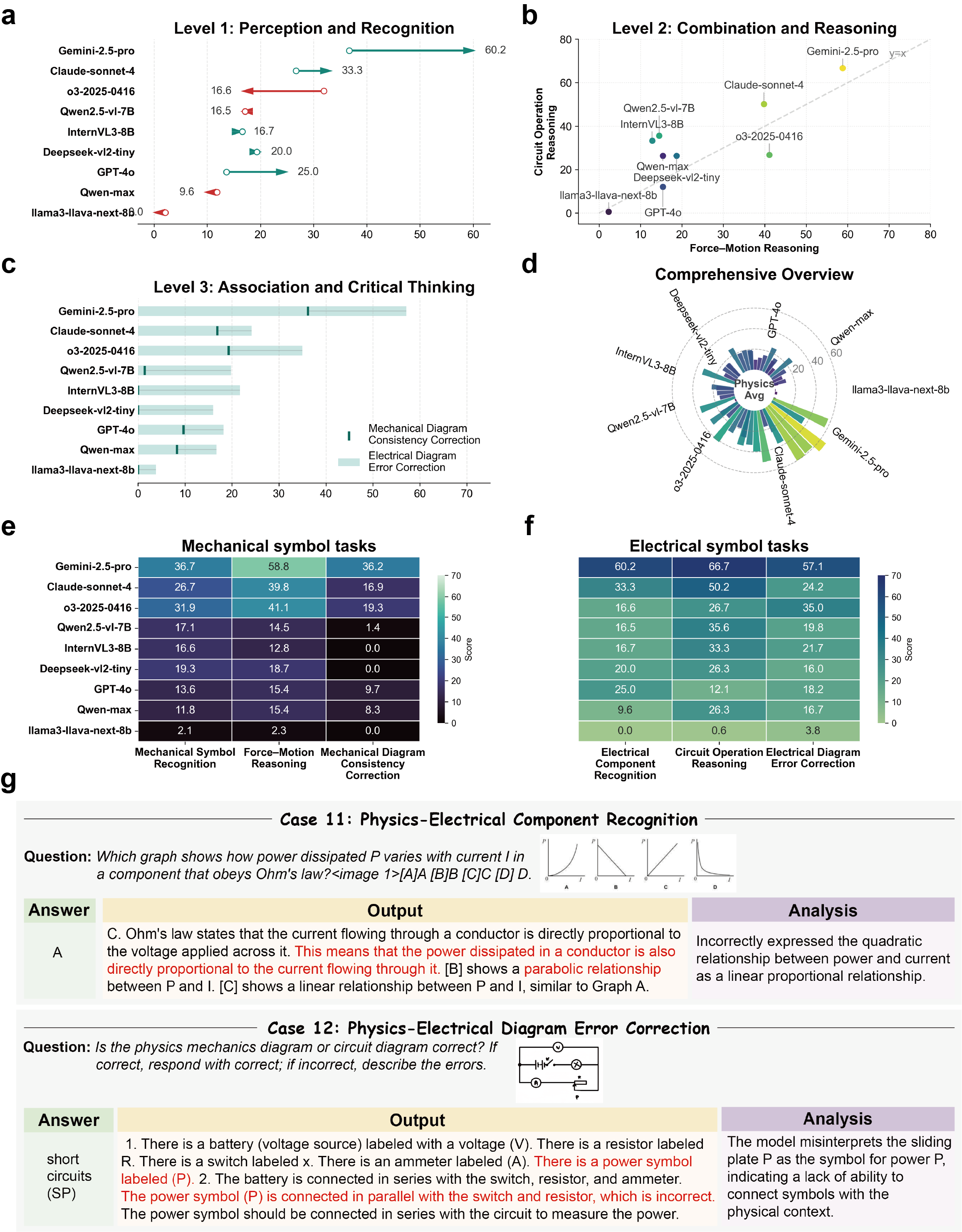}
    \caption{Performance and representative cases for physical-symbol understanding across mechanics and electromagnetism. Sparse physical symbols remain difficult for most models, especially in visually grounded mapping tasks.}
    \label{fig:physics_results}
\end{figure}

Physics tasks show that sparse scientific symbols remain difficult even when the underlying verbal knowledge is familiar. As Figure~\ref{fig:physics_results} indicates, most models perform poorly on basic recognition of formulas, graphs and circuit relations, indicating that they struggle to convert visual physical symbols into stable mathematical or conceptual representations. Many responses can correctly recite the relevant law in words but still fail at the symbolic mapping step that the question actually requires. A model may invoke Ohm's law, for example, yet still confuse the functional form of the quantity being plotted or substitute an incorrect parameter while otherwise following a plausible reasoning chain.

This gap between verbal knowledge and symbolic execution widens with multi-step reasoning. In several cases, the reasoning trace remains locally coherent but is derailed by an early visual error, such as a misread constant, a misidentified graph or an omitted structural relation in the diagram. The physics domain therefore highlights that scientific reasoning in multimodal systems depends not only on correct laws, but also on preserving symbolic consistency from the first visual parse.

The physics results also show that correct verbal knowledge alone is not sufficient for grounded multimodal competence. Some models can reproduce textbook definitions or standard physical principles with impressive fluency, yet once asked to translate a concrete visual configuration into the appropriate symbolic relation, their reasoning becomes unstable. In this sense, physical symbols expose a separation between knowing a law and reading the specific instantiation of that law from a diagram. The more sparsely encoded and low-redundancy the visual representation becomes, the more visible this weakness is, especially in tasks involving circuits, mechanics trajectories and graph-based quantity mapping.

From the perspective of scientific use, this distinction is critical. Physics problems often appear visually simple while actually depending on exact symbolic correspondences among quantities, directions, circuit components and geometric constraints. A system that can narrate the correct principle but cannot preserve those correspondences is not merely making a small perceptual mistake; it is failing at the core interface between visual evidence and formal reasoning. The physical-symbol domain therefore highlights why robust multimodal science requires more than verbal scientific literacy. Supplementary Sec.~\ref{Sec:SupplementaryPhysical} provides the extended physics analyses and representative error patterns.

\subsection*{Chemical symbols}

\begin{figure}
    \centering
    \includegraphics[width=0.95\linewidth]{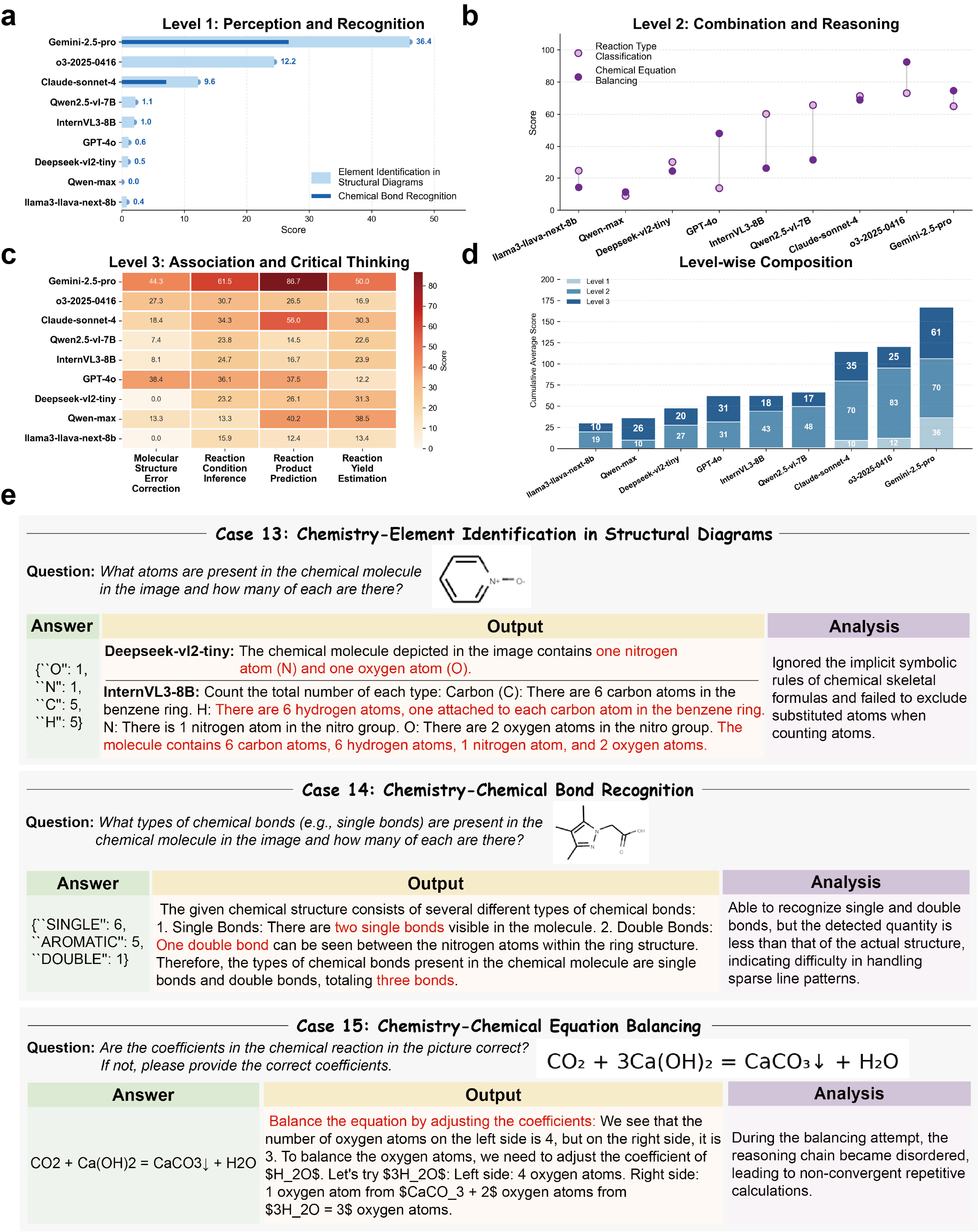}
    \caption{Performance and representative cases for chemical-symbol understanding. The benchmark examines molecular structure recognition, reaction reasoning and higher-level correction or prediction.}
    \label{fig:chemistry_exp}
\end{figure}

Among all domains, chemistry makes the structural nature of complex symbolic understanding especially explicit. Figure~\ref{fig:chemistry_exp} shows that in Level 1 tasks, many models can identify salient individual atom labels or obvious bond types, but they often ignore the implicit topological rules of skeletal formulas. Carbon and hydrogen counts are therefore routinely omitted or inferred incorrectly, aromatic structures are flattened into simpler bond categories, and local visual fragments are combined through memorized templates rather than performing true structural parsing. These errors fundamentally show that current visual encoders still struggle with sparse lines, small overlaps and convention-dependent notation today.

When the task shifts to balancing reactions, identifying conditions or predicting products, performance becomes more dependent on rule-based integration. Some models can detect that a reaction is unbalanced or that a low-temperature reagent is implicated, but their internal reasoning often remains pattern-based rather than causally grounded. Correct answers are possible, yet unstable: once an earlier bond, substituent or coefficient is misread, the downstream chemical chain collapses. Chemistry therefore provides one of the clearest demonstrations that symbolic reasoning without robust symbolic perception is inherently brittle and difficult to generalize.

At the same time, chemistry shows that stronger reasoning-oriented models can sometimes outperform visually stronger models in intermediate tasks that rely heavily on rule integration. This divergence suggests that visual recognition ability and symbolic reasoning ability are not advancing in lockstep. A model may detect that a reaction needs balancing or invoke plausible chemical heuristics, yet still fail to ground those operations in the exact molecular or reaction structure shown in the image. Conversely, a model with better local parsing may still underperform if it cannot organize those symbols into a stable reasoning chain. The domain therefore makes visible the two-sided nature of the cognitive mismatch: current models are weak both at seeing fine-grained chemical symbols and at preserving structural consistency when reasoning over them.

Chemistry is especially valuable here because symbolic compression is unusually high: tiny visual marks encode valency, substitution, reaction conditions and transformation pathways. As a result, a model cannot rely on broad scene semantics to recover the missing meaning. It must preserve exact local structure. The persistent failures on skeletal formulas and reaction diagrams therefore provide strong evidence that the current multimodal pipeline still lacks an effective mechanism for converting sparse visual notation into a stable symbolic substrate on which domain reasoning can reliably operate. Expanded chemical results are presented in Supplementary Sec.~\ref{Sec:SupplementaryChemical}.

\section*{Discussion}

Across five domains, the same pattern recurs. Current MLLMs do not fail merely because symbolic tasks are difficult in a generic sense; they fail because discrete symbol understanding demands a kind of visual precision that can no longer be absorbed into broad language plausibility. Human participants follow a more canonical cognitive trajectory, with high performance on foundational perceptual tasks and gradually lower performance as reasoning becomes more demanding. Many models, by contrast, exhibit the reverse tendency: they underperform on the perceptual baseline and then appear to recover on tasks where linguistic priors, procedural templates or contextual inference can compensate for missing visual evidence. This recognition-reasoning inversion is the operational form of the cognitive mismatch identified in this work.

The implication is not that current MLLMs lack reasoning ability. On the contrary, several models display strong semantic flexibility and can sometimes repair early perceptual errors through downstream inference. The problem is that this compensatory mechanism makes end-task success an unreliable indicator of genuine visual understanding. A model may arrive at the correct answer for the wrong reason, or may fail catastrophically when symbolic precision cannot be approximated by linguistic expectation. This helps explain why sparse handwritten characters, formula graphs, circuit diagrams and chemical structures remain persistent failure points despite impressive progress on natural-image benchmarks.

More broadly, our findings suggest that the dominant training paradigm still privileges jump mapping from vision to language concepts over the formation of stable symbolic visual primitives. Advancing multimodal intelligence in scientific and abstract domains will therefore require stronger supervision on discrete symbols, tighter coupling between perceptual evidence and reasoning, and evaluation protocols that explicitly separate grounded recognition from compensatory inference. Extended related work, fuller per-domain analyses, human-baseline details and additional case studies are provided in Supplementary Secs.~\ref{Sec:SupplementaryRelatedWork},~\ref{Sec:SupplementaryResults},~\ref{Sec:SupplementaryHumanBaseline} and~\ref{Sec:SupplementaryCases}.

The human baseline is important in this regard. Human participants do not show the same inversion pattern. Instead, they follow a more intuitive cognitive trajectory: perceptually grounded tasks are solved well, while performance gradually declines as association, correction and multi-step reasoning become more demanding. This contrast suggests that the current MLLM pipeline does not simply constitute a weaker version of human symbolic cognition; rather, it follows a qualitatively different route, one in which language prior often dominates over visual grounding. That difference matters because it changes how errors emerge. Human symbolic failures often occur after a symbol has been correctly perceived, whereas model failures frequently originate in the earliest perceptual step and are only later concealed by fluent reasoning.

These observations also have implications for model design. The issue is unlikely to be explained by data scarcity alone. Current visual encoders are optimized for continuous natural scenes, in which coarse semantic summaries are often sufficient, but discrete symbols demand preservation of topological and local structural information at a much finer granularity. Future progress will therefore likely require not only more data, but also training objectives and architectural biases that explicitly encourage the formation of stable symbolic visual primitives and force reasoning modules to remain anchored to visual evidence. Without such changes, models may continue to improve on benchmark-style reasoning while remaining fragile on the symbol systems that underpin human science, language and culture.

For evaluation, the same lesson applies. If benchmarks emphasize only final-answer correctness, then compensatory inference can easily be mistaken for grounded understanding. A model may appear strong because it reaches the right answer through prior knowledge, semantic plausibility or pattern completion, even when the underlying visual parse is wrong. Our benchmark is designed to make that discrepancy visible by separating recognition, reasoning and critical symbolic judgment across multiple domains. In doing so, it aims not only to compare current models, but also to provide a clearer target for future multimodal systems that must genuinely perceive and manipulate the discrete symbolic structures through which humans organize knowledge more faithfully.

\section*{Methods}
\label{Sec: Methods}

\subsection*{Benchmark design}

We construct a comprehensive and multidisciplinary benchmark for discrete semantic spaces across five foundational domains: language, culture, mathematics, physics and chemistry. The design follows a rigorous three-level cognitive hierarchy inspired by established principles of human symbolic processing. Level 1, \emph{Perception and Recognition}, tests whether a model can identify the meaning of individual symbolic units from their visual form. Level 2, \emph{Combination and Reasoning}, examines whether multiple symbols can be integrated into a coherent structured interpretation. Level 3, \emph{Association and Critical Thinking}, evaluates whether the model can detect anomalies, correct errors or infer unconventional meanings from symbolic context.

This hierarchy lets us distinguish the nuanced mechanics of visual grounding from downstream reasoning. In continuous natural images, approximate semantic alignment is often sufficient. In discrete symbolic images, however, a small local mistake can alter the global answer entirely, leading to a complete failure of the reasoning chain. Our task design therefore emphasizes precision, compositional structure and cross-level diagnosis rather than focusing only on superficial end-to-end success. The complete task framework is detailed in Supplementary Sec.~\ref{Sec:SupplementaryTaskDesign}.

\subsection*{Task construction across domains}

For language symbols, we evaluate faked-character detection, misspelled-character detection and visual-semantic correction, focusing on handwritten Chinese characters with strong visual compositionality and structural complexity. For cultural symbols, we use emoji sequences to test lexical grounding, idiomatic composition in English and culturally mediated idiom inference in Chinese. For mathematical symbols, we cover function graphs and geometric figures, progressing from entity recognition to property reasoning and consistency checking. For physical symbols, we design tasks in mechanics and electromagnetism that require interpretation of intricate graphs, formulas, schematics and their interdependent relations. For chemical symbols, we evaluate molecular structural formulas and reaction equations, from atom- and bond-level parsing to balancing, condition inference and product prediction.

Across these diverse domains, the benchmark contains more than 13,000 meticulously curated instances. Some datasets are curated and reannotated from existing resources, while others are newly constructed specifically for our symbolic objectives. The goal is not merely to measure domain knowledge, but to test whether models can preserve symbolic fidelity as cognitive demands gradually increase throughout the levels. Additional dataset construction details are given in Supplementary Sec.~\ref{Sec:SupplementaryDatasetConstruction}.

\subsection*{Models, prompting and human baseline}

We evaluate a mixture of closed-source and open-source MLLMs, including GPT-4o, Claude-sonnet-4, Gemini-2.5-pro, o3, Qwen-max, Qwen2.5-VL, InternVL3-8B, Deepseek-vl2-tiny and LLaMA3-LLaVA-Next-8B, representing the current state-of-the-art in multimodal intelligence. Each model is queried with task-specific prompts aligned to the required output format of the benchmark. We keep prompts as direct as possible so that performance reflects genuine symbolic understanding rather than the effects of prompt engineering.

To contextualize the machine results, we also build a robust human baseline on a stratified sample of 1,000 instances from the full benchmark. Five highly educated bilingual annotators complete the tasks using the same prompts as the MLLMs, providing a gold-standard reference for our comparative analysis. This comparison is used to examine whether model behavior follows the same difficulty trajectory as human symbolic cognition when facing increased complexity. Further implementation details, together with the extended human-baseline analysis, are reported in Supplementary Secs.~\ref{Sec:SupplementaryExperiments} and~\ref{Sec:SupplementaryHumanBaseline}.

\subsection*{Evaluation}

Evaluation is tailored to the intrinsic symbolic properties of each domain. For language symbols, we report precision, recall and F1 for error detection, along with exact match and edit distance for correction. For cultural symbols, we use exact-match style metrics at the word and sentence levels, character-overlap measures for partial idiom recovery, and an additional semantic-similarity score to capture cases in which outputs are conceptually related but not lexically identical to the ground truth. For mathematical, physical and chemical symbols, we use accuracy on each task because the target outputs are structurally constrained and logically discrete.

These metrics are intended to separate partial recognition, approximate semantic recovery and exact symbolic correctness within discrete spaces. In the symbolic domains studied here, this distinction is essential: a response can be fluent, plausible and still fundamentally wrong in the precise way that matters most scientifically and structurally. The full metric definitions are provided in Supplementary Sec.~\ref{Sec:SupplementaryExperiments}.

\section*{Data availability}
Datasets introduced in this study are freely available at \href{https://huggingface.co/datasets/Eternity-gaga/SymbolBench}{https://huggingface.co/datasets/Eternity-gaga/SymbolBench}.

\section*{Code availability}
The source code of this study is publicly available on Github at \href{https://github.com/THUKElab/SymbolBench}{https://github.com/THUKElab/SymbolBench}.

\bibliography{sample}

\clearpage
\setcounter{section}{0}
\setcounter{subsection}{0}
\setcounter{subsubsection}{0}
\renewcommand{\thesection}{\Alph{section}}
\renewcommand{\thesubsection}{\thesection.\arabic{subsection}}
\renewcommand{\thesubsubsection}{\thesubsection.\arabic{subsubsection}}
\startcontents[supp]
\begingroup
\hypersetup{linkcolor=black}
\titlecontents{section}[0pt]
  {\large\rmfamily\addvspace{0.9em}}
  {\makebox[2.2em][l]{\thecontentslabel}}
  {}
  {\hfill\thecontentspage}
\titlecontents{subsection}[1.6em]
  {\normalsize\rmfamily\addvspace{0.45em}}
  {\makebox[3.1em][l]{\thecontentslabel}}
  {}
  {\ \titlerule*[0.7pc]{.}\ \thecontentspage}
\titlecontents{subsubsection}[4.2em]
  {\normalsize\rmfamily\addvspace{0.2em}}
  {\makebox[3.5em][l]{\thecontentslabel}}
  {}
  {\ \titlerule*[0.7pc]{.}\ \thecontentspage}
\section*{Contents}
\printcontents[supp]{}{1}{\setcounter{tocdepth}{3}}
\endgroup

\clearpage
\section*{Supplementary}
\addcontentsline{toc}{section}{Supplementary}

\section{Extended Introduction}\label{Sec:SupplementaryIntro}

% Since the advent of the Large Language Models era in Artificial Intelligence, Multimodal Large Language Models (MLLMs) have consistently remained one of the most cutting-edge and prominent research topics~\cite{caffagni2024revolution, song2025bridge, zhang2024mm}. Beyond textual media, researchers have explored how artificial intelligence can see and perceive the world, and truly enhance its understanding of it. Recently, with the rise of Embodied Intelligence, the value of MLLMs research has garnered increasing attention from a growing number of researchers~\cite{xu2025embodied, turgunbaev2025perception, szot2025multimodal}. The fundamental reason behind this trend is that the cognitive paradigm represented by MLLM is closer to human intelligence in understanding the world, and is an essential step towards achieving Artificial General Intelligence (AGI)~\cite{du2025human, fei2024multimodal, shen2025zoomeye}. Consequently, referencing human behavior and performance in the real world is crucial to enhancing the capabilities of MLLMs, which is also \textbf{the core objective of us to promote MLLMs to think in a manner more closely aligned with human thinking, and to achieve more human-like general artificial intelligence}.

Since the advent of the Large Language Models era in Artificial Intelligence, Multimodal Large Language Models (MLLMs) have consistently remained one of the most cutting-edge and prominent research topics~\cite{caffagni2024revolution, song2025bridge, zhang2024mm}. Beyond textual media, MLLMs aim to endow artificial systems with the ability to see, perceive, and reason about the physical world, thereby moving toward a more comprehensive form of intelligence. In recent years, the rapid rise of Embodied Intelligence has further elevated the significance of MLLMs~\cite{xu2025embodied, turgunbaev2025perception, szot2025multimodal}. The fundamental reason behind this trend is that the cognitive paradigm represented by MLLM is closer to human intelligence in understanding the world and represents an essential step towards achieving Artificial General Intelligence (AGI)~\cite{du2025human, fei2024multimodal, shen2025zoomeye}. Consequently, understanding and emulating fundamental mechanisms of human cognition in the real world has become crucial for advancing MLLMs, which is also \textbf{our core objective: to promote MLLMs to reason in a manner more aligned with human thought, thereby fostering more human-like artificial intelligence.}

% Symbols have been the indispensable cornerstone of human cognition and the evolution of intelligence since the dawn of the human species~\cite{perlovsky2007symbols}. From prehistoric rock paintings that encoded survival knowledge to the structured syntax of natural languages, the invention and manipulation of symbolic systems have enabled humans to transcend the limitations of individual sensory experience, facilitating knowledge accumulation and abstract reasoning~\cite{taniguchi2018symbol, bickerton2003symbol, miyagawa2018cross}. As cognitive science represented by semiotic theories has elucidated, human intelligence is inherently symbolic~\cite{rapoport1955role, ahn2017signals}. Crucially, as illustrated in Figure~\ref{fig:continuous_vs_discrete_comparison} (b), the human visual system exhibits a specialized sensitivity to visual symbols, allowing us to efficiently perceive the world by encoding perceptual inputs into \textbf{discrete symbolic representations} (e.g., characters, gestures, or visual motifs). This symbolic capacity is not merely a tool for communication but the very fabric of human thinking—underpinning our ability to conceptualize abstract ideas, solve complex problems, and construct shared realities.

Symbols have been the indispensable cornerstone of human cognition and the evolution of intelligence since the dawn of the human species~\cite{perlovsky2007symbols}. From prehistoric cave paintings encoding survival knowledge to the structured syntax of natural language, symbolic systems allow humans to transcend the limits of individual sensory experience and accumulate abstract knowledge across generations~\cite{taniguchi2018symbol, bickerton2003symbol, miyagawa2018cross}. As cognitive science represented by semiotic theories has elucidated, human intelligence is inherently symbolic, relying on the creation, manipulation, and communication of symbols to support reasoning and collective understanding~\cite{rapoport1955role, ahn2017signals}. Crucially, as illustrated in Figure~\ref{fig:continuous_vs_discrete_comparison}(b), the human visual system exhibits unique sensitivity to visual symbols, enabling perceptual inputs to be rapidly encoded into discrete symbolic representations such as characters, gestures, and schematic patterns. This symbolic capacity is not merely a tool for communication but constitutes the very fabric of human thinking—underpinning our ability to conceptualize abstract ideas, solve complex problems, and construct shared realities.

% Notably, this intrinsic alignment between human cognition and discrete symbolism stands in stark contrast to the mainstream paradigm of MLLMs. As depicted in Figure~\ref{fig:continuous_vs_discrete_comparison} (a), there is a fundamental distinction between continuous and discrete semantic spaces. Mainstream MLLMs are predominantly designed to process continuous visual information (e.g., a photograph of a sunlit grassland) that maps to coherent and unified semantic narratives, such as widely applied tasks like image captioning~\cite{abdulgalil2025next}, VQA~\cite{kuang2025natural}, and grounding~\cite{xiao2024towards}. \textbf{In sharp contrast, the discrete semantic space we focus on in this work is defined by input images composed of semantically independent symbols, where each symbol carries its own distinct semantic value}. For example, in a mathematical equation or a chemical diagram, the model's task is to interpret the meaning of different symbols as well as their potential combinatorial relationships.

However, this intrinsic alignment between human cognition and discrete symbols stands in stark contrast to the dominant training paradigms of current MLLMs. As depicted in Figure~\ref{fig:continuous_vs_discrete_comparison}(a), continuous semantic spaces and discrete semantic spaces differ fundamentally in their representational structure. Most existing MLLMs are optimized to process continuous visual signals—such as natural images of scenes—mapping them to coherent semantic narratives through tasks such as image captioning~\cite{abdulgalil2025next}, Visual Question Answering~\cite{kuang2025natural}, and visual grounding~\cite{xiao2024towards}. \textbf{In contrast, discrete semantic spaces consist of semantically independent symbolic units, where meaning emerges from precise identification and combinatorial relations among symbols.} 
For example, in symbolic images such as mathematical equations or chemical structure diagrams, correct interpretation requires the model to recognize each symbol as a discrete semantic entity and reason over its structured composition. This representational gap poses a fundamental challenge for current MLLMs.

% Across prior research on MLLMs, investigations into discrete symbols remain notably scarce. It is precisely the gap in the mechanisms by which MLLMs handle discrete symbolic visual information that motivates the present work. We aim to decode how these models recognize, interpret, and integrate such discrete symbolic components, comparing their behavior to the visual symbol comprehension mechanism in the human brain shown in Figure~\ref{fig:continuous_vs_discrete_comparison} (c). This mechanism highlights a ``Language Dominance Over Visual Grounding'' process, suggesting a complex neural interplay where linguistic processing guides visual interpretation. By investigating the visual semiotic behaviors of MLLMs in the discrete semantic space we define, we believe that our research not only fills a key gap in current MLLM research but also lays the foundation for developing intelligent systems that are more interpretable and closer to human symbolic cognition.

Across prior research on MLLMs, investigations into discrete symbols remain notably scarce. It is precisely this gap in understanding the mechanisms by which MLLMs process discrete symbolic visual information that motivates the present work. To bridge this, we draw inspiration from hierarchical neural and cognitive pathways underlying human symbolic processing, as illustrated in Figure~\ref{fig:continuous_vs_discrete_comparison}(c). Cognitive neuroscience suggests that humans do not process symbols in a flat, end-to-end manner. Instead, symbolic cognition unfolds along a progressive pipeline, which begins with \emph{recognition and perception}, where raw visual inputs are parsed into recognizable symbolic units, followed by \emph{combination and reasoning}, where symbols are syntactically combined to infer compositional meaning. At the highest level, \emph{association and critical thinking} monitor logical consistency, detect errors, and resolve ambiguities. We argue that true mastery of discrete semantic spaces by MLLMs requires competence across the entire cognitive spectrum, rather than relying solely on statistical correlations or linguistic priors. Thus, by investigating the visual semiotic behaviors of MLLMs in the discrete semantic space we define, we believe that our research not only fills a key gap in current MLLM research but also lays the foundation for developing intelligent systems that are more interpretable and more closely aligned with human symbolic cognition.

To make this benchmark framing more concrete, Figure~\ref{fig:data_Introduction} in Supplementary Sec.~\ref{Sec:SupplementaryIntro} visualizes the three-level hierarchy together with representative task examples across the five domains.

\begin{figure}
    \centering
    \includegraphics[width=0.85\linewidth]{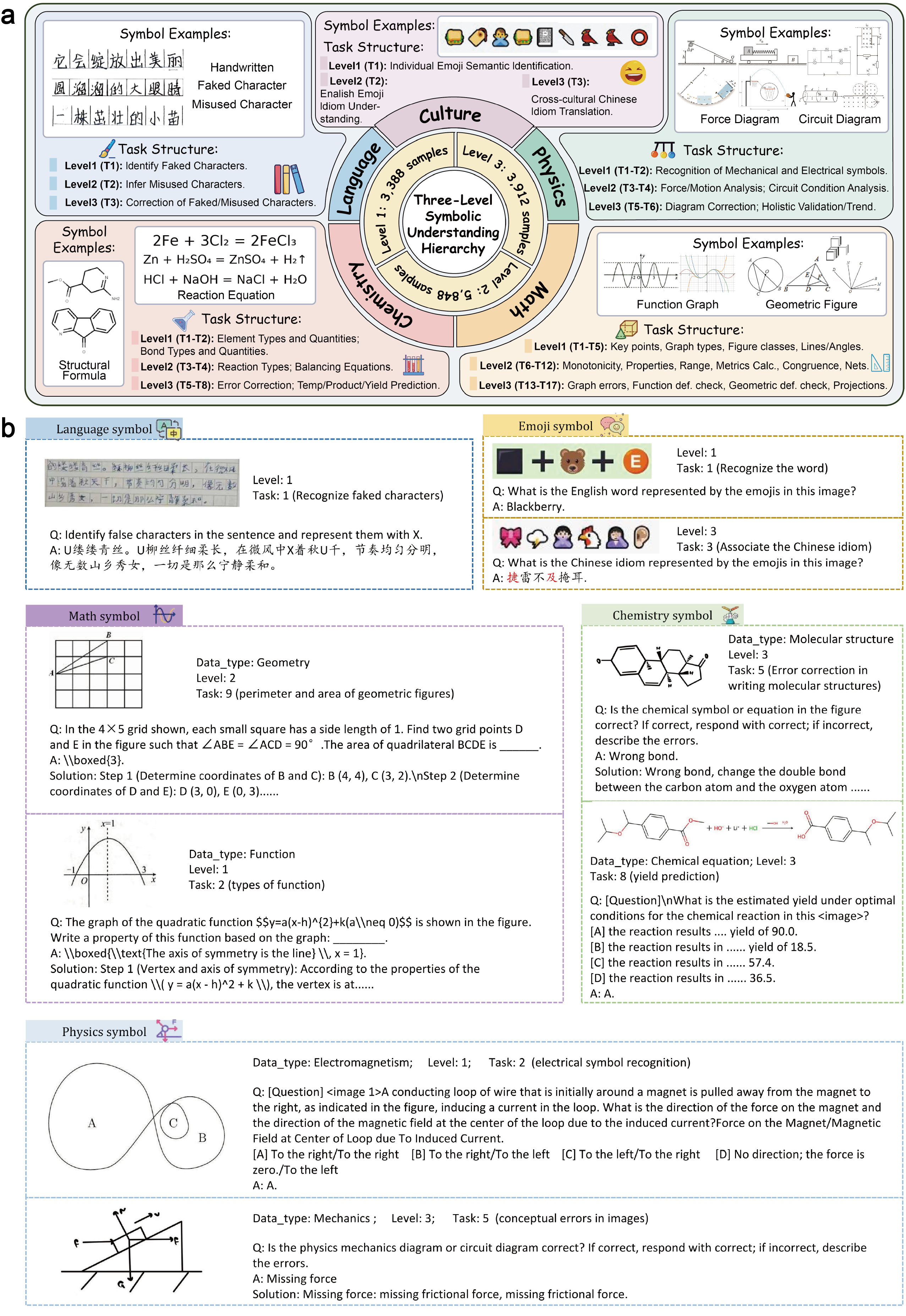}
    \caption{Overview of the benchmark task design framework and illustrative examples. (a) The instantiation of the three-level symbolic understanding hierarchy across five distinct domains: Language, Cultural, Mathematical, Physical, and Chemical symbols. (b) Representative examples of the tasks designed for our benchmark.}
    \label{fig:data_Introduction}
\end{figure}

Operationalizing the above hierarchical cognitive mode, \textbf{we introduce a comprehensive benchmark designed to systematically evaluate the visual symbolic capabilities of MLLMs in discrete semantic spaces}. Unlike prior benchmarks that primarily emphasize natural image understanding or open-ended visual question answering, our framework focuses on structured, abstract, and highly symbolic visual representations that explicitly encode meaning. Importantly, these symbols cannot be trivially recognized through low-level visual capabilities (e.g., OCR). Drawing on insights from human visual neuroscience and cognitive psychology, our framework aligns with these cognitive stages and spans five distinct symbolic domains that mirror the evolution of human knowledge: Language (e.g., handwritten and faked Chinese characters), Culture (e.g., emojis and idioms), Mathematics (e.g., function graphs and geometry), Physics (e.g., circuit diagrams and mechanics), and Chemistry (e.g., molecular structures). To rigorously assess the depth of symbolic understanding, we structure our benchmark across a three-level cognitive hierarchy inspired by Bloom's taxonomy and semiotic theory~\cite{bloom1956taxonomy, eco1979theory}, as shown in Figure \ref{fig:data_Introduction}. The first level assesses \emph{recognition and perception}, evaluating whether models can reliably identify basic symbolic primitives such as handwritten characters, schematic elements in function plots, molecular components, or physical diagram symbols. The second level targets \emph{compositional reasoning}, where symbols must be integrated and interpreted according to domain knowledge, such as inferring functional properties from graphs or analyzing force interactions in mechanics. The third level probes \emph{associative and critical cognition}, requiring models to detect inconsistencies, correct malformed symbols, and interpret non-literal or context-dependent meanings. We collect large-scale raw data from existing public datasets and a large volume of handwritten symbol data from human annotation experts. Based on a strong base of MLLMs, we perform domain classification annotation and question generation, resulting in 38 different sub-tasks and 13k question-image-answer pairs. After strict automated quality verification and manual validation, we obtain a complete evaluation dataset with a corresponding evaluation suite.

%%%%综合图分析

\begin{figure}
    \centering
    \includegraphics[width=0.80\linewidth]{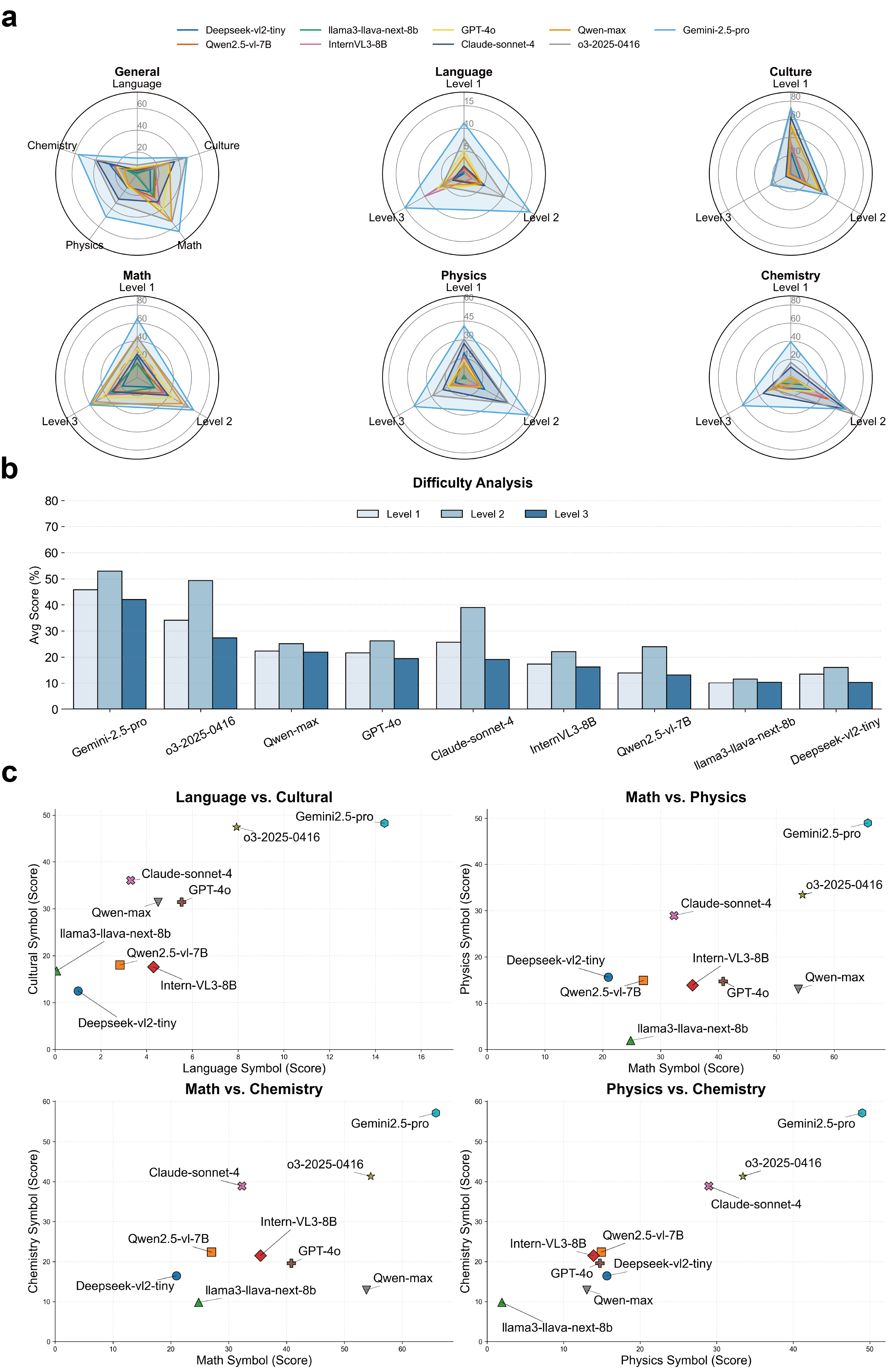}
    \caption{Cross-domain summary of benchmark results. (a) Radar charts illustrate the fine-grained performance of models across General, Language, Culture, Math, Physics and Chemistry domains. (b) Global performance aggregated by difficulty, with accuracy averaged across five symbolic domains. (c) Scatter plots exploring interrelationships between domains.}
    \label{fig:radar}
\end{figure}

We analyze domain performance, cognitive difficulty, and inter-domain correlations (Figure~\ref{fig:radar}). Most models exhibit unbalanced symbolic understanding, with proprietary models showing broader coverage across all domains than open-source counterparts. A critical finding is that \textbf{language symbols represent the most challenging} domain for all tested MLLMs. In contrast, models perform significantly better on natural science symbols, particularly in mathematics and chemistry, suggesting current architectures are more proficient at processing structured molecular and mathematical notations than identifying nuanced anomalies in linguistic characters.
We investigate performance across a three-level hierarchy comprising Level 1 (perception), Level 2 (reasoning), and Level 3 (critical thinking). Figure~\ref{fig:radar} shows a non-linear pattern where average Level 2 scores are frequently higher than or comparable to Level 1 across many models. This counterintuitive ``recognition-reasoning inversion'' suggests that \textbf{MLLMs rely on robust linguistic and structural priors to infer compositional meanings even when fine-grained visual perception of individual symbols is imperfect}. To explore mutual influences, we analyze the correlation between social science symbols (language and culture) and natural science symbols (mathematics, physics, and chemistry). A strong positive correlation exists within the natural sciences; models excelling in mathematics typically demonstrate superior performance in other formalized, rule-based scientific fields. Conversely, the relationship between language and cultural symbols appears more fragmented. While top-tier models lead in both areas, others exhibit specialized capabilities in specific pockets. This divergence indicates that cultural understanding requires a distinct set of semantic knowledge that does not fully overlap with pure linguistic parsing, reflecting the unique difficulty of interpreting non-formalized symbols in discrete spaces.

These extensive evaluations across state-of-the-art MLLMs of varying scales reveal several findings. We observe a counterintuitive \emph{recognition--reasoning inversion}: models often perform better on higher-level reasoning tasks than on foundational perceptual recognition tasks. This suggests that current MLLMs frequently bypass robust visual symbol grounding, instead relying on linguistic priors or memorized patterns. In several domains, particularly chemistry and mathematics, models exhibit procedural imitation, successfully reproducing solution patterns without a genuine understanding of the underlying symbols. Moreover, strong language reasoning capabilities can partially compensate for deficient visual perception, thereby masking perceptual failures through contextual inference. Finally, no single model demonstrates consistent performance across symbolic domains, indicating that current strengths remain largely domain-dependent and data-driven rather than systematic.

We provide the broader benchmark-level summary and the inter-domain comparison in Figure~\ref{fig:radar} in Supplementary Sec.~\ref{Sec:SupplementaryResults}, where the imbalance between domains and the recognition-reasoning inversion can be seen more globally.

Taken together, these findings expose a fundamental cognitive mismatch in contemporary MLLMs and underscore the necessity for benchmarks that explicitly disentangle perception, reasoning, and critical symbolic understanding. Thinking further, the observed limitations are rooted in the fundamental divergence between the \textit{continuous representational bias} of current visual encoders (e.g., CLIP-based ViTs) and the \textit{compositional rigor} required by discrete semiotics. While MLLMs excel at directing visual signals to high-level linguistic concepts, this process often bypasses the intermediate structural parsing essential for symbolic semiosis. Unlike natural images, where semantic ``gist'' is preserved through spatial redundancy, discrete symbols exhibit high information density where a single stroke deletion (e.g., in faked characters or chemical bonds) triggers a total semantic shift. This ``Cognitive Mismatch'' suggests that current architectures lack a \textit{structural bottleneck} capable of preserving the topological integrity of symbols, representing a foundational barrier to achieving Human-aligned Artificial General Intelligence. The main contributions of this paper are summarized as follows:
\begin{itemize}
    \item \textbf{A symbolic perspective on MLLM evaluation}: We introduce the first framework dedicated to assessing MLLMs in discrete semantic spaces, shifting the focus from continuous perception to structured symbolic interpretation.
    \item \textbf{A hierarchical, multi-domain benchmark}: We construct a large-scale, high-quality benchmark spanning five symbolic domains and three cognitive levels, enabling fine-grained diagnosis of model capabilities.
    \item \textbf{Insights into fundamental cognitive limitations}: Our comprehensive analysis reveals systematic deficiencies in fine-grained visual symbol grounding and highlights the persistent reliance of current MLLMs on heuristic linguistic shortcuts, offering valuable new directions for advancing embodied and symbolic intelligence.
\end{itemize}

\subsection{Related Work}\label{Sec:SupplementaryRelatedWork}
\subsubsection{General Benchmarks}

The evaluation landscape for Multimodal Large Language Models (MLLMs) has evolved into a multifaceted ecosystem, transitioning from foundational general capabilities to complex cognitive and interactive intelligence. Comprehensive benchmarks \cite{hiippala2021ai2d, qin-etal-2024-infobench, tang2025mtvqa, wang2024mmlu} typically employ meticulously crafted multiple-choice or open-ended questions to assess dimensions such as vision–language understanding, world knowledge, and multi-step reasoning. Complementing these, fine-grained perception benchmarks \cite{fu2024blink, pmlr-v235-ying24a, fu2025mme, meng2025mmiu, yue2024mmmu, liu2024mmbench, jiang2024mantis, chen2025megabench, wang2025muirbench, wang2025mfcbench} evaluate not only object and scene recognition but also the ability to infer semantic relations, action intentions, and underlying logical connections. Visual grounding tasks \cite{kazemzadeh2014referitgame, mao2016generation, li2022elevater} further assess precise localization of target regions based on textual descriptions, while hallucination and safety suites \cite{li-etal-2023-evaluating
, guan2024hallusionbench, lin2022truthfulqa, li2023halueval} aim to quantify faithfulness and mitigate ungrounded generation. To probe advanced intelligence, researchers have introduced challenges in abstract reasoning \cite{zeng2024mrben, rein2024gpqa, hao2025can, mialon2024gaia}, code synthesis \cite{10.1145/3520312.3534862, uniyal-etal-2024-one, liu2023your, jain2025livecodebench}, and long-context processing \cite{shaham2023zeroscrolls, jin2025longcontext, zhang-etal-2024-bench, wang-etal-2025-multimodal}. More recently, the frontier has shifted toward dynamic and interactive capabilities, incorporating video understanding benchmarks \cite{li2024mvbench, fang2024mmbench, zhou2025mlvu, wu2024longvideobench, wang2025lvbench, liu-etal-2024-tempcompass, fu2025video} to assess the comprehension of temporal information, alongside autonomous decision-making in agentic GUI environments \cite{cheng2024seeclick, qiu2025can, rawles2025androidworld, li2024on, li2025screenspotpro, xie2025scaling}. Despite this expansive breadth, existing benchmarks primarily focus on naturalistic scenes, often overlooking the structured, abstract symbolic systems that underpin human civilization.

\subsubsection{Symbolic Benchmarks}
Semiotics is the study of how symbols carry and convey meaning \cite{de1985linguistic, key2017analysis, valdez2024semiotics, danesi2024ai}. In semiotic theory, a sign is not the object itself but consists of two components: the signifier, referring to the form of the symbol, and the signified, denoting the concept or meaning it represents \cite{hatt2025semiotics, iskanderova2024semiotics}. In the social sciences, evaluation has moved from modern OCR \cite{fu2025ocrbench, yu2025benchmarkingvisionlanguagemodelschinese, li-etal-2024-towards-real} to deciphering complex ancient scripts like Oracle Bone Inscriptions \cite{chen2025obibench}, Egyptian Hieroglyphs \cite{fuentes2025recognition}, and Ancient Yi \cite{liu2024ancient}. Cultural assessment has transitioned from emoji-based sentiment analysis \cite{liu2021improving, zheng2025irony} to sophisticated semantic generation \cite{kuang2025express}, geo-diverse VQA \cite{nayak-etal-2024-benchmarking}, and interactive art critique \cite{zheng2025artmentor}. Recent work like WildScore \cite{mundada-etal-2025-wildscore} further probes the structural reasoning of musical scores.
In the natural sciences, mathematical benchmarks have evolved from static formula parsing \cite{lu2024mathvista, wang2024measuring, zhang2024mathverse} to dynamic program-based synthesis \cite{zou2025dynamath} and fine-grained error correction \cite{nath-etal-2025-vision}. Physics evaluation now encompasses circuit analysis \cite{shi2025amsbench}, grounded reasoning \cite{shen2025phyx, dai-etal-2025-physicsarena}, and university-level problem solving that resists textual shortcuts \cite{wang2025physunibench, xiang2025seephys}. Similarly, chemistry suites focus on table structure extraction \cite{zhai2021chemtables}, versatile real-world scenarios \cite{huang2024chemeval}, and molecular elucidation via spectral data \cite{guo2024can}. Despite this proliferation of datasets, most existing work evaluates reasoning in a terminal fashion, focusing on the final answer. Our work addresses this critical gap by mirroring a human-like cognitive progression, providing a diagnostic hierarchy from discrete symbol identification to compositional logic and emergent semantic inference across five foundational domains.

\subsubsection{Architectures for Symbolic Domain}

In recent years, Multimodal Large Language Models (MLLMs) \cite{10.1093/nsr/nwae403, 10.1145/3695053.3731410, mckinzie2024mm1, wu2024visionllm} have developed rapidly. Early models such as CLIP \cite{radford2021learning} and ALIGN \cite{jia2021scaling} laid the foundation through large-scale image-text contrastive learning, followed by BLIP-2 \cite{li2023blip} and the LLaVA series \cite{liu2023visual, lin-etal-2024-video, liu2024improved, li2025llavaonevision}, which further advanced the performance of MLLMs in image understanding, visual question answering, and open-domain dialogue. More recent studies have shifted their focus to architectural efficiency and native multimodal integration. Key innovations include M-ROPE for temporal-spatial alignment \cite{bai2025qwen2, yang2025qwen3}, Cascade Reinforcement Learning for scientific reasoning \cite{wang2025internvl3}, and unified understanding-generation architectures \cite{yang2025mmada, fu2025vita}. To handle symbolic data, specialized paradigms have emerged. For text-intensive perception, models utilize window attention \cite{liu2024textmonkey, yu2024texthawk} and layout-compressed query embeddings \cite{nacson2025docvlm}. In cultural reasoning, approaches like NotaGPT \cite{tang-etal-2025-nota} align 2D symbols with text sequences, while ArtCoT \cite{jiang2025multimodal} and ArtSeek \cite{fanelli2025artseek} apply evidence-based Chain-of-Thought (CoT) to minimize hallucinations. For scientific symbols, methodologies emphasize structural rigor through geometric element alignment \cite{gao2025gllava, shi-etal-2024-math}, symbolic verification mechanisms \cite{pan2025enhancing, shi2025multimodal}, and external simulator integration \cite{zhu2025maps, wiesner2025towards}. Molecular modeling has similarly shifted from string translation \cite{edwards-etal-2022-translation} to discrete token-level fusion \cite{tan2025chemmllm} and high-resolution image compression \cite{zhao2024chemdfm, li2025chemvlm}. However, these approaches remain fragmented across specific domains; our benchmark provides a unified framework to drive the development of models capable of integrated, multi-level symbolic reasoning.

\section{Extended Results}\label{Sec:SupplementaryResults}
\subsection{Language Symbols}\label{Sec:SupplementaryLanguage}

\subsubsection{Weak Recognition Ability for Faked Characters}
% In the Faked Character Detection task, as shown in Table 1, the overall performance of most models was extremely poor,
In task 1 (faked character detection), the overall performance of most models was extremely poor, with F1 scores universally below 2. Only Gemini-2.5-pro, o3, and GPT-4o achieved slightly higher results. In contrast, most open-source models performed particularly poorly; LLaMA3-llava-next-8b, for instance, often defaulted to outputting a templated ``cannot analyze the image'' prompt without attempting to identify or correct the faked characters. This phenomenon reflects a deficiency in the underlying visual encoding capability of current MLLMs for sparse character structures, especially in abnormal cases involving missing strokes or faint handwriting, where they fail to establish a stable character-space representation.

Qualitative analysis reveals two primary failure modes. First, some models did not recognize the faked characters as errors but instead automatically replaced them with the most similar legal glyphs in their outputs, as seen with the characters ``\cn{推}'' (push) and ``\cn{荐}'' (recommend) in Case 1 of Figure~\ref{fig:language_results}. This demonstrates a typical forced normalization behavior, whereby the model repairs anomalous strokes at the visual stage into a symbol that can be mapped to its linguistic vocabulary, thereby erasing the anomalous features at the perceptual level. Second, while some models could follow the instruction to mark errors, they lacked precise symbol discrimination ability, often mistaking normal characters for anomalous ones and thus producing incorrect localizations and redundant annotations. For example, in Case~\ref{case:langL1_2} in Supplementary Sec.~\ref{Sec:SupplementaryCases}, a model misidentified the correct character ``\cn{违}'' (violate) as a faked character. This behavior shows an inability to distinguish between poorly written yet correct characters and structurally incorrect faked characters, leading to an imbalanced detection result characterized by a high X\_count\_pred but a low F1 score.

\subsubsection{Insufficient Recognition of Misused Characters}

In task 2 (contextual character misuse identification), models must not only recognize individual characters but also integrate visual recognition with contextual semantics to identify word or sentence-level misspellings. The results show that while Gemini-2.5-pro and o3 maintained their lead, the F1 scores of GPT-4o and the Qwen series were clustered at the low level of around 5. This indicates that current models still struggle to integrate character recognition with syntax and semantics into a coherent comprehension structure when faced with semantically dependent symbolic tasks. Notably, both Deepseek-vl2-tiny and LLaMA3-llava-next-8b scored 0, but for different reasons. LLaMA3-llava-next-8b failed to correctly perceive the text in the image, often hallucinating entirely irrelevant characters and sentences and attempting to analyze this fabricated content. In contrast, Deepseek-vl2-tiny, as shown in Case~\ref{case:langL2_1} in Supplementary Sec.~\ref{Sec:SupplementaryCases}, could accurately recognize the entire sentence via OCR but treated it as an indivisible whole, inserting it directly into the JSON output, thus demonstrating a lack of a character-level decomposition and comparison mechanism.

Interestingly, the 7B parameter Qwen2.5-vl achieved a score comparable to the much larger GPT-4o and close to its sibling Qwen-max. This suggests that in such highly specific symbolic tasks, performance is not solely dependent on model scale but may be influenced by the training data composition. The inclusion of fine-grained textual and handwritten document data during the Qwen series' training might explain its relative advantage on this task.

From a qualitative perspective, models exhibit shortcomings when required to combine semantic information to judge errors. In Case 2 of Figure~\ref{fig:language_results}, when the task involved distinguishing between the homophones ``\cn{的}'' (of) and ``\cn{地}'' (adverbial particle), models often visually identified ``\cn{的}'' (of) as a legitimate character but could not determine whether its grammatical position was correct based on the context. In practice, ``\cn{的}'' (of) typically precedes nouns, while ``\cn{地}'' (adverbial particle) precedes verbs; however, the models failed to understand this grammatical dependency and instead randomly generated unrelated replacement characters (e.g., incorrectly swapping ``\cn{室}'' (room) for ``\cn{试}'' (try)).

\subsubsection{Lack of a Mechanism for Maintaining Semantic Consistency}

In task 3 (visual-semantic character correction), under the exact match metric, only Gemini-2.5-pro and InternVL3-8B achieved double-digit scores, whereas LLaMA3-llava-next-8b scored zero, indicating it never once successfully completed a correction. The edit distance metric reveals the reason for this performance disparity. Gemini-2.5-pro and InternVL3-8B had a very low edit distance, suggesting that while their corrections were not perfectly accurate, they were directionally relevant, generating semantically similar characters. Some of their outputs even exhibited an ``analysis-correction'' chain-of-thought process. This indicates they have established a tighter synergistic mechanism between visual perception and semantic reasoning, allowing the models not only to ``see'' the error but also to understand and correct it accordingly.

In contrast, the high edit distance of models like Claude-sonnet-4, Deepseek-vl2-tiny, and Qwen2.5-vl indicates that their outputs deviated severely from the correct answers at both the character and semantic levels. These models often lack stable capabilities for error localization and controlled correction. For example, in Case 3 of Figure~\ref{fig:language_results}, a model misidentified an incorrect character as the visually similar character ``\cn{烧}'' (burn), leading it to generate the completely unrelated phrase ``\cn{烧烤板}'' (grill plate). This is a case where erroneous visual perception dominated semantic generation, leading to severe semantic drift.

The most extreme case was LLaMA3-llava-next-8b, whose edit distance reached an astonishing 179.7. This was not merely a failure of correction; rather, the model, unable to recognize the image, experienced severe hallucinations, generating a large volume of irrelevant text. It even repeatedly used templated excuses in its output, such as ``due to the low image quality,'' to directly admit its failure to recognize valid characters. Drawing on the results from Level 1 (perception and recognition) and Level 2 (combination and reasoning), it can be inferred that most models failed to correctly perform error localization in the earlier stages, which naturally precluded them from making targeted corrections in Level 3 (association and critical thinking).

% \begin{CJK}{UTF8}{gbsn}
% \begin{tcolorbox}[
%     colback=natureblue!30!white,
%     colframe=natureblueframe,
%     boxrule=1pt,
%     rounded corners,
%     drop shadow={opacity=0.4},
%     % equal height group=LangL2,% 新增：将两框高度统一
%     valign=top,
%     boxsep=1mm, left=1mm, right=1mm, top=1mm, bottom=1mm,
%     sidebyside,
%     sidebyside align=top seam,
% ]
% % =================== 左栏内容 (问题描述与图片) ===================
% \noindent\textcolor{naturedarkblue}{\small\bfseries Question}
% \smallskip
% % deepseek：1192
% Task 3: Sentence correction. Output the fully corrected sentence according to the image content.
% \begin{center}
%     % 图片宽度设为 \linewidth，使其自动适应左栏宽度
%     \includegraphics[width=0.5\linewidth]{case/language3_2.jpg}
% \end{center}
% % --- Analysis 放在左边底部 ---
% \noindent\textcolor{naturedarkblue}{\small\bfseries Analysis}
% \smallskip
% Incorrect visual perception dominates semantic generation, leading to severe semantic drift.
% \tcblower % <<< 分隔符：此命令之后的内容将出现在右栏
% % =================== 右栏内容 (答案对比与分析) ===================
% % 答案与模型输出对比
% \begin{answerbox}
%     Correct: 跷跷板也很好玩。
    
%     Trans.: The seesaw is also a lot of fun.
% \end{answerbox}
% \begin{modelbox}
%     Model: \textcolor{red}{烧烤}板也很好玩。
    
%     Trans.: The \textcolor{red}{barbecue} grill is also a lot of fun.
% \end{modelbox}
% \end{tcolorbox}
% \captionof{case}{Case of Language-Level 3-Correction of faked and misspelled characters.}
% \label{case:langL3_1}
% \end{CJK}

\subsection{Cultural Symbols}\label{Sec:SupplementaryCultural}
\subsubsection{Models perform well at the low level but still face challenges}

Compared to the two Chinese tasks, MLLMs performed better on English. As shown in Figure~\ref{fig:culture_exp} (a) and (b), GPT-4o achieved impressive F1 scores of 55.8 and 35.2 on the word-level and sentence-level tasks for English word and idiom recognition, respectively. This is likely because the model encountered more similar English text during training, making it more adept at reasoning with English words. In Case 4 of Figure~\ref{fig:culture_exp}, the model accurately identified pine on the left and apple on the right, successfully integrated their semantics, and inferred the compound word pineapple, demonstrating strong capabilities in semantic composition and association.

However, MLLMs consistently struggle with hallucinations when decoding emojis as visual codes. As shown in Case 5 of Figure~\ref{fig:culture_exp}, the model recognized the smiling face and lightbulb emojis but ignored the critical semantic constraint of the third ``prohibition'' symbol. It proceeded to hallucinate the positive idiom ``bright idea,'' which is semantically opposite to the correct answer, ``Not the brightest bulb.'' This reveals that once the models capture the linguistic meaning of an individual emoji, they tend to immediately retrieve related words or idioms from their internal knowledge while ignoring the crucial context provided by the surrounding emojis.

% Gemini有着更好的视觉能力（Claude的推理能力虽然强，但是往往在第一步的表情符号理解上就出错了），达到了最好的性能
\subsubsection{Models show limited performance in Chinese idiom tasks}

We evaluate four-character and multi-character idioms. MLLMs perform poorly: GPT-4o achieves accuracy scores of 3.3 and 5.0 on these tasks. However, even the strongest open source model Qwen2.5-VL lags behind GPT-4o; the differing strengths of the models remain apparent on our more challenging benchmark. In Case 6 of Figure~\ref{fig:culture_exp}, the model interpreted and reasoned about the first two emojis along the dimension of ``color,'' generating a four-character idiom that was unrelated to the latter two emojis. In reality, the key to this image lies in reasoning from object semantics and homophonic relationships—for example, recognizing that the ``bucket'' (\cn{桶}, tǒng) shown in the image is a homophone for the character meaning ``same'' (\cn{同}, tóng). Tasks of this nature require not only symbol recognition but also cross-modal divergent thinking and the ability to perform phonetic-semantic associative reasoning. Accuracy at the Chr-1 level is significantly higher, indicating that MLLMs can perform basic text translations corresponding to individual emojis, but they have limited visual intuitive semiosis capabilities to further infer the corresponding linguistic meanings based on the relevant emoji context, particularly for the reasoning.

\subsubsection{Semantic Similarity Analysis Reveals Random Guessing Patterns in Association and Critical Thinking} 

We further compute the semantic similarity between the responses and the ground truth, applying LLM to score from 1 to 5. As shown in Figure \ref{fig:culture_exp} (c), the average scores are low, with the English task significantly higher than those on the Chinese task. When carefully observing the distribution, we observe that 1)for the Chinese task, most of the scores are concentrated in 1 and 2, indicating the poor performance of MLLMs; 2)while for the English task, most of the scores are concentrated in 1 and 5, demonstrating that the MLLMs can either predict the answer correctly, or get irrelevant answers.

In the Chinese multi-character idiom task of Case 7 of Figure~\ref{fig:culture_exp}, the model successfully recognized the first two emojis, ``\cn{一}'' (one) and ``\cn{日}'' (day), accurately capturing the initial semantic cue. However, it failed to establish a logical connection among the subsequent symbols. Notably, although there were a total of 8 character elements in the image, the model ultimately output only a 4-character idiom. This indicates that when MLLMs are faced with long-sequence, composite symbol combinations, their visual attention tends to focus on the beginning of the input, while the subsequent content is either ignored or overridden by early semantic hypotheses.

\subsection{Mathematical Symbols}\label{Sec:SupplementaryMathematical}

% \begin{figure}
%     \centering
%     \includegraphics[width=0.8\linewidth]{figures/math_performance.pdf}
%     \caption{The performance of different models in Math symbols.}
%     \label{fig:enter-label}
% \end{figure}

The overall experimental results reveal significant hierarchical disparities and contradictions in model performance across various task tiers. These disparities not only reflect the inherent difficulty of the tasks but also expose a deeper imbalance in how multimodal models develop their visual perception and language reasoning capabilities. Notably, most models perform worse on basic recognition tasks than on complex reasoning tasks, contradicting intuitive cognitive expectations.

\subsubsection{Visual perception ability}

Firstly, the fundamental difference in task design is one of the core reasons for this phenomenon. Level 1 (perception and recognition) heavily relies on precise visual localization and symbolic semantic mapping, with extremely high data sensitivity—any pixel-level deviation can lead to failure—and very low error tolerance, where answering incorrectly means failing completely. For example, in task 2 (function type classification), the model must accurately identify curves like $ e^x $ as ``exponential functions.'' However, minor rendering differences such as jagged edges or axis compression often cause misclassification. Even advanced models like Qwen2.5 achieve only 37.3 points on this task. In contrast, Level 2 (combination and reasoning) and Level 3 (association and critical thinking) tasks focus more on multimodal logical reasoning and rule generalization, allowing partial compensation through linguistic logic. Take task 15 (geometric definition consistency check) as an example: even if the model fails to precisely locate corners or edges, it can still detect errors via logical rules such as ``the sum of internal angles in a triangle is not equal to 180,'' enabling Llama3 to score 66.7. These structural differences expose the visual limitations of models when facing Level 1 (perception and recognition) tasks.

In Case 8 of Figure~\ref{fig:math_exp}, the model was asked to identify the type of function shown in the graph. Although its pixel perception abilities were limited, the model was able to perform reasoning by elimination using its existing knowledge base of function definitions. It first analyzed the characteristics of exponential and logarithmic functions, then, based on the curve in the image exhibiting an upward-opening, U-shaped trajectory, it ultimately inferred that the graph corresponds to a quadratic function.

\subsubsection{Model performance comparison}

Secondly, the adaptability of models significantly influences their performance across different levels of tasks. Smaller models like Deepseek-vl2-tiny excel in clearly defined reasoning tasks (e.g., task6 function monotonicity reasoning), as they tend to bypass complex visual details and instead rely on language pattern matching. Larger models like GPT-4o demonstrate stronger adaptability in open-ended error detection tasks (Level 3), compensating for visual localization errors through joint vision-language representations. However, this divergence also leads to anomalous cases. For instance, Qwen2.5 performs exceptionally well on geometric computation tasks (e.g., area calculation), likely due to its use of built-in geometric knowledge templates such as the Pythagorean theorem to circumvent visual shortcomings. Yet, it scores only 8.3 points on task 3 (geometric shape classification), revealing a severe lack of visual generalization capability. This indicates that while large models possess strong language reasoning abilities, their foundational visual encoding capabilities still require improvement.

As shown in Case 9 of Figure~\ref{fig:math_exp}, this task could have been completed quickly through direct visual judgment. The function in the graph is a monotonically decreasing straight line, and the required range for x could have been obtained simply by reading the two intersection points with the coordinate axes. However, the model did not directly utilize this explicit visual information. Instead, it engaged in a three-step symbolic computation process: first identifying the coordinates of the intersection points with the x and y axes, then substituting these into the equation to solve for the slope 
k, and finally deriving the corresponding x range via algebraic operations. This indicates that even when provided with sufficient visual clues, the model still tends to bypass the image information and instead relies on linguistic logic to perform lengthy symbolic reasoning.

% \begin{CJK}{UTF8}{gbsn}
% \begin{tcolorbox}[
%     colback=natureblue!30!white,
%     colframe=natureblueframe,
%     boxrule=1pt,
%     rounded corners,
%     drop shadow={opacity=0.4},
%     % equal height group=LangL2,% 新增：将两框高度统一
%     valign=top,
%     boxsep=1mm, left=1mm, right=1mm, top=1mm, bottom=1mm,
%     sidebyside,
%     sidebyside align=top seam,
% ]
% % =================== 左栏内容 (问题描述与图片) ===================
% \noindent\textcolor{naturedarkblue}{\small\bfseries Question}
% \smallskip
% %Intern VL3 math 92
% The graph of the linear function y = kx + b is shown in the figure. Find the solution set of the inequality $0\le kx+b< 5$.
% \begin{center}
%     % 图片宽度设为 \linewidth，使其自动适应左栏宽度
%     \includegraphics[width=0.25\linewidth]{case/math2_1.png}
% \end{center}
% % --- Analysis 放在左边底部 ---
% \noindent\textcolor{naturedarkblue}{\small\bfseries Analysis}
% \smallskip
% Even with clear visual cues, the model still favors linguistic reasoning.
% \tcblower % <<< 分隔符：此命令之后的内容将出现在右栏
% % =================== 右栏内容 (答案对比与分析) ===================
% % 答案与模型输出对比
% \begin{answerbox}
%     (0,2]
% \end{answerbox}
% \begin{modelbox}
%      $x\le 2$ and $x> 0$. 1.Identify the points of intersection: substituting x = 2 and y = 0 into the equation y = kx + b. 2. \textcolor{red}{Write the equation of the line: $y = -\frac{5}{2}x + 5$}. 3. \textcolor{red}{Solve the inequality $0\le kx+b< 5$}. The solution set of the inequality $0\le kx+b< 5$ is the intersection of $x\le 2$ and $x> 0$.
% \end{modelbox}
% \end{tcolorbox}
% \captionof{case}{Case of Math-Level 2-Function.}
% \label{case:mathL2_1}
% \end{CJK}

\subsubsection{Fine-grained task analysis}

Additionally, the misalignment in mainstream pre-training objectives contributes to poor performance on Level 1 (perception and recognition) tasks. Most current multimodal models are trained to encourage “jump mapping” from vision to language concepts rather than fine-grained visual localization. For example, VQA and captioning tasks emphasize generating coherent language descriptions rather than identifying every detail in images. This training approach leaves models ill-equipped to handle tasks requiring precise visual analysis. GPT-4o scores only 26.3 on task-4 (Geometric Element Attribution) but achieves 77.7 on task-14 (Function Definition Validation), indicating its strength lies in semantic-based plausibility judgment rather than visual element enumeration. In Case 10 of Figure~\ref{fig:math_exp}, when asked to determine the number of zeros of the function, the model directly read the printed text ``$y=\log_{2}{(x)} $'' from the image. It thereby bypassed the need to identify the function type from the curve's shape, avoiding a more fine-grained visual task.

% \begin{CJK}{UTF8}{gbsn}
% \begin{tcolorbox}[
%     colback=natureblue!30!white,
%     colframe=natureblueframe,
%     boxrule=1pt,
%     rounded corners,
%     drop shadow={opacity=0.4},
%     % equal height group=LangL2,% 新增：将两框高度统一
%     valign=top,
%     boxsep=1mm, left=1mm, right=1mm, top=1mm, bottom=1mm,
%     sidebyside,
%     sidebyside align=top seam,
% ]
% % =================== 左栏内容 (问题描述与图片) ===================
% \noindent\textcolor{naturedarkblue}{\small\bfseries Question}
% \smallskip
% %Claude math 99
% How many zeros does this function have?
% \begin{center}
%     % 图片宽度设为 \linewidth，使其自动适应左栏宽度
%     \includegraphics[width=0.25\linewidth]{case/math1_2.jpg}
% \end{center}
% % --- Analysis 放在左边底部 ---
% \noindent\textcolor{naturedarkblue}{\small\bfseries Analysis}
% \smallskip
% Directly reads the text ``$y=\log_{2}{(x)}$'' from the image, avoiding more fine-grained visual tasks.
% \tcblower % <<< 分隔符：此命令之后的内容将出现在右栏
% % =================== 右栏内容 (答案对比与分析) ===================
% % 答案与模型输出对比
% \begin{answerbox}
%     1
% \end{answerbox}
% \begin{modelbox}
%      Let me examine the graph step by step: 1) Identify the function: \textcolor{red}{The graph shows $y=\log_{2}{(x)}$}, which is a logarithmic function with base 2.
% \end{modelbox}
% \end{tcolorbox}
% \captionof{case}{Case of Math-Level 1-Function.}
% \label{case:mathL3_1}
% \end{CJK}

Ultimately, we arrive at an important conclusion: the weakness of current models on Level 1 (perception and recognition) tasks is not entirely due to insufficient capability but rather to their selective reliance on language reasoning mechanisms, which suppress the development of the visual modality. This phenomenon reveals a key contradiction in multimodal learning—when the language modality becomes too dominant, the growth potential of the visual modality is compressed, preventing the model from truly “understanding the world visually”. This further confirms the necessity of constructing high-quality multimodal symbolic datasets. Only through such datasets can we break the current “language-dominant, vision-passive” paradigm and push multimodal models toward higher-level cognitive abilities.

\subsection{Physical Symbols}\label{Sec:SupplementaryPhysical}

% \begin{table}[htbp]
% \centering
% \caption{Accuracy Scores Across Models and Tasks in Physical Symbol}
% \label{tab:scores}
% \begin{tabular}{l *{9}{c}}
% \toprule
% \textbf{Model} & \multicolumn{3}{c}{\textbf{Level 1}} & \multicolumn{3}{c}{\textbf{Level 2}} & \multicolumn{3}{c}{\textbf{Level 3}} \\
% \cmidrule(lr){2-4} \cmidrule(lr){5-7} \cmidrule(lr){8-10}
% & \textbf{task1} & \textbf{task2} & \textbf{avg} 
% & \textbf{task3} & \textbf{task4} & \textbf{avg} 
% & \textbf{task5} & \textbf{task6} & \textbf{avg} \\
% \midrule
% Deepseek-vl2-tiny (3B)     & 19.3 & 20.0 & 19.6 & 18.7 & 26.3 & 19.1 &  0.0 & 16.0 &  8.2 \\
% Qwen2.5-vl (7B)            & 17.1 & 16.5 & 16.9 & 14.5 & 35.6 & 15.7 &  1.4 & 19.8 & 12.2 \\
% llama3-llava-next-8b       &  2.1 &  0.0 &  1.8 &  2.3 &  0.6 &  2.1 &  0.0 &  3.8 &  1.9 \\
% InternVL3-8B               & 16.6 & 16.7 & 16.7 & 12.8 & 33.3 & 13.2 &  0.0 & 21.7 & 11.8 \\ \midrule
% GPT-4o                     & 13.6 & 25.0 & 14.1 & 15.4 & 12.1 & 15.3 &  9.7 & 18.2 & 14.8 \\
% Claude-sonnet-4            & 26.7 & 33.3 & 27.1 & 39.8 & 50.2 & 40.4 & 16.9 & 24.2 & 19.4 \\
% Qwen-max                   & 11.8 &  9.6 & 10.1 & 15.4 & 26.3 & 16.1 &  8.3 & 16.7 & 12.9 \\
% o3-2025-0416               & 31.9 & 16.6 & 30.7 & 41.1 & 26.7 & 40.9 & 19.3 & 35.0 & 28.7 \\
% Gemini-2.5-pro             & 36.7 & 60.2 & 41.2 & 58.8 & 66.7 & 59.5 & 36.2 & 57.1 & 46.3 \\
% \bottomrule
% \end{tabular}
% \end{table}

\subsubsection{Limited recognition of sparse physical symbols across most models}

As shown in Figure~\ref{fig:physics_results}, most models performed poorly on the Level 1 (perception and recognition) task for physics symbols, with mean accuracies below 30\%. For example, GPT-4o achieved an accuracy of 14.1\%, Qwen2.5-vl 16.9\%, and LLaMA3-llava-next-8b was almost entirely non-functional, reaching 1.8\%. This widespread low performance indicates that current MLLMs have a deficiency in their ability to translate visual symbols into expressions of physical quantities and formulaic structures.

It is evident that models are generally able to recognize and recite the textual descriptions of physical laws but fail to correctly perform the mathematical mapping at the symbolic level. For instance, in Case 11 of Figure~\ref{fig:physics_results}, a model could correctly state the definition of Ohm's law relevant to the problem but, during symbolization, incorrectly wrote the quadratic relationship between power and current as a linear proportional one, and misidentified the straight-line graph in option B as a parabola. Such errors reflect that the model's understanding of formula structures remains at a linguistic level, lacking both geometric intuition for the functional relationships between physical quantities and an awareness of symbolic consistency.

Furthermore, in Case~\ref{case:physicalL1_2} in Supplementary Sec.~\ref{Sec:SupplementaryCases}, although a model could identify the required principle of conservation of energy, it made a critical parameter error during substitution: the original problem stated $H=\frac{3mg}{k}$, but the model incorrectly wrote it as $H=\frac{mg}{k}$. Although its reasoning path was logically sound for the most part, the deviation in the numerical stage led to an incorrect final conclusion. This indicates that when faced with a complex physics symbol system involving multiple steps and intertwined concepts, the stability and accuracy of its reasoning chain are easily disrupted.

The reasoning process of Deepseek-vl2-tiny was opaque and unstable. On some problems, such as Case~\ref{case:physicalL1_2} in Supplementary Sec.~\ref{Sec:SupplementaryCases}, it was able to directly generate the correct answer without providing any intermediate reasoning, whereas on others (like Case 11 of Figure~\ref{fig:physics_results}), it offered an incorrect explanation that ``power increases linearly with current.'' This inconsistency suggests that its correct answers may originate from template retrieval or pattern matching of its training corpora, rather than from systematic deduction based on physical principles. In contrast, Gemini-2.5-pro and o3 demonstrated more stable performance. Notably, Gemini-2.5-pro achieved a 60.2\% accuracy on the electrical symbol recognition task, showing that its visual encoder possesses higher parsing precision when handling formula symbols with low pixel density and partial overlaps.

\subsubsection{Improved performance through symbolic reasoning yet with marked divergence}

In the Level 2 (combination and reasoning) task, the score distribution among models follows a similar pattern to Level 1 (perception and recognition), with Gemini-2.5-pro and o3 still significantly ahead, while GPT-4o and the Qwen series generally hover around the low level of 15 points. As seen in Case~\ref{case:physicalL2_1} in Supplementary Sec.~\ref{Sec:SupplementaryCases}, models are capable of listing numerous general formulas related to the motion of charged particles, indicating they possess a certain reserve of physics knowledge. However, when required to combine these formulas with specific conditions from the problem, such as the electric field width, the models often show a disconnect. They lack systematic task-planning ability and are unable to select the correct chain of formulas for simultaneous solving. As a result, the reasoning process remains at the level of formula stacking and superficial pattern matching, ultimately halting after listing several equations or producing a seemingly plausible yet physically incorrect answer through the model’s guessing mechanism. This phenomenon indicates that while current multimodal models may know which physical laws to apply on a knowledge level, they have a significant deficiency in dynamically integrating symbols, parameters, and spatial constraints, lacking the ability to establish a continuous logical pathway from visual input to formulaic deduction.

\subsubsection{Global correction and prediction remain challenging for most models}

In the Level 3 (association and critical thinking) task, task 5 (mechanical diagram consistency correction) emerged as a common bottleneck for all models. With the exception of Gemini-2.5-pro, o3, and Claude-sonnet-4, which achieved relatively acceptable scores, the remaining models almost universally failed, with some even scoring zero. This result highlights a significant deficiency in the capabilities of current Multimodal Large Language Models for symbolic error correction and physical logic consistency reasoning within complex visual scenarios.

As shown in Case 12 of Figure~\ref{fig:physics_results}, although the model is able to accurately identify multiple basic components in the circuit diagram, this recognition often relies on the semantic assistance of letter markings in the diagram, such as 'A' for Ammeter and 'V' for Voltmeter, rather than on a genuine visual analysis of the symbols' morphology. However, the model still lacks the ability to integrate visual symbol semantics with the physical context. It mistakes the slider P of the sliding rheostat for the symbol P representing power, and proceeds to make incorrect logical deductions on this basis.

\subsection{Chemical Symbols}\label{Sec:SupplementaryChemical}

To investigate the cognitive limits of MLLMs in the highly symbolic domain of chemistry, we designed and implemented a benchmark with three difficulty levels. This benchmark simulates the cognitive progression from basic chemical knowledge recognition to advanced chemical reasoning and error correction, encompassing the comprehension of core chemical notations, such as structural formulas and reaction equations.

\subsubsection{Most models show limited capability in basic chemical symbol recognition}
We can observe the overall results for the Level 1 (perception and recognition) tasks in Figure \ref{fig:chemistry_exp}. Although most models exhibit some preliminary recognition capabilities at a basic level, their overall accuracy remains low. Gemini-2.5-pro was the most prominent performer, achieving accuracies of 46.1\% and 26.7\% in task 1 (element identification in structural diagrams) and task 2 (chemical bond recognition), respectively, with a mean score of 39.4\% that was superior to other models. In contrast, mainstream multimodal models, including GPT-4o and Qwen2.5-vl, generally scored lower on task 2 than on task 1, with some even dropping to zero.

From a qualitative analysis, most models can identify explicitly rendered chemical symbols in an image, such as the clearly labeled N and O atoms in Case 13 of Figure~\ref{fig:chemistry_exp}. However, they universally ignore the implicit rules of chemical skeletal formulas, wherein each vertex and endpoint in the skeletal structure represents a carbon atom, and the number of attached hydrogen atoms must be inferred based on valency. Because the models fail to understand this, both C and H atoms are omitted, leading to incorrect parsing of the molecular structure. While some models can identify high-level structures like ``benzene ring'' and ``nitro group,'' demonstrating an ability to capture local features, their atom-counting process relies on memorized templates rather than actual structural analysis. For example, they directly combine the formulas for a ``benzene ring'' and a ``nitro group,'' mechanically adding them to get 6 hydrogen atoms, but fail to infer from the image that one hydrogen on the ring had been substituted by the nitro group; the correct result should be 5 hydrogen atoms.

In Case 14 of Figure~\ref{fig:chemistry_exp}, most models are able to identify single and double bonds, but their counts are significantly lower than in the actual structure. This issue reflects the difficulty current MLLM visual encoders have in processing sparse lines, intersecting connections, and structural abbreviations, making them unable to effectively distinguish the types and quantities of bonds. Furthermore, models fail to identify the aromatic ring structure present in the diagram, which is neither a pure single nor a pure double bond. This indicates that current models lack the ability to jointly model symbolic hierarchy and chemical semantics, and thus still struggle to truly understand the chemical meaning behind the structural symbols.

\subsubsection{Rule-based symbolic integration drives reasoning yet exposes structural gaps}
When the task shifted from static symbol recognition to dynamic reasoning requiring the application of chemical laws, the performance gap among models widened. The results show that o3 performed best at Level 2 (combination and reasoning), followed by Claude-sonnet-4 and Gemini-2.5-pro. Notably, although Gemini-2.5-pro exhibited the strongest symbol recognition ability, its overall score in Level 2 (combination and reasoning) was lower than o3's. This suggests that o3 possesses stronger capabilities for reasoning chain integration and chemical knowledge transfer; despite being slightly weaker in visual parsing, it holds an advantage in tasks that require quantitative balancing and reaction type judgment. In contrast, the performance of LLaMA3-llava-next-8b and Qwen-max lagged, reflecting their deficiencies in constructing stable reasoning chains and integrating information from both the symbol and rule levels.

From a qualitative perspective, in Case 15 of Figure~\ref{fig:chemistry_exp}, some models were able to correctly identify chemical symbols in an equation and detect coefficient anomalies, demonstrating a preliminary capacity for anomaly detection. However, their proposed corrections were often incorrect, indicating that the models rely primarily on pattern-matching reasoning—for example, simply identifying that ``the equation needs balancing''—rather than a true understanding of the principle of atom conservation. More typically, even when models detected an imbalance, their reasoning chains became disordered during the balancing attempt. There was a lack of a causal link between adjusting formulas and verifying counts, leading them to cyclically output the same step or get stuck in non-convergent, repetitive calculations.

\subsubsection{Only a few models succeed in global chemical correction and prediction}

In the Level 3 (Association and Critical Thinking) tasks, models perform multi-level reasoning across complex chemical reaction diagrams and textual descriptions, including reaction type discrimination, product prediction, and condition analysis. Such tasks present a comprehensive challenge to a model's capabilities in symbol understanding, rule transfer, and causal generation. Overall, Gemini-2.5-pro again performed best, achieving a high score of 86.7 in task 7 (reaction product prediction), which demonstrates its capacity to generate results grounded in chemical knowledge structures. Conversely, although o3 and Claude-sonnet-4 performed adequately on the Level 2 (Combination and Reasoning) reasoning tasks, they were significantly weaker on the multi-dimensional prediction tasks in Level 3 (Association and Critical Thinking). This is likely due to their weaker symbol recognition capability, which limits their capacity to integrate information globally. Interestingly, Qwen-max, which consistently lagged in Level 1 and Level 2, jumped to third place in the Level 3 average score. This suggests that its stronger global understanding may have compensated for its deficiencies in symbol parsing, enabling it to complete higher-level chemistry tasks in a leapfrogging manner. This could also be related to the inclusion of chemistry topics in its training set.

Case~\ref{case:chemistryL3_1} in Supplementary Sec.~\ref{Sec:SupplementaryCases} shows that some models can identify key reactants in a chemical equation, such as an alkyl halide and a lithium reagent, and proactively invoke internal knowledge during their analysis, such as ``lithium reagent reactions typically need to be conducted at extremely low temperatures.'' They can then use this knowledge to filter options and ultimately generate the correct answer. This indicates that models are beginning to exhibit a principle-based heuristic reasoning capability. However, this ability is not yet stable. Errors in some models often stem from early-stage symbol recognition deviations, such as misreading bond types or omitting substituents, which causes the entire subsequent logical chain to fail.

\begin{figure}[t]
    \centering
    \includegraphics[width=1\linewidth]{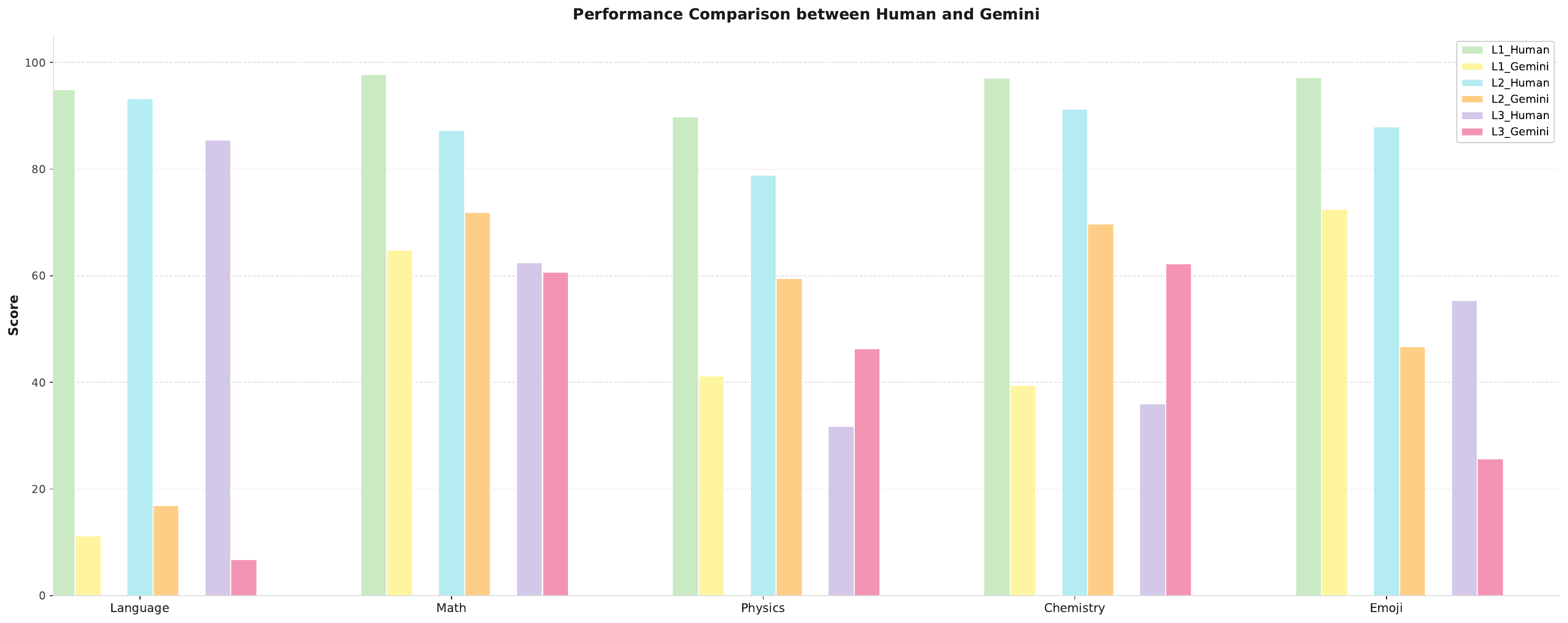}
    \caption{Human performance on the benchmark compared with the best-performing model Gemini.}
    \label{fig:human}
\end{figure}

\subsection{Human Performance Baseline}\label{Sec:SupplementaryHumanBaseline}

To further benchmark the cognitive alignment of Multimodal Large Language Models (MLLMs), we establish a human performance baseline to evaluate the processing of complex abstract visual symbols in real-world scenarios. By comparing human performance across diverse domains and hierarchical levels, we aimed to discern whether model reasoning stems from true visual perception or a reliance on linguistic priors.

We conduct a stratified sampling of 1,000 instances from our comprehensive dataset of over 13,000 samples, ensuring a 95\% confidence level. The sample was proportionally distributed across the cognitive hierarchy: 253 samples for Level 1 (Perception and Recognition), 439 for Level 2 (Combination and Reasoning), and 308 for Level 3 (Association and Critical Thinking). Five professional volunteers, all holding Master’s degrees or higher and possessing high bilingual proficiency in Chinese and English, participated in the experiment. They were instructed to complete the tasks using the identical prompts provided to the MLLMs to ensure a rigorous and fair comparison. As shown in Figure~\ref{fig:human}, the experimental results from the human-baseline comparison demonstrate that human performance follows a canonical cognitive trajectory, where accuracy steadily declines as the hierarchical level and task complexity increase. This trend empirically validates our three-level framework as a true reflection of escalating cognitive demand. In stark contrast, even the most advanced MLLMs exhibit a profound performance gap at the foundational level. In the Language and Chemistry domains specifically, humans achieved near-ceiling performance in Level 1 tasks, confirming these as intuitive, perceptual baselines for the human visual system.
Conversely, MLLMs frequently exhibit the "Recognition–Reasoning Inversion" phenomenon, where they perform significantly better on complex reasoning tasks than on basic symbol recognition. This disparity confirms that current models do not adhere to human-like visual-cognitive logic. Instead, they appear to rely on learned linguistic probabilities and procedural imitation rather than robust visual grounding. Consequently, the models' ``visual understanding'' is often governed by what they statistically expect to see rather than the actual visual input, a cognitive mismatch that inherently leads to biases and hallucinations.

\subsection{Comprehensive analysis}\label{Sec:SupplementaryComprehensiveAnalysis}

\begin{figure}[t]
    \centering
    \includegraphics[width=0.75\linewidth]{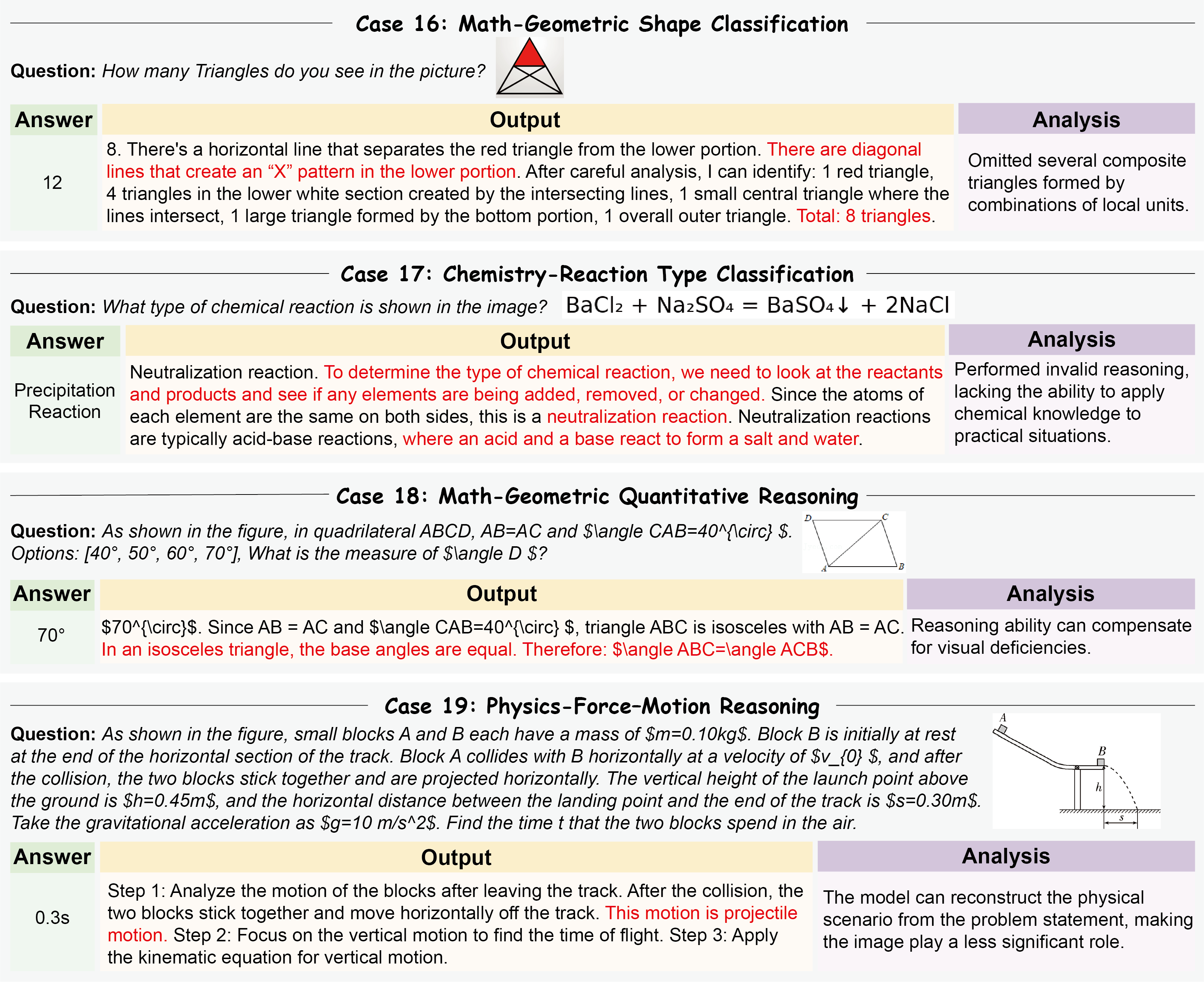}
    \caption{Case studies from the comprehensive analysis, highlighting model failures in recognizing and integrating symbols.}
    \label{fig:comprehensive_case}
\end{figure}

\subsubsection{Models generally lack the ability to recognize complex symbols}

% \begin{figure}
%     \centering
%     \includegraphics[width=0.9\linewidth]{figures/comprehensive_case.png}
%     \caption{Case studies of comprehensive analysis.}
%     \label{fig:comprehensive_case}
% \end{figure}

% 模型普遍缺乏复杂符号的识别能力
Contrary to the common intuition that recognition is simpler than reasoning, our results indicate that foundational Level 1 (perception and recognition) remains a significant weakness for nearly all models, a pattern illustrated by the representative failures in Figure~\ref{fig:comprehensive_case}. This phenomenon was particularly prominent for language, chemistry, and physics symbols, where almost all models, including top-tier ones like Gemini 2.5 Pro and GPT-4o, exhibited high error rates. This suggests that while the visual systems of current MLLMs excel at processing the holistic features of natural images, such as texture, shape, and semantics, they lack the capacity for systematic and compositional parsing when faced with formalized, abstract symbolic systems. In one representative comprehensive case, the model demonstrates good descriptive capabilities in its output. It identifies the large triangle, the red areas, and the intersecting line segments, and attempts a hierarchical count based on geometric intuition. However, it ultimately arrives at only 8 triangles, omitting several composite triangles formed by the combination of local units. It fails to treat adjacent basic shapes as a whole to perceive and construct new symbolic entities formed by the combination of these parts and superimposed upon the original figure.

\subsubsection{The models exhibit rote memorization rather than genuine understanding}
%模型展现出“死记硬背”而非真实理解
The success of models on certain Level 2 (combination and reasoning) procedural tasks stands in contrast to their failures in Level 1 (perception and recognition), suggesting that their success stems not from genuine understanding but from a form of procedural imitation. This phenomenon is most pronounced in the chemistry domain. For example, Qwen-max scored 0 in Level 1 yet achieved a certain score in Level 2. This discrepancy indicates that its success does not follow a logically coherent process of first recognizing, then understanding, and finally solving the problem. Instead, it is more likely relying on powerful pattern-matching capabilities, effectively ``memorizing'' the overall visual paradigms of numerous chemical equations in its training data. A similar phenomenon can be observed in mathematics, where the excellent performance of the Qwen series and Claude-sonnet-4 is related to the large amount of mathematical content included in their training corpora.

In another representative comprehensive case, although the model could correctly identify the reactants and products in the image, it chose to determine the reaction type by comparing whether the types and quantities of elements changed before and after the reaction. However, in a correctly balanced chemical equation, the types and quantities of elements are inherently conserved, rendering this process an invalid reasoning step. Ultimately, the model concluded that it was a ``neutralization reaction'' and could correctly provide the definition of one, but this definition was irrelevant to the given problem and could not be genuinely applied to the current reaction. In other words, the model was merely reciting memorized chemical knowledge at a semantic level but failed to apply it correctly to the specific context.

\subsubsection{Reasoning ability sometimes exists as a compensatory mechanism}
%高级推理能力有时作为一种补偿机制而存在
A model's language reasoning ability can, to some extent, compensate for its deficiencies in visual perception. For instance, InternVL-2.6-8B performed better on the Level 3 (association and critical thinking) in the linguistic symbols domain than on the foundational tasks in Level 1 (perception and recognition) and Level 2 (combination and reasoning). Similarly, Gemini-2.5-pro's performance on the comprehensive physics reasoning task surpassed its results in the symbol recognition stage. This suggests that the vision and language modules in these models are not yet deeply integrated, functioning more as complementary components. When visual recognition is inaccurate, the powerful language reasoning module often infers an answer based on contextual patterns or common logical paths, thereby masking or even compensating for the shortcomings of the visual system.

In a representative geometric reasoning task, the model did not explicitly identify which edges or angles were collinear from a visual standpoint. However, by leveraging its logical grasp of an ``isosceles triangle'' from the textual condition ``AB = AC,'' it automatically deduced that ``$\angle ABC=\angle ACB$'' and proceeded to calculate the correct answer based on the knowledge chain that ``adjacent angles of a parallelogram are supplementary.'' In this process, the role of the image was weakened to that of scenario confirmation, while the model’s strong language reasoning ability bypassed the need for a precise perception of geometric relationships from the image.

\begin{figure}
    \centering
    \includegraphics[width=0.75\linewidth]{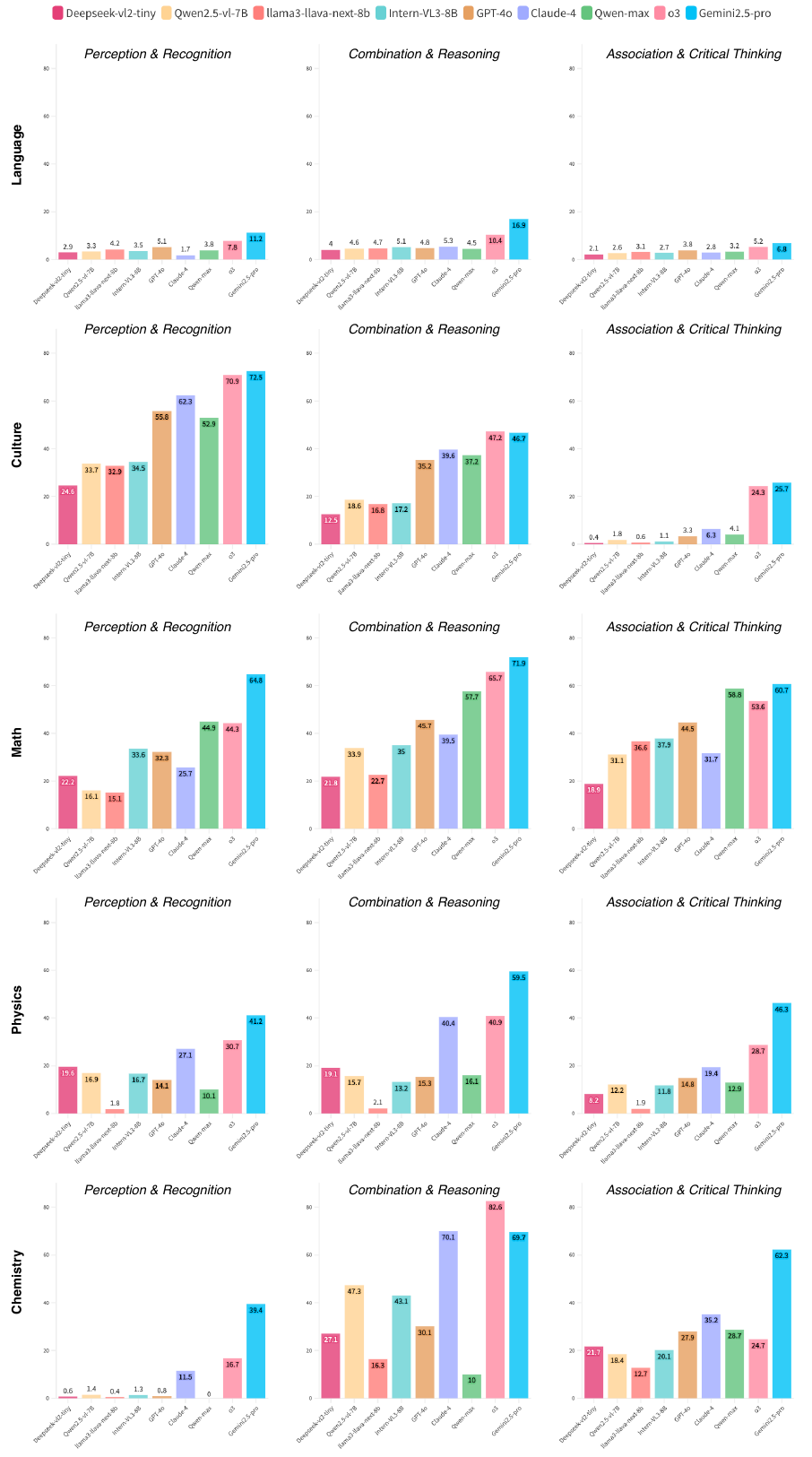}
    \caption{Overall performances of different models across the five symbolic domains.}
    \label{fig:main_results}
\end{figure}

Similarly, in a representative physics scenario, even if the model failed to recognize structural features like the track and object positions during the image analysis phase, it could still reconstruct the physical scenario from the text of the problem description. By identifying phrases such as ``launch point height h = 0.45 m'' and ``B is at rest at the end of the horizontal section of the track,'' it automatically invoked kinematics knowledge to deduce the time. This behavior indicates that the model's reasoning relies more on the dispatch of linguistic knowledge and logical chains than on directly extracting quantitative information from the visual input.

\subsubsection{There is no model that excels in all domains}
%跨域迁移不足，没有完美的模型
% No single model demonstrated robust and consistent performance across all five symbolic domains. \qy[This paragraph has only one sentence?]{}

None of the models exhibit robust and consistent performance across all five symbolic domains. As summarized in Figure~\ref{fig:main_results}, strong performance was typically concentrated in domains where pre-training data is likely more abundant and diverse, such as cultural symbols and mathematics. For instance, models like o3 and Gemini-2.5-pro achieved their highest scores on Level 1 (perception and recognition) tasks in the cultural symbols domain, likely benefiting from the vast corpora of related content in web data. Similarly, models with a focus on reasoning showed strong capabilities in mathematics, particularly on programmatic tasks. In stark contrast, these same top-performing models often exhibited a significant performance decline in domains less represented in pre-training data, such as chemistry and physics. This indicates that the strengths of current MLLMs are more associative and empirical, adept at leveraging massive corpora from common domains, rather than deductive and systematic.

\subsubsection{The model structure may be the culprit for insufficient understanding of discrete symbolic.}
Our analysis suggests that the failure of MLLMs in sparse symbol recognition (e.g., skeletal formulas and circuit diagrams) is not merely a consequence of data scarcity, but an inherent artifact of the Vision Transformer (ViT) architecture. Standard ViTs partition images into fixed-size patches, processed via global self-attention with coarse-grained positional encodings. This mechanism is optimized for capturing low-frequency semantic features in natural scenes but effectively acts as a \textit{spatial low-pass filter} for discrete symbols. In sparse structures like chemical bonds (Case 14) or electrical junctions (Case 12), the precise coordinates and connectivity are often "blurred" or aliased within the attention maps, leading to the observed inability to maintain symbolic consistency. The model’s reliance on linguistic priors thus becomes a compensatory survival strategy for its deficient perceptual resolution at the token-grid level.

\section{Extended Methods}\label{Sec:SupplementaryExtendedMethods}

To evaluate the visual and intuitive understanding of complex symbols by multimodal large language models (MLLMs), we construct a comprehensive benchmark spanning multiple domains, including language, art, mathematics, physics, and chemistry. The symbols involved include incorrect or substituted characters, emojis, musical notations, function symbols, geometric shapes, mechanical and electrical symbols, as well as chemical structural diagrams. These symbols represent a broad array of conventional forms commonly encountered across various facets of human culture. 

\begin{figure}[t]
    \centering
    \includegraphics[width=0.9\linewidth]{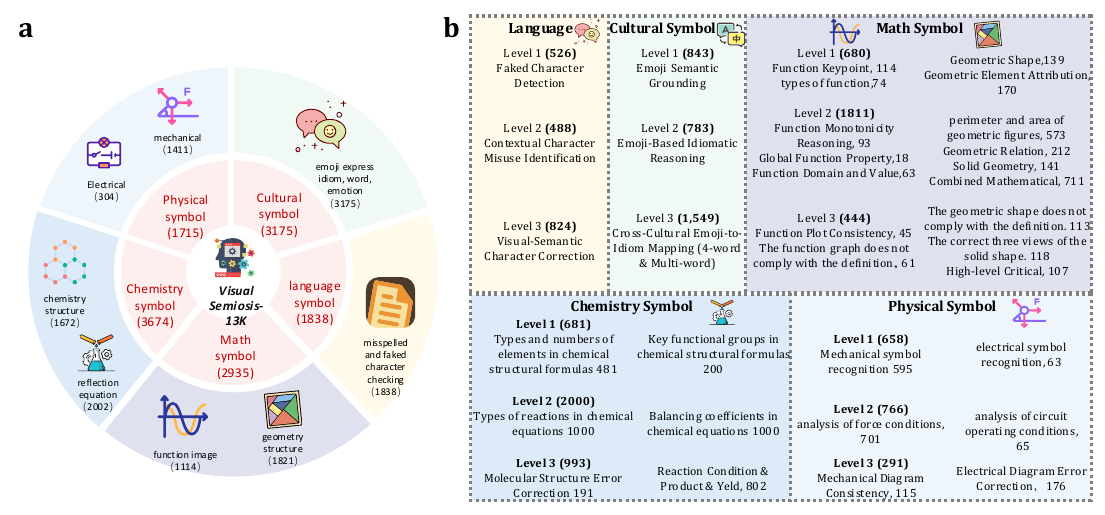}
    \caption{Overview of the benchmark task design, including language symbols, cultural symbols, math symbols, chemistry symbols and physical symbols.}
    \label{fig:data_sta}
\end{figure}

We propose a three-level symbolic understanding task framework to systematically assess MLLMs across different modalities. The tasks range from basic recognition of individual symbols to more sophisticated associative and inferential reasoning. Each domain-specific task set is designed to reflect this progression, allowing us to investigate how models respond to different types of symbols under increasing cognitive demands. We curated datasets from as many fields as possible, with new annotations tailored to our task objectives, and conducted in-depth evaluations and analyses of multiple state-of-the-art multimodal models. The full task-organization schema is shown in Figure~\ref{fig:data_sta}.

\subsection{Task Design}\label{Sec:SupplementaryTaskDesign}

% According to the hierarchy of human symbolic cognition, our benchmark defines three levels of symbolic understanding tasks. At the first level, models are expected to recognize the meaning of individual symbols, such as identifying atomic types in a chemical structure. The second level involves multi-symbol composition, requiring models to perform reasoning or computation based on the combination of multiple symbolic elements, such as determining the monotonicity of a function. The third level demands critical thinking and symbolic creativity—models must correct erroneous symbol usage and infer abstract linguistic meanings from visual representations, such as identifying typos in handwritten text or interpreting the idiomatic meaning of an emoji sequence. For each domain, we construct task sets that cover these three levels, aiming to thoroughly evaluate symbolic understanding in context. 

According to the hierarchy of human symbolic cognition, our benchmark defines three levels of symbolic understanding tasks. 

At the first level, \textbf{``Perception and Recognition''}, the model is required to understand the symbolic semantics represented by individual basic symbols. For the understanding of a single complex symbol, it is necessary to first extract its visual features (e.g., lines, shapes, colors, etc.), learn the visual semantics of the visual input itself, and then convert it into a meaningful symbolic semantic unit. For example, seeing a ``red light'' can identify that it has the symbolic semantics of ``danger''. After continuous training through daily life and social culture, the human brain can directly convert visual symbolic input into corresponding semantics. We aim to observe the model's recognition ability at this level, thereby exploring whether multimodal large language models can learn such human-like ability to directly understand the abstract textual semantics corresponding to complex visual symbols under the training of large-scale image-text data currently available.

At the second level, \textbf{``Combination and Reasoning''}, the model needs to further perform more global understanding and reasoning in the discrete semantic space of images. It is required to combine multiple symbols, understand the joint semantics expressed by these symbols in the discrete semantic space, and even conduct further reasoning. For example, at Level 1, it is only necessary to understand the meaning of individual electrical symbols in a circuit diagram, while at Level 2, it is necessary to combine multiple electrical symbols and their connecting wires to infer the connection mode and working state of the entire circuit. This necessitates the model to possess semantic knowledge of symbols and their interrelationships on the premise of symbol comprehension, and also to examine the model's reasoning ability.

Finally, building on the reasoning ability for symbol combination, the third level focuses on high-level associative and critical thinking abilities, \textbf{``Association and Critical Thinking''}. At the first two levels, almost all symbols presented to the model are correct, and most of the information they convey is the inherent meaning of the symbols themselves. However, at the third level, the model is required to break through this limitation. When encountering incorrect symbols or symbols that convey unconventional symbolic semantics, it must perform in-depth semantic judgment and reasoning based on the context. Among them, associative ability is the support of critical thinking. The model first associates through the contextual context, based on knowledge reserves such as the conventional semantics of symbols, relevant cultural backgrounds, and usage patterns. It then engages in critical thinking based on this associative information to complete the prediction of the rationality of symbols, correction of errors, and interpretation of special semantics. Specifically, in real-world scenarios, humans construct critical cognition of symbols precisely through association based on contextual cues: on the one hand, when facing deviated or defective symbols (e.g., the typo ``chioce''), the human brain employs an autonomous correction mechanism, which can automatically associate it with the correct symbol (``choice'') according to the textual or situational context in which it is located, without disrupting the understanding of the overall context. This process requires not only associative ability to guide the direction of correction but also critical thinking to verify the rationality of the correction. On the other hand, for symbol combinations with special usages (e.g., homophonic emoji expressions), it is necessary to first associate information such as the inherent meaning of symbols, their connection with the context, and homophonic conventions in social culture. Critical thinking is then used to overcome the interference of the inherent meaning of symbols and to interpret the unconventional semantics that differ from these inherent meanings. Therefore, this level primarily examines the model's ability to accurately capture subtle differences in symbolic semantics, as well as the high-level semantic understanding and reasoning ability, guided by association and centered on critical thinking within contextual scenarios.

Overall, targeting the special discrete semantic visual information of visual complex symbols, we have constructed a multi-level benchmark covering individual symbol understanding, combined symbol reasoning, and high-level association and critical thinking based on the special understanding mechanism of the human brain. Corresponding tasks have been designed and datasets established in multiple fields to achieve comprehensive evaluation and exploration of the symbol understanding mechanism of multimodal large language models.

% \qy[it may be hard to distinguish level 2 and 3 for some domains. Seems like there is some overlap. The results also show this?]{} \Response[We have clarified the task design, especially for giving examples for level 2 and 3, with detailed explaination.]{}

\subsubsection{Language Symbols}

Language constitutes one of the most fundamental symbolic systems in human cognition and serves as the primary modality processed by large language models. In this benchmark, we focus on Chinese characters as a representative linguistic symbol system with strong visual compositionality. Unlike alphabetic writing systems, Chinese characters encode semantic and phonetic information through structured visual components and strokes, making them an ideal testbed for evaluating visually grounded symbolic understanding.

At the first level, the benchmark evaluates the model’s ability to perceive and recognize invalid or non-existent characters, referred to as faked characters. These symbols arise from stroke-level or component-level perturbations and do not belong to any standardized character set. Since such characters cannot be resolved through lexicon lookup or OCR-based transcription, successful identification requires fine-grained visual perception and structural analysis of character morphology.

Building upon basic recognition, the second level focuses on contextual semantic reasoning over valid but incorrectly used characters. Misused characters are visually or phonetically similar to the intended ones and are grammatically valid in isolation, yet semantically incompatible with the surrounding context. This level examines whether the model can integrate visual recognition with linguistic context to detect semantic inconsistencies and identify the source of misuse.

At the third level, the benchmark targets cross-modal correction capability. The model is required to actively correct faked characters or misused characters by jointly leveraging visual structure, phonetic similarity, and contextual semantics. This level reflects a higher-order symbolic competence, assessing whether the model can construct stable mappings between visual perception and linguistic reasoning to perform autonomous error correction.

\subsubsection{Cultural Symbols}

With the proliferation of online communication and social media, emojis have evolved into a globally shared yet culturally nuanced symbolic system. Although emojis are standardized in appearance, their meanings are not fixed; they emerge through collective usage, cultural conventions, and contextual adaptation. As a result, emojis function not only as visual icons but also as culturally grounded symbols that convey emotions, intentions, and idiomatic meanings beyond their literal depiction.

At the first level, the benchmark evaluates the model’s ability to ground individual emojis in their commonly accepted semantic representations. This task examines whether the model can map visual emoji symbols to corresponding lexical meanings, reflecting basic visual-semantic alignment.

The second level advances to compositional reasoning, where sequences of emojis jointly express idiomatic meanings in English. In these tasks, emojis act as substitutes for words or phrases, requiring the model to infer sentence-level semantics from their symbolic composition. By restricting target interpretations to widely recognized English expressions, this level emphasizes symbolic combination and syntactic reasoning while minimizing ambiguity.

The third level introduces cross-cultural and cross-linguistic symbolic reasoning. Given an emoji sequence, the model must infer its corresponding Chinese idiom, which often relies on phonetic resemblance, metaphorical association, or culturally specific conventions. This level explicitly examines the model’s ability to transcend surface visual semantics and capture culturally mediated meanings, highlighting emojis as a form of cross-cultural symbolic language shaped by shared practices.

\subsubsection{Mathematical Symbols}

Mathematical reasoning represents a high-level form of symbolic cognition, characterized by abstraction, formal structure, and strict logical dependencies. In human cognition, mathematical understanding is not limited to linear symbolic expressions but heavily relies on visual representations such as function graphs and geometric diagrams, which provide spatial intuition for abstract concepts. In contrast to prior benchmarks that focus primarily on text-based formulas encoded in \LaTeX{}, this benchmark targets visually grounded mathematical symbols and evaluates whether models can establish stable semantic and logical relationships directly from images.

To reflect the internal structure of visual mathematics, our tasks are organized around two major sub-fields: \emph{function graphs} and \emph{geometric figures}. Across both sub-fields, we design a three-level hierarchy that progressively evaluates perception, reasoning, and critical verification.

At the first level (\emph{Perception and Recognition}), the model is required to identify fundamental mathematical entities and structural components from images. In the function sub-field, this includes recognizing key point sets (e.g., zeros, extrema, and inflection points) and classifying the functional type (such as polynomial, exponential, or trigonometric). In the geometric subfield, tasks involve identifying the class of planar figures and attributing special geometric elements, including notable lines and angles. This level assesses whether the model can reliably extract symbolic primitives and local structures that serve as the basis for higher-level reasoning.

At the second level (\emph{Combination and Reasoning}), the benchmark shifts from isolated symbol recognition to integrated reasoning over complete mathematical structures. For function graphs, the model must infer global properties such as monotonicity, parity, periodicity, domain, and range, as well as perform value-based reasoning at specific points. For geometry, tasks require quantitative computation (e.g., length, area, or volume), reasoning about congruence and similarity, and understanding three-dimensional configurations through net diagrams and spatial relationships. These tasks examine whether the model can align local visual symbols with global mathematical constraints and axioms.

At the third level (\emph{Association and Critical Thinking}), the benchmark evaluates the model’s ability to verify, diagnose, and correct holistic mathematical representations. Tasks include detecting inconsistencies in function plots, judging whether a curve satisfies the formal definition of a function, validating whether geometric figures meet their claimed definitions, and identifying correct orthographic projections of solid objects. This level emphasizes structural consistency checking, error localization, and corrective reasoning, reflecting advanced mathematical understanding beyond direct computation.

\subsubsection{Chemical Symbols}

Chemical knowledge is encoded through a highly structured symbolic system that includes molecular structural formulas and chemical reaction equations. Structural diagrams represent atoms, bonds, and functional groups, while reaction equations describe transformation processes governed by conservation laws and energetic constraints. Together, these representations form a precise symbolic language that supports both static description and dynamic reasoning.

To reflect this internal structure, the chemical symbol tasks are organized into two sub-fields: \emph{molecular structural formulas} and \emph{chemical reaction equations}, evaluated across three hierarchical levels.

At the first level (\emph{Perception and Recognition}), the benchmark focuses on visual parsing of molecular structures. The model must identify element types, count atomic occurrences, and recognize different bond types within structural formulas. This level assesses whether the model can translate pixel-level information into a structured symbolic representation that faithfully captures molecular composition.

At the second level (\emph{Combination and Reasoning}), the tasks move beyond static structures to symbolic reasoning under chemical laws. The model is required to classify reaction types and balance chemical equations by enforcing element conservation. These tasks evaluate whether the model can reason over symbolic representations in accordance with fundamental chemical principles.

At the third level (\emph{Association and Critical Thinking}), the benchmark examines advanced chemical reasoning and correction abilities. Tasks include detecting and correcting errors in molecular formulas or reaction equations, inferring missing reaction conditions such as temperature, predicting reaction products, and estimating reaction yields. This level requires the integration of symbolic reasoning with domain-specific chemical knowledge, reflecting a mature understanding of chemical processes.

\subsubsection{Physical Symbols}

Similar to mathematics, physics relies heavily on symbolic and graphical representations to support understanding and reasoning. Especially in core fields like mechanics and electromagnetism, visual symbolic systems such as schematics, force diagrams, and circuit diagrams are widely used to abstractly depict complex physical processes. These diagrams encode rich information within a compact two-dimensional space, including force directions, field distributions, and structural connection relationships. Accurately interpreting these symbolic diagrams requires not only precise visual perception but also multi-level logical reasoning in conjunction with physical laws. Whereas existing evaluations primarily focus on text-based physics problems, which mainly assess the ability to extract parameters from text and perform numerical calculations, our benchmark instead proposes a more foundational and challenging paradigm: it emphasizes the understanding of visual symbolic content and the modeling of its physical meaning. It examines whether models can extract physical laws from symbolic graphics and achieve the cross-modal leap from visual perception to theoretical reasoning. Accordingly, we organize the physical symbol tasks into two sub-fields: \emph{mechanics} and \emph{electricity}, each evaluated through a three-level hierarchical framework.

In \emph{Perception and Recognition}, the benchmark evaluates whether the model can correctly parse fundamental physical symbols from diagrams. In mechanics, this involves recognizing force symbols, their directions, magnitudes, and points of application, as well as associating them with the corresponding objects. In electricity, tasks require identifying electronic components such as power sources, resistors, and wires, along with their connection patterns. This level focuses on the extraction of symbolic primitives and their local semantic roles.

\begin{figure}
    \centering
    \includegraphics[width=0.80\linewidth]{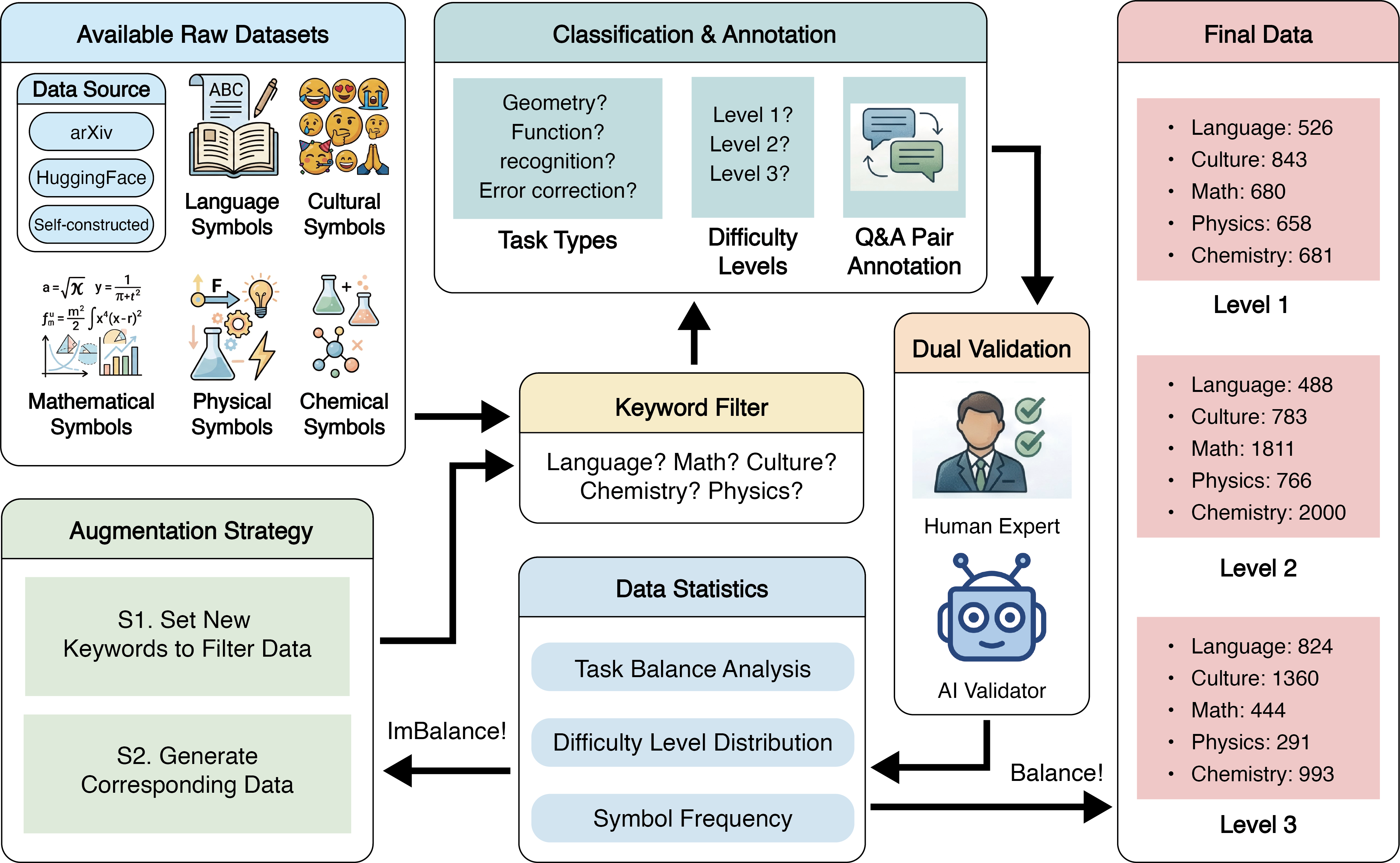}
    \caption{The data construction pipeline of our benchmark.}
    \label{fig:pipeline}
\end{figure}

Then, at \emph{Combination and Reasoning}, the model must integrate multiple symbols and infer system-level physical behavior. For mechanics, this includes reasoning about net forces, equilibrium conditions, acceleration directions, and motion tendencies under given constraints. For electrical systems, the model must infer current directions and component states based on circuit topology and source polarity. These tasks assess whether the model can map symbolic compositions to the governing laws.

At the third level (\emph{Association and Critical Thinking}), the benchmark targets high-order validation and correction capabilities. The model is required to detect and correct inconsistencies in mechanical or electrical diagrams, such as contradictory force directions or invalid circuit connections. Additionally, it must perform comprehensive reasoning in coupled systems, predicting system stability, energy flow, or behavioral changes under modified conditions. This level emphasizes holistic consistency checking and physically grounded critical reasoning.

\subsection{Dataset Construction}\label{Sec:SupplementaryDatasetConstruction}

After establishing the task domains and definitions for our hierarchical benchmark on complex symbol understanding, we proceed with the construction of the dataset. As illustrated in Figure~\ref{fig:pipeline}, the process begins with domain-specific data collection, followed by task-type and difficulty-level classification based on the design principles described in the previous section. Each data sample is annotated with a corresponding question-answer pair. Following the initial annotation, both automated and manual validation steps are performed to ensure data quality. Once the dataset distribution is statistically analyzed, we supplement additional samples where necessary to maintain a balanced representation across all task types.

\subsubsection{Data Sources}

To build our benchmark, we draw on a wide range of existing open-source multimodal benchmarks, selectively extracting data relevant to our task domains. For linguistic symbols, we utilize the VisualC3 dataset \cite{li-etal-2024-towards-real}, a high-quality collection of handwritten Chinese character errors and grammatical mistakes extracted from real-world student essays. This dataset provides authentic samples of miswritten and misused characters in handwritten form. For emoji symbols, data are sourced from eWe-bench \cite{kuang2025express}, which collects idioms and phrases expressed using emoji from internet communication. The data have undergone multiple rounds of automatic filtering and human verification to ensure both expression quality and representativeness.

For mathematical symbols, we extract function and geometry-related samples from several multimodal mathematics benchmarks, including MultiMath-300K \cite{peng2024multimath} and MathVista \cite{lu2024mathvista}. Chemistry-related data come from ChemBench-4K \cite{zhang2024chemllm}, Mini-CASIA-CSDB \cite{ding2022large}, and annotated middle and high school chemistry materials, which provide molecular structural diagrams, chemical equations, and their associated prediction problems. Physical symbol data are obtained from multi-disciplinary benchmarks such as OlympiadBench \cite{he2024olympiadbench}, MMMU-Pro \cite{yue2024mmmu}, and Gaokao-Bench \cite{zhang2023evaluating}. From these, we extract physics-related tasks involving mechanics and electrical circuits, covering a range of difficulties from university entrance exams to higher education-level problems.

\subsubsection{Data Annotation}

During annotation, we perform filtering, classification, and task-specific labeling of the raw data. The process begins with a coarse filtering step to extract data samples relevant to our task design. For example, from over 300K mathematical examples, we select items involving function plots and geometric figures. When existing domain tags are available, we retain and reclassify the samples accordingly—for instance, categorizing chemical structure images from Mini-CASIA-CSDB into the chemical structure domain. For unlabeled data, we use keyword matching or large language models (LLMs) to identify relevant content. The keywords and prompts used for each domain are detailed in Supplementary Sec.~\ref{Sec:SupplementaryExtendedMethods}.

After preliminary classification, we align the samples with our task framework and re-annotate each with an appropriate question and answer. For some datasets, this process is relatively straightforward. In VisualC3, for instance, we reorganize the image-question-answer triplets according to our three task levels, assigning the corresponding task labels. For chemistry tasks, we convert SMILES strings into rendered molecule images, replace them with placeholders in the original text, and redesign the questions and answers to match our framework.

Mathematics tasks require more extensive re-annotation, as most existing data do not include fine-grained task-level labels. We first explain the 17 task types defined across three difficulty levels, provide concrete examples, and use LLMs to generate initial annotations for domain type (function, geometry, or other), task level (Level 1–3), and task type (Task 1–17). These annotations are subsequently verified and corrected by human experts. Physics tasks follow a similar annotation pipeline.

In cases where data could not be sourced from existing datasets, we recruited trained annotators to construct examples manually. For error detection and correction tasks in chemistry and physics, which require uncommon miswritten symbols, domain experts with bachelor's degrees or higher were invited to hand-draw erroneous diagrams. Each annotator was provided with a correct reference image and instructed to introduce intentional symbolic mistakes while recording the error types.

For chemical structural symbols, error types include atom omission (EA), extra atoms (WB), bond type errors (CHG), charge errors (CHG), and stereochemistry errors (SC). For chemical equations, we identify element imbalance (UB), incorrect conditions (WC), invalid reaction arrows (WA), incorrect states (MS), and catalyst annotation errors (CAT). For mechanical diagrams, we define force direction errors (FD), missing forces (FM), extra forces (FE), and incorrect analysis objects (DA). For electrical circuit diagrams, the error types include incorrect component symbols (CE), short circuits (SP), open circuits (OL), labeling errors (CD), and misconnected meters (MP). To maintain balanced datasets, annotators were instructed to produce an approximately equal number of examples for each error category.

\subsubsection{Data Validation}

To ensure the high quality of the benchmark, we implemented a two-stage validation process combining automated and manual review. Automated validation is used to detect duplicate or missing entries and to verify the integrity and readability of image files. Manual validation is more nuanced and domain-specific. For linguistic and emoji symbol tasks, human experts evaluate whether the expressions align with natural usage and filter out samples containing nonstandard or inappropriate content. This is especially important for emoji data sourced from online platforms, where we remove samples involving discrimination, violence, or profanity to ensure ethical and safe data use.

For mathematics, physics, and chemistry tasks, human reviewers confirm the completeness and correctness of problem statements and answers. In mathematics tasks, particular attention is given to verifying task type and difficulty assignments. For chemistry and physics error-detection tasks, reviewers inspect whether annotated error types accurately describe the errors depicted in the image. After this validation phase, we retained approximately 96.7\% of the data, demonstrating the effectiveness of our annotation protocol and the overall reliability of the constructed dataset.

\subsubsection{Data Statistics and Analysis}

\begin{table*}[t]
\centering
\caption{Distribution of tasks across domains, sub-fields, and hierarchical levels.}
\label{tab:task_distribution}
\resizebox{\textwidth}{!}{
\begin{tabular}{llllrrr}
\toprule
Field & Sub-field & Task ID & Task Name & Level & Instances & Level Total \\
\midrule

\multirow{3}{*}{Language}
& Character & T1 & Faked Character Detection & Perception and Recognition & 526 & 526 \\
& Character & T2 & Contextual Character Misuse Identification & Combination and Reasoning & 488 & 488 \\
& Character & T3 & Visual-Semantic Character Correction & Association and Critical Thinking & 824 & 824 \\

\midrule
\multirow{4}{*}{Culture}
& Emoji & T1 & Emoji Semantic Grounding & Perception and Recognition & 843 & 843 \\
& Emoji & T2 & Emoji-Based Idiomatic Reasoning & Combination and Reasoning & 783 & 783 \\
& Emoji & T3 & Cross-Cultural Emoji-to-Idiom Mapping (4-word) & Association and Critical Thinking & 1,360 & \multirow{2}{*}{1,549} \\
& Emoji & T4 & Cross-Cultural Emoji-to-Idiom Mapping (Multi-word) & Association and Critical Thinking & 189 &  \\

\midrule
\multirow{17}{*}{Mathematics}
& Function & T1 & Function Keypoint Recognition & Perception and Recognition & 114 & \multirow{5}{*}{680} \\
& Function & T2 & Function Type Classification & Perception and Recognition & 74 &  \\
& Geometry & T3 & Geometric Shape Classification & Perception and Recognition & 139 &  \\
& Geometry & T4 & Geometric Element Attribution & Perception and Recognition & 170 &  \\
& General & T5 & Basic Mathematical Symbol Recognition & Perception and Recognition & 183 &  \\

& Function & T6 & Function Monotonicity Reasoning & Combination and Reasoning & 93 & \multirow{7}{*}{1811} \\
& Function & T7 & Global Function Property Inference & Combination and Reasoning & 18 &  \\
& Function & T8 & Function Domain and Value Reasoning & Combination and Reasoning & 63 &  \\
& Geometry & T9 & Geometric Quantitative Reasoning & Combination and Reasoning & 573 &  \\
& Geometry & T10 & Geometric Relation Inference & Combination and Reasoning & 212 &  \\
& Geometry & T11 & Solid Geometry Structure Reasoning & Combination and Reasoning & 141 &  \\
& General & T12 & Combined Mathematical Reasoning & Combination and Reasoning & 711 &  \\

& Function & T13 & Function Plot Consistency Verification & Association and Critical Thinking & 45 & \multirow{5}{*}{444} \\
& Function & T14 & Function Definition Validation & Association and Critical Thinking & 61 &  \\
& Geometry & T15 & Geometric Definition Consistency Check & Association and Critical Thinking & 113 &  \\
& Geometry & T16 & Orthographic Projection Identification & Association and Critical Thinking & 118 &  \\
& General & T17 & High-level Mathematical Analysis & Association and Critical Thinking & 107 &  \\

\midrule
\multirow{6}{*}{Physics}
& Mechanics & T1 & Mechanical Symbol Recognition & Perception and Recognition & 595 & \multirow{2}{*}{658} \\
& Electricity & T2 & Electrical Component Recognition & Perception and Recognition & 63 &  \\
& Mechanics & T3 & Force--Motion Reasoning & Combination and Reasoning & 701 & \multirow{2}{*}{766} \\
& Electricity & T4 & Circuit Operation Reasoning & Combination and Reasoning & 65 &  \\
& Mechanics & T5 & Mechanical Diagram Consistency Correction & Association and Critical Thinking & 115 & \multirow{2}{*}{291} \\
& Electricity & T6 & Electrical Diagram Error Correction & Association and Critical Thinking & 176 &  \\

\midrule
\multirow{8}{*}{Chemistry}
& Structure & T1 & Element Identification in Structural Diagrams & Perception and Recognition & 481 & \multirow{2}{*}{681} \\
& Structure & T2 & Chemical Bond Recognition & Perception and Recognition & 200 &  \\
& Reaction & T3 & Reaction Type Classification & Combination and Reasoning & 1,000 & \multirow{2}{*}{2,000} \\
& Reaction & T4 & Chemical Equation Balancing & Combination and Reasoning & 1,000 &  \\
& Structure & T5 & Molecular Structure Error Correction & Association and Critical Thinking & 191 & \multirow{4}{*}{993} \\
& Reaction & T6 & Reaction Condition Inference & Association and Critical Thinking & 300 &  \\
& Reaction & T7 & Reaction Product Prediction & Association and Critical Thinking & 202 &  \\
& Reaction & T8 & Reaction Yield Estimation & Association and Critical Thinking & 300 &  \\

\bottomrule
\end{tabular}
}
\end{table*}

Before finalizing the dataset, we conducted a detailed statistical analysis to examine the sample distribution across task types and ensure that no subtask was severely underrepresented. For subtasks with insufficient data, we revisited the data collection and annotation pipeline to augment them. This involved identifying additional data sources or applying self-instruct methods, in which large language models are guided to generate new samples based on example images, task descriptions, and annotated templates. Through this iterative augmentation strategy, we ensured a balanced and comprehensive benchmark.

The final benchmark dataset consists of \textbf{13,148 samples}, covering five symbolic domains and three hierarchical cognitive levels. Table~\ref{tab:task_distribution} provides an overview of the dataset composition across domains, levels, and sub-fields. From a hierarchical perspective, the dataset is relatively balanced across the three cognitive levels, with \textbf{3,388 samples} at the \emph{Perception and Recognition} level, \textbf{5,848 samples} at the \emph{Combination and Reasoning} level, and \textbf{3,912 samples} at the \emph{Association and Critical Thinking} level. 
This distribution reflects our design intent: while higher-level reasoning tasks are essential for evaluating symbolic intelligence, they are grounded in a substantial volume of lower-level perceptual and compositional tasks.

Across domains, the dataset spans a diverse range of symbolic systems, including Language (\textbf{1,838 samples}), Culture (\textbf{2,986 samples}), Mathematics (\textbf{2,935 samples}), Physics (\textbf{1,715 samples}), and Chemistry (\textbf{3,674 samples}). 
Each domain is internally structured according to its own symbolic characteristics and sub-fields, while adhering to the same three-level cognitive hierarchy.

\begin{figure}
    \centering
    \includegraphics[width=0.85\linewidth]{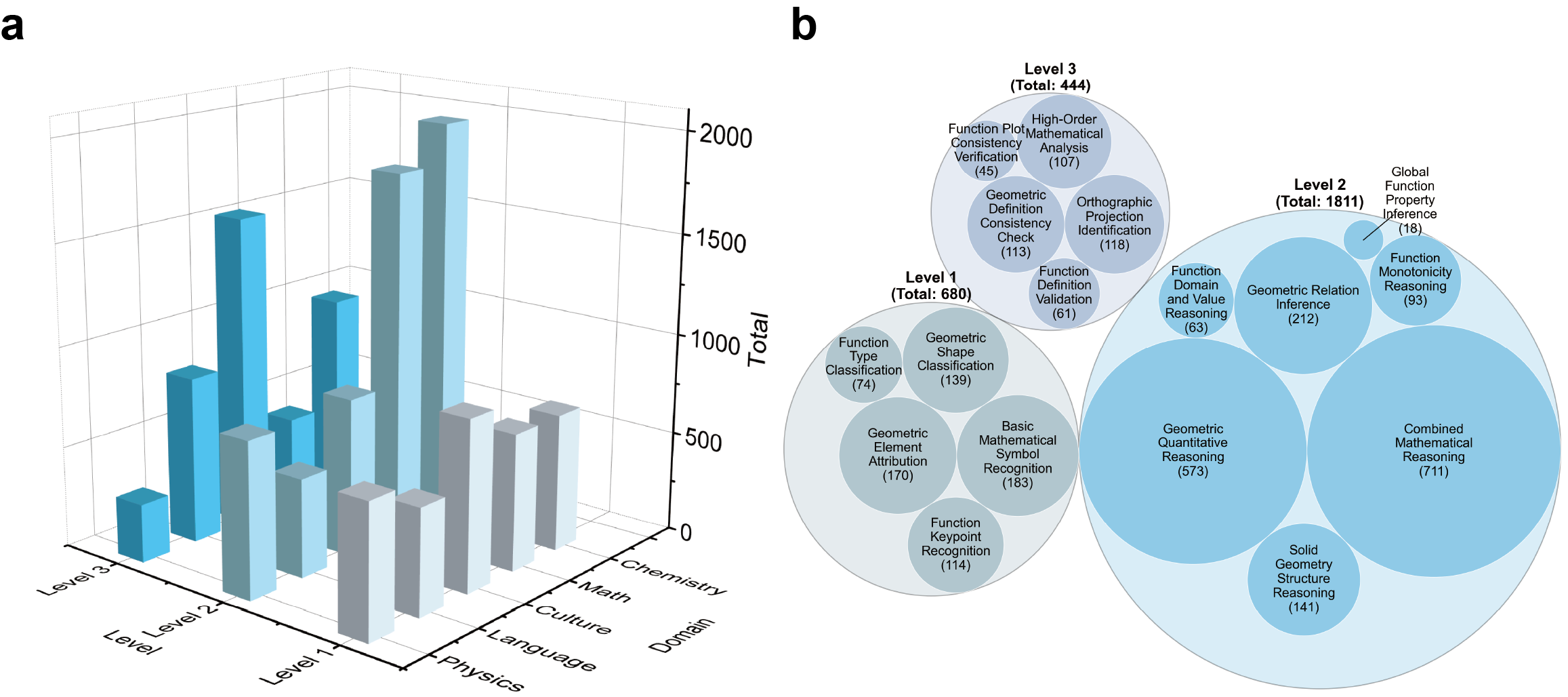}
    \caption{Dataset distribution. (a) Overall data distribution across five domains. (b) Detailed task distribution specifically for Mathematical symbols.}
    \label{fig:quantity_statistics}
\end{figure}

In the \emph{Language Symbols} domain, samples are distributed across all three levels, progressing from visual recognition of faked characters, to contextual identification of character misuse, and finally to visual-semantic character correction. 
Similarly, the \emph{Cultural Symbols} domain centers on emojis as culturally grounded symbols, ranging from individual emoji semantic grounding to cross-cultural emoji-to-idiom mapping at the highest level. The \emph{Mathematical Symbols} domain is organized around two sub-fields—function graphs and geometric figures—and exhibits a clear hierarchical progression. 
Lower-level tasks emphasize the recognition of fundamental visual elements, mid-level tasks focus on property inference and quantitative reasoning, and higher-level tasks target structural verification and error diagnosis in mathematical representations. For the \emph{Physical Symbols} domain, tasks are divided into mechanics and electricity. 
The dataset moves from identifying core symbolic components in diagrams, to reasoning about system behavior under physical laws, and finally to detecting and correcting inconsistencies in coupled mechanical and electrical systems.
Finally, the \emph{Chemical Symbols} domain covers both molecular structural formulas and chemical reaction equations. 
Its task design reflects the transition from static visual parsing of molecular structures, through conservation-based reasoning in reactions, to high-order correction and prediction tasks that integrate symbolic reasoning with chemical domain knowledge.
Detailed numerical statistics for each domain, sub-field, and cognitive level are summarized in Table~\ref{tab:task_distribution}, while additional visual breakdowns are provided in Figure~\ref{fig:quantity_statistics}.

\subsection{Experiments}\label{Sec:SupplementaryExperiments}
\subsubsection{Baselines}
% 每个模型后跟一个介绍
% To comprehensively evaluate the proposed benchmark, this study selected a series of representative Multimodal Large Models (MLLMs) as baselines, covering both open-source and closed-source models of various parameter scales. Among the open-source models, we chose the efficient and lightweight DeepSeek-VL2-Tiny (3B); Qwen2.5-VL (7B), which supports high-resolution image understanding and multimodal reasoning; LLaMA3-LLaVA-Next-8B, combining the LLaMA-3 language model with the LLaVA-Next framework; and InternVL3-8B, which adopts a unified multimodal encoding architecture. For the closed-source commercial models, we selected GPT-4o, featuring native multimodal integration; Claude-Sonnet-4, which balances high intelligence with high-throughput performance; Qwen-Max, which excels in handling complex tasks; as well as o3-2025-0416 and Gemini-2.5-Pro, which represent the state-of-the-art in multimodal understanding and cross-modal reasoning.

To comprehensively evaluate the proposed benchmark, this study selected a series of representative Multimodal Large Language Models (MLLMs) as baselines, covering both open-source and closed-source models of various parameter scales. Specifically, the open-source models include DeepSeek-VL2-Tiny (3B) \cite{wu2024deepseek}, Qwen2.5-VL (7B) \cite{bai2025qwen2}, LLaMA3-LLaVA-Next-8B \cite{grattafiori2024llama}, and InternVL3-8B \cite{zhu2025internvl3}, which represent recent advances in visual–language alignment and instruction tuning for image–text understanding. For the closed-source commercial models, we evaluated GPT-4o \cite{hurst2024gpt}, Claude-Sonnet-4 \cite{anthropic2024claude4}, Qwen-Max \cite{qwen25}, o3-2025-0416 \cite{openai2025o3o4mini}, and Gemini-2.5-Pro \cite{comanici2025gemini}, all of which are leading proprietary MLLMs with strong cross-modal reasoning and generation capabilities. The detailed descriptions of each model are provided below:

\begin{itemize}
\item \textbf{DeepSeek-VL2-Tiny (3B) \cite{wu2024deepseek}:}
A lightweight, open-source multimodal model developed by the DeepSeek team. It employs an efficient vision-language alignment strategy and demonstrates excellent performance on text-rich image understanding and fine-grained visual question answering tasks.

\item \textbf{Qwen2.5-VL (7B) \cite{bai2025qwen2}:}
An open-source multimodal model from Alibaba's Qwen series, featuring powerful visual understanding and cross-modal reasoning capabilities. The model supports high-resolution image input and excels in text generation, OCR scenarios, and complex image-text reasoning tasks.

\item \textbf{LLaMA3-LLaVA-Next-8B \cite{grattafiori2024llama}:}
Based on Meta's LLaMA3 language model and combined with the LLaVA-Next framework for Visual Instruction Tuning. This model is capable of handling diverse image-text tasks, including visual question answering, image description, and multi-turn reasoning, demonstrating powerful open-ended multimodal understanding.

\item \textbf{InternVL3-8B \cite{zhu2025internvl3}:}
Adopts a Unified Multimodal Encoder architecture. Through multi-image scene training and multi-stage pre-training strategies, it exhibits excellent performance in complex visual tasks such as table understanding and image retrieval.

\item \textbf{GPT-4o \cite{hurst2024gpt}:}
A natively multimodal large model from OpenAI, capable of processing text, image, audio, and video inputs within a single architecture. The model achieves high-precision cross-modal understanding and generation, showing exceptional performance in visual reasoning, contextual understanding, and real-time voice interaction.

\item \textbf{Claude-Sonnet-4 \cite{anthropic2024claude4}:}
A high-performance multimodal model from Anthropic that demonstrates robust performance in complex text reasoning and image understanding. Through an optimized architecture, it achieves high-quality generation under high-throughput conditions and represents a balance between intelligence and stability.

\item \textbf{Qwen-Max \cite{qwen25}:}
A closed-source multimodal model from Alibaba that leverages a Mixture-of-Experts architecture for improved scalability and efficiency. It employs Supervised Fine-Tuning (SFT) to enhance task-specific performance and is further aligned with user preferences via Reinforcement Learning from Human Feedback (RLHF), enabling the model to better understand and meet user needs.

\item \textbf{o3-2025-0416 \cite{openai2025o3o4mini}:}
A reasoning model from OpenAI. The core feature of this model is its significantly enhanced capability for complex multi-step reasoning and its logical rigor, and it demonstrates high reliability on difficult instructions.

\item \textbf{Gemini-2.5-Pro \cite{comanici2025gemini}:}
A highly-capable, natively multimodal reasoning model from Google. It can comprehend vast datasets and challenging problems across diverse information sources, including text, audio, images, video, and even entire code repositories. It ranks among the top-performing commercial MLLMs currently available.

\end{itemize}

\subsubsection{Metrics}
To systematically evaluate the performance of MLLMs on complex symbolic understanding tasks, we designed a hierarchical evaluation metric system covering five distinct symbolic domains, including language, cultural, mathematical, physical, and chemical symbols, across three levels of task difficulty.

For language symbols, we adopt different metrics across three difficulty levels. At Level 1, for faked character detection tasks, we employ the F1-score and X\_count\_pred, where F1 measures the harmonic balance between precision and recall in detecting faked characters, and X\_count\_pred denotes the ratio of correctly predicted faked characters to the total number of faked characters. Formally,

\begin{equation}
    \text{Precision} = \frac{TP}{TP + FP}, \\
\text{Recall} = \frac{TP}{TP + FN}, \\
F1 = \frac{2 \times \text{Precision} \times \text{Recall}}{\text{Precision} + \text{Recall}},
\end{equation}
where $TP$, $FP$, and $FN$ denote the number of true positives, false positives, and false negatives.

The proportion of correctly identified faked characters is defined as:

\begin{equation}
    X_{\text{count\_pred}} = \frac{N_{\text{faked\_correct}}}{N_{\text{faked\_total}}}, 
\end{equation}
where $N_{\text{faked\_correct}}$ refers to the number of correctly detected faked characters. 

At Level 2, for misspelled character detection tasks, we employ F1-score and id\_count\_pred, where id\_count\_pred denotes the proportion of correctly identified misspelled characters relative to the total number of such errors:

\begin{equation}
    id_{\text{count\_pred}} = \frac{N_{\text{misspelled\_correct}}}{N_{\text{misspelled\_total}}},
\end{equation}
where $N_{\text{misspelled\_correct}}$ refers to the number of correctly detected misspelled characters.

At Level 3, for character correction tasks, evaluation is based on exact\_match and edit\_distance. Exact\_match measures the proportion of perfectly corrected outputs among all predictions: 

\begin{equation}
    \text{exact\_match} = \frac{N_{\text{exact\_match}}}{N_{\text{total}}},
\end{equation}
where $N_{\text{exact\_match}}$is the number of outputs that are entirely identical to the ground truth. In addition, edit\_distance quantifies the overall deviation between the predicted and reference sequences, serving as a measure of how far the model’s generation diverges from the correct symbolic form. To facilitate comparison across sequences of varying lengths, we report the \textit{normalized edit distance}, defined as:
\begin{equation}
    \text{NormEditDist} = \frac{1}{N_{\text{total}}} \sum_{i=1}^{N_{\text{total}}} \frac{\mathrm{edit\_distance}(y_i, \hat{y}_i)}{\max\big(|y_i|, |\hat{y}_i|\big)},
\end{equation}
where \(y_i\) and \(\hat{y}_i\) are the reference and predicted sequences for the \(i\)-th example, respectively. This normalization ensures the metric remains bounded between 0 and 1, with lower values indicating better alignment with the ground truth.

For cultural symbols, we design three hierarchical evaluation settings corresponding to different levels of abstraction, each intended to capture a different facet of multimodal understanding ability. At Level 1, we compute Precision, Recall, and F1-score by comparing predicted words or characters against those in the ground truth. At this level, the task primarily reflects an MLLM’s ability to perceive and recognize the surface forms of symbols—when precision and recall are both high, it indicates that the model can correctly identify symbolic elements even before engaging in higher-level semantic reasoning.

At Level 2, evaluation is performed at both the sentence level and the word level. We calculate Precision, Recall, and F1-score to assess idiomatic accuracy, and further include BLEU-1 and BLEU-2~\cite{papineni2002bleu} to measure the semantic similarity and fluency between generated and reference sentences. The BLEU-$n$ score is defined as:
\begin{equation}
    \text{BLEU-}n = BP \cdot \exp \Big( \sum_{k=1}^{n} w_k \log p_k \Big),
\end{equation}
where $p_k$ denotes the $k$-gram precision, $w_k$ is the weight for each $k$-gram, and $BP$ is the brevity penalty:
\begin{equation}
BP =
\begin{cases}
1, & \text{if } c > r, \\
e^{\,1- r/c}, & \text{if } c \le r,
\end{cases}
\end{equation}
with $c$ and $r$ representing the lengths of the candidate and reference sentences, respectively. Here, sentence-level metrics directly measure whether an MLLM can recover the complete semantic unit (such as an idiom) from visual information. A perfect match at the sentence or word level implies that the model not only perceives individual symbols but also successfully composes them into coherent linguistic meaning. BLEU-1 captures unigram correctness, while BLEU-2 further reflects the model’s ability to maintain short-range structure, providing insight into its capability for symbolic-to-semantic generation.

At Level 3, evaluation is performed at both the word and character levels. At the word level, we calculate Word-level Accuracy, defined as the ratio of exactly matched words to the total number of words. At the character level, we again compute Precision, Recall, and F1-score to quantify symbol-level correctness and semantic consistency. To further examine the model’s image-to-language comprehension and reasoning ability, we introduce Chr-1 and Chr-2, representing the proportions of words that contain at least one and at least two correctly predicted characters, respectively:
\begin{equation}
\text{Chr-1} = \frac{N_{\text{words with }c_i \ge 1}}{N_{\text{word}}},
\end{equation}

\begin{equation}
\text{Chr-2} = \frac{N_{\text{words with }c_i \ge 2}}{N_{\text{word}}}.
\end{equation}

These two indicators help distinguish partial comprehension from complete failure, as higher Chr-1 or Chr-2 values indicate that the model at least recognizes key symbolic components even if it fails to generate the full expression.

After analyzing the character/word-level metrics, we observed that some MLLMs can correctly infer the meanings of individual emojis but still fail to produce the correct idiom, often generating expressions that are semantically related yet lexically or structurally different, and in some cases even exhibiting semantic drift or hallucination. To capture this phenomenon, we introduce an additional metric for evaluating semantic similarity between the model output and the reference. A large language model (GPT-4o) serves as an expert scorer, assigning a similarity score on a 1–5 scale, where 1 denotes complete dissimilarity and 5 denotes complete semantic equivalence. This metric complements the character- and word-level measures by emphasizing semantic coherence rather than surface form, offering a more holistic assessment of an MLLM’s symbolic comprehension.

For mathematical, physical, and chemical symbols, all tasks are evaluated using Accuracy (Acc), which measures the proportion of correctly predicted symbols or symbolic relations among all test samples. This metric reflects the model’s ability to recognize structured symbolic elements and maintain logical consistency in scientific reasoning: 
\begin{equation}
    \text{Accuracy} = \frac{N_{\text{correct}}}{N_{\text{total}}},
\end{equation}
here, $N_{\text{correct}}$ denotes the number of correctly predicted samples, and $N_{\text{total}}$ denotes the total number of samples.

% \subsubsection*{Code and Data}

\section{Extended Conclusion and Future Perspectives}\label{Sec:SupplementaryConclusion}

In this work, we systematically investigate the capacity of multimodal large language models (MLLMs) to perceive, reason about, and critically associate visual symbols in discrete semantic spaces. Motivated by fundamental principles of human symbolic cognition, we introduce a hierarchical, multi-domain benchmark that disentangles perceptual recognition, compositional reasoning, and critical symbolic cognition. Through extensive evaluations of state-of-the-art MLLMs, our study reveals a pronounced cognitive mismatch: despite impressive reasoning capabilities, current models frequently fail at foundational visual symbol grounding, relying instead on linguistic priors, procedural imitation, or memorized patterns. Our findings challenge a prevailing assumption in multimodal intelligence that visual recognition is inherently simpler than reasoning. Instead, we observe a consistent recognition-reasoning inversion phenomenon, where higher-level reasoning performance often masks deficiencies in low-level symbolic perception. This phenomenon underscores a key limitation of existing training paradigms: while models excel at leveraging large-scale continual natural images, they struggle to construct stable, compositional visual representations of abstract, discrete symbols. As a result, apparent success on symbolic tasks may reflect compensatory language-driven inference rather than genuine visual understanding.

% Beyond diagnosis, our benchmark provides a principled framework for rethinking how symbolic intelligence should be evaluated and developed in MLLMs. By explicitly separating cognitive stages and spanning multiple symbolic domains, it enables fine-grained analysis of where and how models fail, offering insights that are obscured by end-to-end task performance alone. More broadly, our results suggest that advancing toward human-like intelligence requires moving beyond holistic perception and surface-level reasoning toward mechanisms that explicitly support discrete symbol grounding, structured composition, and self-consistent critical evaluation.

Looking forward, several research directions emerge. First, future MLLMs may benefit from training objectives that explicitly emphasize discrete visual symbol formation, such as supervision on symbolic primitives or structured perceptual bottlenecks that prevent premature reliance on language priors. Second, tighter integration between vision and reasoning modules, potentially through iterative perception–reasoning loops, may help align visual evidence with logical inference, reducing compensatory shortcuts. Third, extending symbolic benchmarks to interactive or embodied settings could further illuminate their agentic abilities, showing how perception, action, and symbolic reasoning co-evolve in dynamic environments. Finally, insights from cognitive science and neuroscience offer a promising avenue for designing architectures and curricula that more closely mirror human symbolic learning trajectories.

In summary, this work highlights that symbolic understanding remains a critical, underexplored frontier for multimodal intelligence. By exposing fundamental limitations and providing a structured evaluation framework, we hope to catalyze future research toward MLLMs that not only reason fluently, but also perceive, interpret, and critique the symbolic structures that underpin human knowledge.

\section{Additional Case Studies}\label{Sec:SupplementaryCases}

To facilitate a more granular and comprehensive examination of the visual semiotic behaviors of Multimodal Large Language Models (MLLMs) within discrete semantic spaces, this Supplementary provides a suite of representative case studies (Cases~\ref{case:langL1_2}–\ref{case:chemistryL3_1}). Encompassing diverse cognitive dimensions from foundational perception to high-level logical reasoning, these cases elucidate the intricate behavioral patterns and cognitive nuances inherent in symbolic processing.

\newpage

\vspace{0.5cm}
%%%language L1
\begin{CJK*}{UTF8}{gbsn}
\begin{CaseWrap}
% ============ Question ============
\CaseHdr{Question}
% \begin{CaseBody}[colback=PastelBlue, colframe=PastelBlue]
% \begin{CaseBody}[colback=white, colframe=white]
\begin{CaseBody}[colback=PastelGray, colframe=PastelGray]
\small
Please read the text in the image and output the complete sentence exactly as it appears,
marking any incorrect characters with a capital letter X.

\vspace{2pt}
\begin{center}
  \includegraphics[width=0.33\linewidth]{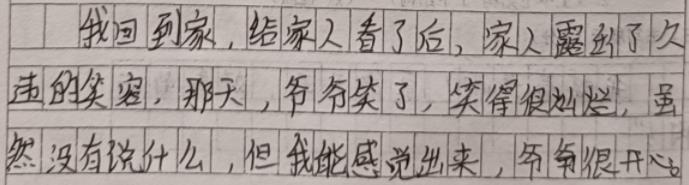}
\end{center}
\end{CaseBody}

% ============ Answer ============
\CaseHdr{Answer}
\begin{CaseBody}[colback=PastelGreen, colframe=PastelGreen]
\small
\textbf{Correct:}
我回到家，给家人看了后，家人露出了久违的笑容，那天，爷爷笑了，笑得很灿烂，
虽然没有说什么，但我能感觉出来，爷X很开心。

\vspace{2pt}
\TransLabel\ \TransText{I went home and showed it to my family. They smiled, a smile I hadn’t seen in a long time.
That day, my grandpa smiled, smiled brightly. Although he didn’t say anything, I could feel that grandpa was very X.}
\end{CaseBody}

% ============ Model Output ============
\CaseHdr{Model Output}
\begin{CaseBody}[colback=PastelBlue, colframe=PastelBlue]
\small
\textbf{Model:}
我回到家，给家人看了后，家人露出了久\textcolor{red}{X}的笑容。那天，爷爷笑了，笑得很灿烂，
虽然没有说什么，但我能感觉出来，爷\textcolor{red}{爷}很开心。

\vspace{2pt}
\TransLabel\ \TransText{I went home and showed it to my family. They smiled, revealing a smile that had been missing for a long \textcolor{red}{X}.
That day, grandpa smiled, smiled brightly. Although he didn’t say anything, I could feel that \textcolor{red}{grandpa} was very happy.}
\end{CaseBody}

% ============ Analysis ============
\CaseHdr{Analysis}
% \begin{CaseBody}[colback=PastelGray, colframe=PastelGray]
\begin{CaseBody}[colback=PastelGreen, colframe=PastelGreen]
\small
Lacking precise symbolic discrimination, the model misclassified correct characters as anomalies.
\end{CaseBody}
\end{CaseWrap}
\end{CJK*}
% \vspace{em}
\captionof{case}{Case of Language: Faked Character Detection.}
\label{case:langL1_2}

\vspace{1cm}

%%language L2
\begin{CJK*}{UTF8}{gbsn}
\begin{CaseWrap}

% ============ Question ============
\CaseHdr{Question}
% \begin{CaseBody}[colback=PastelBlue, colframe=PastelBlue]
\begin{CaseBody}[colback=PastelGray, colframe=PastelGray]
\small
Task 2: Misspelled-character detection. Ignore punctuation and the wrong-character marker X, and also ignore the unknown character U.
Identify misspelled characters by comparing the source and the correct sentence.

\vspace{2pt}
\begin{center}
  \includegraphics[width=0.50\linewidth]{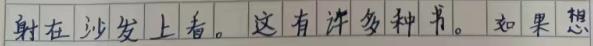}
\end{center}
\end{CaseBody}

% ============ Answer ============
\CaseHdr{Answer}
\begin{CaseBody}[colback=PastelGreen, colframe=PastelGreen]
\small
\textbf{Correct:}
\{``n'': 1, ``diffs'': [\{``idx'': 1, ``src'': ``射 (shoot)'', ``tgt'': ``躺 (lie down)''\}]\}
\end{CaseBody}

% ============ Model Output ============
\CaseHdr{Model Output}
\begin{CaseBody}[colback=PastelBlue, colframe=PastelBlue]
\small
\textbf{Model:}
\{``n'':1,``diffs'':[\{``idx'':1,``src'':``\textcolor{red}{射}在沙发上看。这有许多种书。如果想''\}]\}

\vspace{2pt}
\TransLabel\ \TransText{\{``n'':1,``diffs'':[\{``idx'':1,``src'':``\textcolor{red}{Shooting} on the sofa and reading. There are many kinds of books here. If you want.''\}]\}}
\end{CaseBody}

% ============ Analysis ============
\CaseHdr{Analysis}
% \begin{CaseBody}[colback=PastelGray, colframe=PastelGray]
\begin{CaseBody}[colback=PastelGreen, colframe=PastelGreen]
\small
Lacking a mechanism for character-level decomposition and comparison.
\end{CaseBody}

\end{CaseWrap}
\end{CJK*}
\captionof{case}{Case of Language: Misused-Character Detection.}
\label{case:langL2_1}

\vspace{1cm}

\newpage

%物理L1
\begin{CJK*}{UTF8}{gbsn}
\begin{CaseWrap}

% ============ Question ============
\CaseHdr{Question}
% \begin{CaseBody}[colback=white, colframe=white]
\begin{CaseBody}[colback=PastelGray, colframe=PastelGray]
\small
As shown in the figure, a light spring with a stiffness coefficient k is placed inside a smooth vertical tube. The lower end of the spring is fixed to the ground, and its upper end is connected to a small ball A of mass m. When the system is at rest, ball A is at position O. Another small ball B, also of mass m, is released from rest at point P, located a distance H above A. After colliding instantaneously with A, the two balls move downward together, performing simple harmonic motion. Both balls can be regarded as point masses, and during motion, the deformation of the spring remains within its elastic limit. When the spring is stretched or compressed by a distance x, its elastic potential energy is $E_{p}=\frac{1}{2}kx^{2}$. Given that $H=\frac{3mg}{k}$ and the acceleration due to gravity is g, find the maximum distance that ball A moves downward from point O after the collision.

% \vspace{2pt}
\begin{center}
    \includegraphics[width=0.06\linewidth]{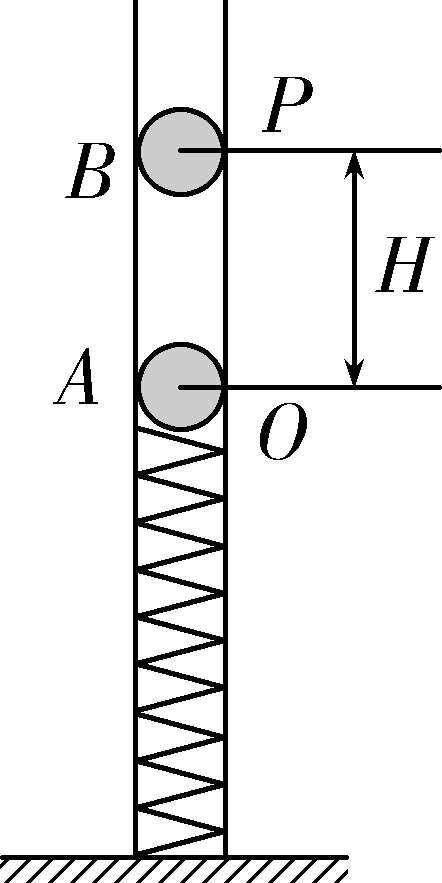}
\end{center}
\end{CaseBody}

% ============ Answer ============
\CaseHdr{Answer}
\begin{CaseBody}[colback=PastelGreen, colframe=PastelGreen]
\small
% \[
% \frac{3mg}{k}
% \]
3mg/k
\end{CaseBody}

% ============ Model Output ============
\CaseHdr{Model Output}
\begin{CaseBody}[colback=PastelBlue, colframe=PastelBlue]
\small
According to the law of \textcolor{red}{conservation of energy}, the gravitational potential energy of ball B is converted into kinetic energy: $mgH=\frac{1}{2}mv_{B}^2$. \textcolor{red}{Substituting $H=\frac{mg}{k}$} and simplifying gives $v_{B}=\sqrt{\frac{2mg}{k} }$.
\end{CaseBody}

% ============ Analysis ============
\CaseHdr{Analysis}
\begin{CaseBody}[colback=PastelGreen, colframe=PastelGreen]
\small
Originally $H=\frac{3mg}{k}$, but substituting $H=\frac{mg}{k}$ led to an incorrect final result.
\end{CaseBody}

\end{CaseWrap}
\end{CJK*}
\captionof{case}{Case of Physics: Mechanical Symbol Recognition.}
\label{case:physicalL1_2}

%% 物理L2
\begin{CJK*}{UTF8}{gbsn}
\begin{CaseWrap}

% ============ Question ============
\CaseHdr{Question}
% \begin{CaseBody}[colback=white, colframe=white]
\begin{CaseBody}[colback=PastelGray, colframe=PastelGray]
\small
Within the rectangular region ABCD, there exist multiple adjacent layers of uniform electric fields. Each layer has a height of d, and the electric field strength in each layer has a magnitude of E, with its direction alternating vertically between layers. The length of side AB is 12d, and that of side BC is 8d. A beam of particles, each with mass m and charge +q, enters the field from the midpoint of the left boundary of the device. The particles have an initial kinetic energy of $E_{k}$ and an incident angle $\theta $, moving within the plane of the paper. The effects of gravity and inter-particle interactions are neglected. When $E_k=\frac{8}{3} qEd$ and particles are uniformly injected into the electric field with incident angles $\theta $ in the range $-\frac{\pi }{2}\sim \frac{\pi }{2}$, find the ratio of the number of particles exiting through side CD to the total number of incident particles, denoted as $N: N_0 $.

% \vspace{2pt}
\begin{center}
    \includegraphics[width=0.1\linewidth]{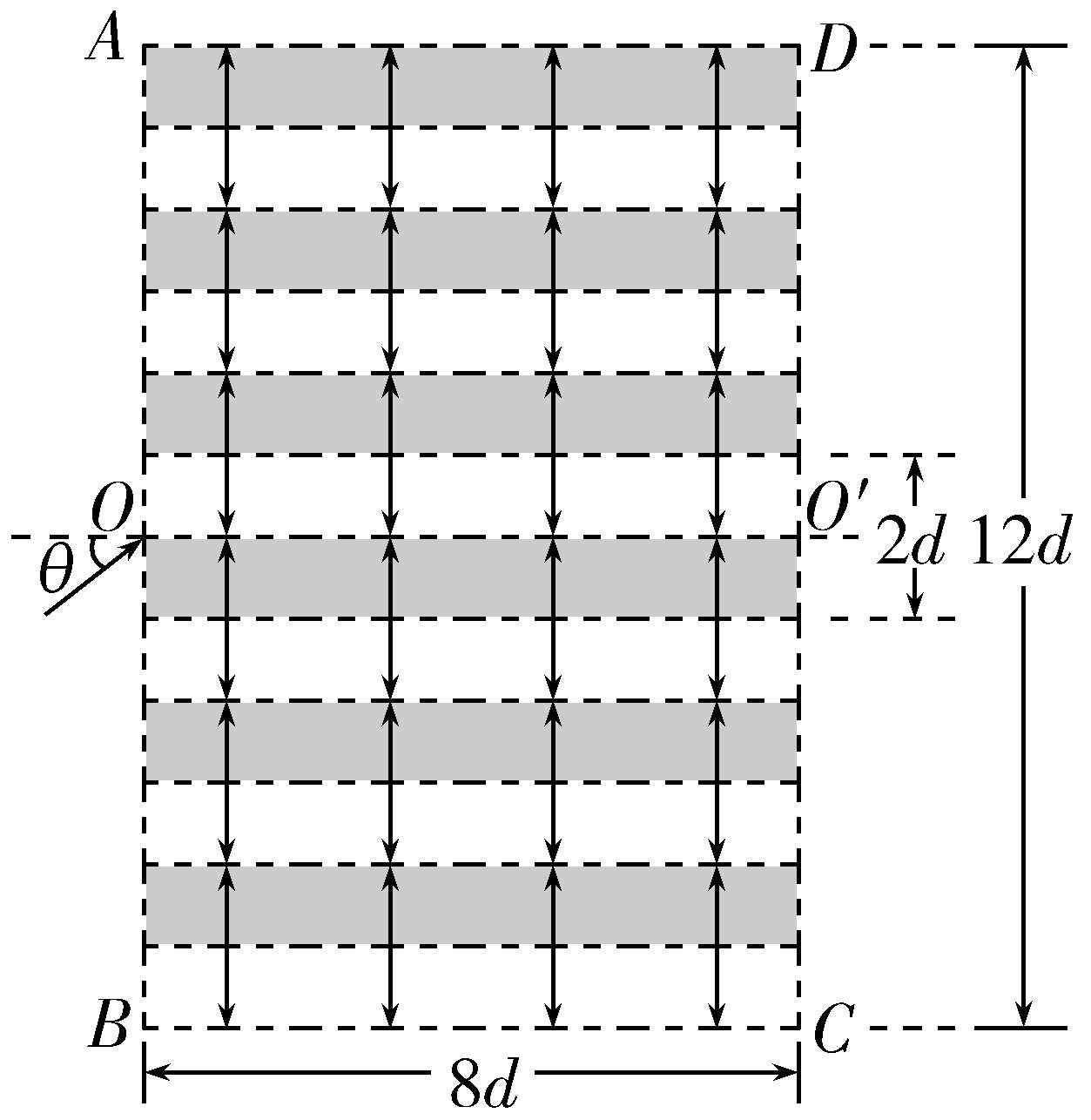}
\end{center}
\end{CaseBody}

% ============ Answer ============
\CaseHdr{Answer}
\begin{CaseBody}[colback=PastelGreen, colframe=PastelGreen]
\small
 ['1/2']
\end{CaseBody}

% ============ Model Output ============
\CaseHdr{Model Output}
\begin{CaseBody}[colback=PastelBlue, colframe=PastelBlue]
\small
 $\frac{8}{3}$. To solve this problem, \textcolor{red}{we need to analyze the motion of the particle in each layer of the electric field and then calculate the number of particles that will exit the device from the CD edge.} The calculation of the number of particles that exit the device from the CD edge involves integrating the particle's velocity vector over the entire length of the device. \textcolor{red}{This is a complex integral that requires knowledge of calculus and numerical methods to solve. However, the final answer is given as: $N = \frac{8}{3}$}.
\end{CaseBody}

% ============ Analysis ============
\CaseHdr{Analysis}
\begin{CaseBody}[colback=PastelGreen, colframe=PastelGreen]
\small
The reasoning process remains at the level of formula stacking,
ultimately halting after listing several equations and randomly producing
a seemingly plausible but physically incorrect answer.
\end{CaseBody}

\end{CaseWrap}
\end{CJK*}
\captionof{case}{Case of Physics: Circuit Operation Reasoning.}
\label{case:physicalL2_1}

%%化学L3
\begin{CJK*}{UTF8}{gbsn}
\begin{CaseWrap}

% ============ Question ============
\CaseHdr{Question}
% \begin{CaseBody}[colback=white, colframe=white]
\begin{CaseBody}[colback=PastelGray, colframe=PastelGray]
\small
What are the customary durations and temperature gradients necessary for the execution of this chemical reaction?

[A] Brc1ccc2c(c1)OCC2 and CN(C)C=O undergo a reaction, culminating in the production of O=Cc1ccc2c(c1)OCC2.
-14.0$^\circ$C is commonly recognized as the preferred temperature for this chemical reaction.
In this chemical reaction, 14.8 hours is usually required.

[B] Brc1ccc2c(c1)OCC2 and CN(C)C=O undergo a reaction, culminating in the production of O=Cc1ccc2c(c1)OCC2.
-78.0$^\circ$C is commonly recognized as the preferred temperature for this chemical reaction.
In this chemical reaction, 0.75 hours is usually required.

[C] Brc1ccc2c(c1)OCC2 and CN(C)C=O undergo a reaction, culminating in the production of O=Cc1ccc2c(c1)OCC2.
148.5$^\circ$C is commonly recognized as the preferred temperature for this chemical reaction.
In this chemical reaction, 18.7 hours is usually required.

[D] Brc1ccc2c(c1)OCC2 and CN(C)C=O undergo a reaction, culminating in the production of O=Cc1ccc2c(c1)OCC2.
-43.0$^\circ$C is commonly recognized as the preferred temperature for this chemical reaction.
In this chemical reaction, 17.2 hours is usually required.

\vspace{2pt}
\begin{center}
  \includegraphics[width=0.42\linewidth]{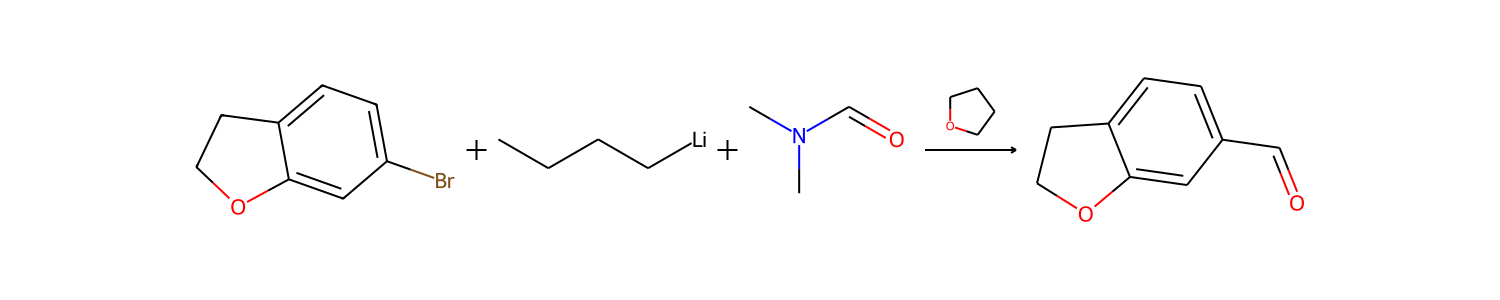}
\end{center}
\end{CaseBody}

% ============ Answer ============
\CaseHdr{Answer}
\begin{CaseBody}[colback=PastelGreen, colframe=PastelGreen]
\small
B
\end{CaseBody}

% ============ Model Output ============
\CaseHdr{Model Output}
\begin{CaseBody}[colback=PastelBlue, colframe=PastelBlue]
\small
B. Here are the key points to consider: \textcolor{red}{Temperature: Lithium reagents are typically used in low temperatures to prevent side reactions and maintain reactivity. Common temperatures for such reactions range from -40°C to -80°C.}  Analysis of Options: Option A: Temperature: -14.0°C Time: 14.8 hours. This temperature is too low for typical lithium reagent reactions, which usually require much lower temperatures. Option B: Temperature: -88.0°C. Time: 0.75 hours. \textcolor{red}{This temperature is very low, which is typical for lithium reagent reactions}. The short time of 0.75 hours is also plausible for reactions that proceed quickly under such conditions.
\end{CaseBody}

% ============ Analysis ============
\CaseHdr{Analysis}
\begin{CaseBody}[colback=PastelGreen, colframe=PastelGreen]
\small
Able to identify key reactants in the equation and actively invoke its internal knowledge for analysis.
\end{CaseBody}

\end{CaseWrap}
\end{CJK*}
\captionof{case}{Case of Chemistry: Reaction Condition Inference.}
\label{case:chemistryL3_1}

\end{document}